# Superhuman performance in urology board questions by an explainable large language model enabled for context integration of the European Association of Urology guidelines: the UroBot study

**Authors:** Martin J. Hetz[a, †], Nicolas Carl[a,b, †], Sarah Haggenmüller[a], Christoph Wies[a,c], Maurice Stephan Michel[b], Frederik Wessels[b*], Titus J. Brinker[a#*]

**Affiliations:**

a: Digital Biomarkers for Oncology Group, German Cancer Research Center (DKFZ), Heidelberg, Germany

b: Department of Urology, University Medical Center Mannheim, Ruprecht-Karls University of Heidelberg, Mannheim, Germany

c: Medical Faculty, University of Heidelberg, Heidelberg, Germany

† These authors contributed equally.

* These authors jointly supervised this work.

# Corresponding author: titus.brinker@dkfz.de

Word count: 2433/2500

Figures & Tables: 3 Figures, 1 Table

# Abstract

**Background:** Large Language Models (LLMs) are transforming medical Question-Answering (medQA) by leveraging vast amounts of medical literature. Despite promising results, LLM performance is limited by outdated training data and lack of explainability, impeding clinical translation.

**Objective:** To develop and evaluate an urology-specialised chatbot (UroBot) against state-of-the art models and urologists' performance in answering urological board questions in a fully clinician-verifiable manner.

**Methods:** We developed UroBot based on the GPT-3.5, GPT-4, and GPT-4o models of OpenAI, utilising retrieval augmented generation (RAG) and the 2023 European Association of Urology (EAU) guidelines. UroBot was benchmarked along the models GPT-3.5, GPT-4, GPT-4o and Uro_Chat. The evaluation involved ten runs against 200 European Board of Urology (EBU) In-Service Assessment (ISA) questions, with the performance measured by the mean Rate of Correct Answers (RoCA).

**Results:** UroBot-4o achieved the highest RoCA, with an average of 88.4%, outperforming GPT-4o (77.6%) by 10.8%, is clinician-verifiable and demonstrated the highest level of agreement between runs as measured by Fleiss' Kappa ($\kappa = 0.979$). In comparison, the average performance of urologists on urological board questions is 68.7% as reported by the literature.

**Limitations:** The reliance on 200 multiple choice questions from the EBU committee is not representative of the full range of scenarios in clinical practice.

**Conclusions and clinical implication:** UroBot is clinician-verifiable and substantially more accurate as compared to both performance of published models and urologists in answering board-questions, encouraging translation to care. We provide code and instructions to rebuild UroBot for further development.

# 1. Introduction

Researchers are exploring the potential of Large Language Models (LLMs) to tackle medical queries, a frontier that promises to extend how knowledge is accessed in healthcare [1–3]. LLMs are artificial neural networks comprising billions of parameters, trained with a broad spectrum of texts mainly sourced from the internet, which includes medical text sources [3–5]. A recent study assessed the performance of multiple LLMs in answering over 2000 oncological multiple-choice questions. The OpenAI developed LLM *GPT-4* achieved the highest rate of correct answers with 68.7% [6]. The growing interest across medical specialties in utilising LLMs for medical Question-Answering (medQA) is culminating in the performance evaluation of LLMs in written medical examinations [3,7–10]. Although LLM models demonstrate remarkable performance in zero-shot abilities, their capabilities are constrained by the training-data used, which can be wrong or outdated rapidly. The performances presented by *Rydzewski et al.*[6] and other studies evaluating LLM performance in medQA show impressive but in total limited capabilities [2-7]. If LLMs would be employed for medical purposes, their accuracy, reliability and explainability become critical, as these models could significantly impact healthcare decisions, diagnosis, and treatment [3].

It is, however, possible to augment commercial or open source LLMs in order to increase the performance by fine-tuning methods or by using Retrieval Augmented Generation (RAG) [11–13]. This has led *Khene et al.*[13] to adapt a LLM based on evidence-based knowledge with the aim to improve performance. This so-called Uro_Chat was based on GPT-3.5 turbo[14] using RAG to implement the uro-oncological guidelines published by the European Association of Urology (EAU) [13,15]. In a subsequent performance test, Uro_Chat was able to answer 61 out of 100 In-Service Assessment questions (ISA) of the European Board of Urology (EBU) correctly which are integrated in board exams e.g. in Austria, Switzerland and the Netherlands and thus, would have barely passed a fictitious EBU exam [16, 17]. Notably, a performance evaluation of GPT-3.5, GPT-4, and Bing-AI (now Copilot [18]) using 100 EBU ISA questions revealed 58-62%, 63-77% and 73-81% correct answers [8]. The average urologist completes the ISA of the EBU with a grade of 68.7% (SD = 6.62) [19]. These results suggest that Uro_Chat performs comparably to GPT-3.5 but is less effective than GPT-4, Copilot or the average human ISA participant. Nonetheless, Uro_Chat presented an interesting approach by incorporating evidence-based knowledge from guidelines into its design.

While these developments are exciting, LLMs suffer from the so-called hallucination problem, which describes a phenomenon where the model generates text that is incorrect, nonsensical, or not real. Accordingly, both the European General Data Protection Regulation as well as clinicians demand explainability of AI for end-users, ensuring verifiability of decisions especially in clinical settings [20,21]. Thus, development should involve continual collaborations between AI developers and clinical end users [22].

Conclusively, the objective of this interdisciplinary work is to develop and evaluate an explainable urology-specialised chatbot based on current EAU guidelines against state-of-the art models and urologists' performance in answering urological board questions in a clinician-verifiable manner.

The design of UroBot will incorporate all 2023 guidelines published by the EAU and is engineered to display to which parts of the corresponding documents its answer was based on. UroBots' accuracy is benchmarked against the most recent LLMs, including GPT-3.5, GPT-4, and GPT-4o. Uro_Chat will be re-benchmarked in order to provide a direct comparison to our optimised model. We investigate whether a substantial enhancement (defined as an increase of more than 5 percent points in accuracy) is achievable compared to the currently most accurate LLM (GPT-4o). A technical approach to auto-update its knowledge database for its context-based decisions is introduced. All code is made available and instructions are provided for full reproducibility of our study.

# 2. Materials and Methods

## 2.1 Materials

The EAU guidelines were downloaded as PDF files from the EAUs online resources on March 12, 2024 [23]. In total the 20 PDF files contained over 2000 pages of text. The raw text was extracted from the PDA files using Python and split into text chunks with a size of approximately 1000 characters, with each chunk tagged with metadata indicating whether it is a paragraph or a table and the corresponding page. Following the segmentation of the text into discrete chunks, the data was transformed into vector embeddings via the open source embedding model 'mxbai-embed-large-v1' by Mixedbread-ai [24], with the resulting vectors being stored in a Chroma database [25]. The instructions and Python code for running UroBot and reproducing our experiments are available via Github (*https://github.com/marjohe/UroBot*) [26].

In order to assess the performance of UroBot and the competing models under investigation, 200 multiple-choice questions provided by the EBU Committee were digitised by transferring them to an Excel spreadsheet. The ISA EBU questions are confidential and are only obtainable through purchase on the official EBU website [16].

## 2.2 Question Answering Pipeline

To answer the questions, the OpenAI text generation API was used with the models 'gpt-3.5-turbo-0125', 'gpt-4-turbo-2024-04-09' and 'gpt-4o-2024-05-13' (referred to as 'GPT-3.5', 'GPT-4' and 'GPT-4o'). Subsequently, RAG was utilised to make the models urology-informed and based UroBot-3.5, UroBot-4 and UroBot-4o on the respective LLM. In order to provide UroBot with the necessary context, the query was vectorised using the embedding model described above. Similar vectors and their corresponding text chunks are retrieved from the database. The system prompt of UroBot is then modified to contain the retrieved context and prompted to answer the question based on the retrieved content. A visual representation of our question-answering pipeline is illustrated in **Figure 1**. The exact prompts utilised in this study can be accessed in the **Supplementary Material**. In all experiments, the number of retrieved chunks was set to ten and the sampling temperature to 0.1. The sampling temperature in a text generation model is a factor in determining the randomness of the generated text. A lower temperature (close to 0) results in a more focused and predictable output, with the selection of words tending towards the most probable.

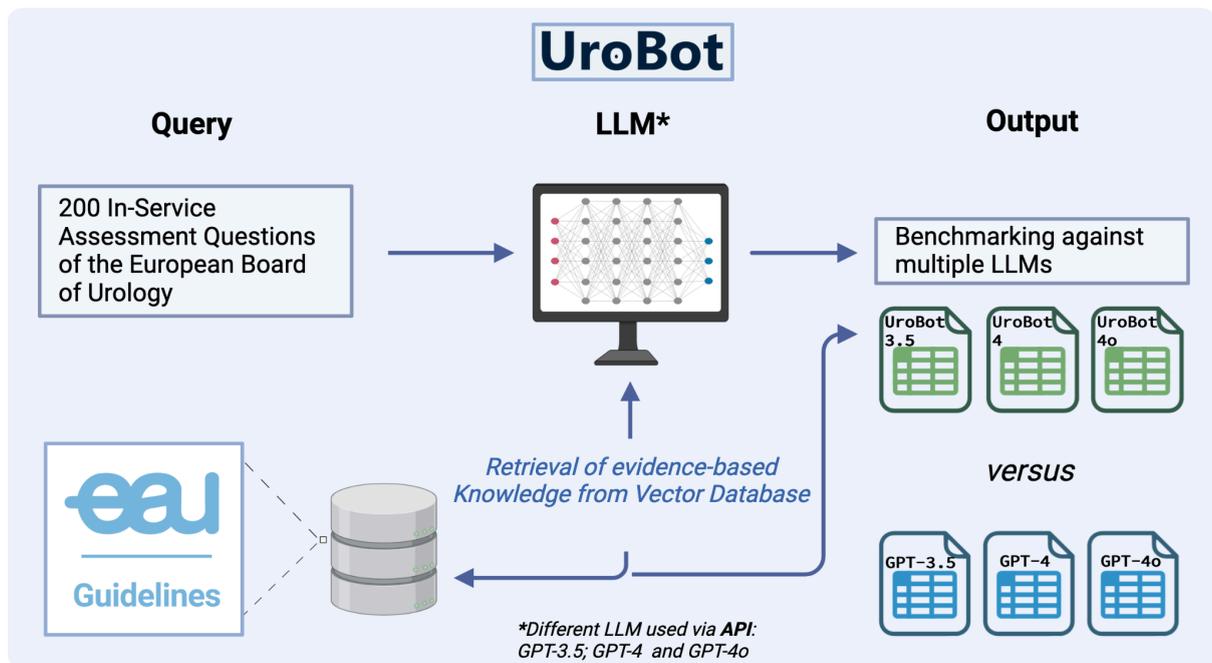

**Figure 1. Design of UroBot and the benchmarking procedure.** A vector database was created by embedding full-text information of all 20 available EAU guidelines. The LLM retrieves information from the vector database to create outputs. 200 ISA EBU questions are used to benchmark UroBot against all models under investigation (i.e., GPT-3.5, GPT-4, and GPT-4o). LLM = Large Language Model, API = application programming interface.

We provided the design of the end-user interface via GitHub [26]. It displays a query and respective answer with the exact document and text snippet where the information was received from in order to provide clinician-verifiable LLM outputs (Figure 2).

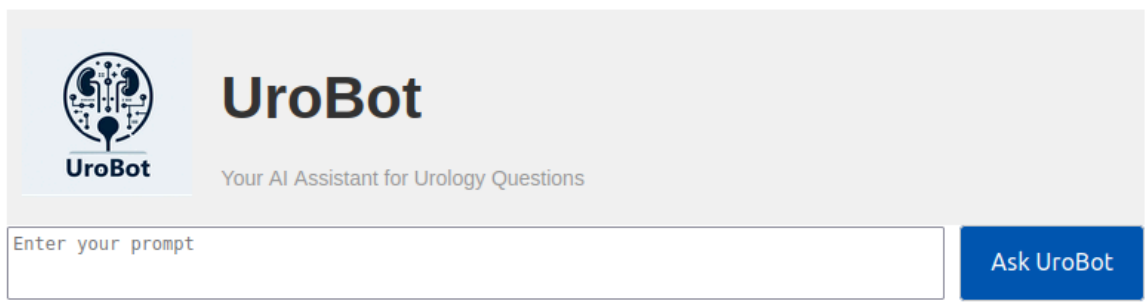

**UroBot**

Your AI Assistant for Urology Questions

Enter your prompt

Ask UroBot

## Predefined Questions

How common is bladder cancer?        What is the prognosis of renal carcinoma?

How is penile cancer treated?        I have blood in my urine, what could that be?

You:
How common is bladder cancer?

UroBot:
Bladder cancer is the seventh most commonly diagnosed cancer in the male population worldwide, and it ranks tenth when considering both genders. The worldwide age-standardised incidence rate is 9.5 per 100,000 person/years for men and 2.4 for women (Document ID 806). Can I help you with any other questions on this topic?

Document ID 806: 2.2 The 2021 publication was peer reviewed prior to print. Review Future goals 2.3 The findings of the ongoing 'Individual Patient Data Validation of the Definition of bacillus Calmette-Guérin (BCG) Failure/BCG Unresponsive in Patients with Non-muscle Invasive Urothelial Carcinoma of the Bladder: an international multicentre retrospective study' will be included in the future update of the NMIBC Guidelines. 3. EPIDEMIOLOGY, AETIOLOGY AND PATHOLOGY Epidemiology 3.1 Bladder cancer (BC) is the seventh most commonly diagnosed cancer in the male population worldwide, and it is the tenth when both genders are considered [13]. The worldwide age-standardised incidence rate (per 100,000 person/years) is 9.5 in men and 2.4 in women [13]. In the European Union, the age-standardised incidence rate is 20 in men and 4.6 in women [13]. Worldwide, the BC age-standardised mortality rate (per 100,000 person/years) is 3.3 for men vs. 0.86 for women [13]. Bladder cancer incidence and mortality rates vary across countries due to differences in risk factors, detection and diagnostic practices, and variations in access to, and delivery of, healthcare. Additionally, epidemiological variations have

**Figure 2. Screenshot of the user-interface of UroBot.** An end-user (e.g. a urologist) can ask any urology question to UroBot and receives an answer in less than 5 seconds. The lower part of the image displays exactly which documents and text-snippets within the documents were used to provide the answer, making the answer verifiable for the end-user.

## 2.3 Evaluation

A total of 200 ISA EBU questions were posed to all models. In order to analyse the consistency of the different models, we repeated this procedure 10 times. The mean Rate of Correct Answers (RoCA), including 95% confidence intervals (CI), was used as a performance metric and is calculated by dividing the number of correct answers per run by the total number of questions, averaged over ten runs. With regard to the OpenAI-based models, an automated benchmark was conducted utilising the text generation API of OpenAI, without the use of a chat history. With regard to Uro_Chat, all questions were entered into the provided web interface in a consecutive manner ten times, with the responses then being entered into an Excel spreadsheet. Upon the presentation of each new question, the web interface of Uro_Chat was reloaded.

Fleiss' Kappa κ was used to evaluate the consistency of the LLM answers, whilst simultaneously accounting for any agreement that might occur by chance. Fleiss' Kappa ranges from -1 to 1, with 1 representing perfect agreement, 0 denoting agreement expected by chance, and -1 indicating perfect disagreement. Therefore, higher values indicate a higher degree of agreement between runs. For statistical comparisons of the LLM performance across all 200 questions, pairwise 2-sided t-tests were applied. A significance level of alpha 0.05 was set for all analyses. Significance levels were adjusted to 0.005 (m = 10) according to Bonferroni correction[27] in case of multiple tests to adjust for the increased risk of type I errors due to multiple comparisons.

# 3. Results

## 3.1 Performance of the Models under Investigation

An overview of the results is provided in **Table 1**. The highest RoCA was reached by UroBot-4o with an average of 0.884 (95% CI: 0.881-0.886). The highest mean RoCA of a standard model (without RAG) was reached by GPT-4o with 0.776 (95% CI: 0.771-0.781). UroBot-4o outperformed the best standard model by a Δ of 0.108 pairwise 2-sided t-test, p<0.001; see **Table 1**). The performance of UroBot was dependent on the LLM used. The mean RoCA were 0.722, (95% CI: 0.717-0.728) for UroBot-3.5, 0.863 (95% CI: 0.860-0.867) for UroBot-4, and 0.884 (95% CI: 0.881-0.886) for UroBot-4o.

The lowest performance was observed for Uro_Chat with a RoCA of 0.547 (95% CI: 0.538-0.555) and GPT-3.5 turbo with 0.492 (95% CI: 0.484-0.500).

|  | Uro_Chat | GPT-3.5 | GPT-4 | UroBot-3.5 | GPT-4o | UroBot-4 | UroBot-4o |
|---|---|---|---|---|---|---|---|
| **Mean Rate of Correct Answers (95% CI)** | 0.547 (0.538-0.555) | 0.492 (0.484-0.500) | 0.719 (0.715-0.723) | 0.722 (0.717-0.728) | 0.776 (0.771-0.781) | 0.863 (0.860-0.867) | 0.884 (0.881-0.886) |
| **Majority Voting Rate of Correct Answers** | 0.57 | 0.5 | 0.72 | 0.715 | 0.78 | 0.865 | 0.885 |
| **p-value (vs UroBot-4o)** | < 0.001 | < 0.001 | < 0.001 | < 0.001 | < 0.001 | < 0.001 | ref. |
| **Fleiss' Kappa Value** | 0.704 | 0.866 | 0.945 | 0.92 | 0.943 | 0.966 | 0.979 |

**Table 1. Benchmark results for all models aggregated over ten runs.** The mean Rate of Correct Answers (RoCA), majority vote RoCA, p-values and the Fleiss' Kappa value (κ) are shown. The results demonstrate that UroBot-4o exhibits a markedly higher RoCA than its competitors. Furthermore, it has the highest κ, indicating the highest consistency in answering. An overview of the results is provided in **Figure 3**.

## 3.2 Consistency of Performance across Test Runs

The reliability of the generated model answers was assessed by presenting the same question to various models on ten separate occasions. Generally, the agreement between test runs was *substantial* across all LLMs and test runs according to the interpretation guideline of *Landis and Koch* [28]. UroBot-4o showed the highest agreement, demonstrating almost perfect consistency between test runs (κ = 0.979), followed by UroBot-4 (κ = 0.966) and GPT-4 (κ = 0.945). The lowest agreement was observed with Uro_Chat, which nevertheless showed *substantial agreement* between test runs (κ = 0.70).

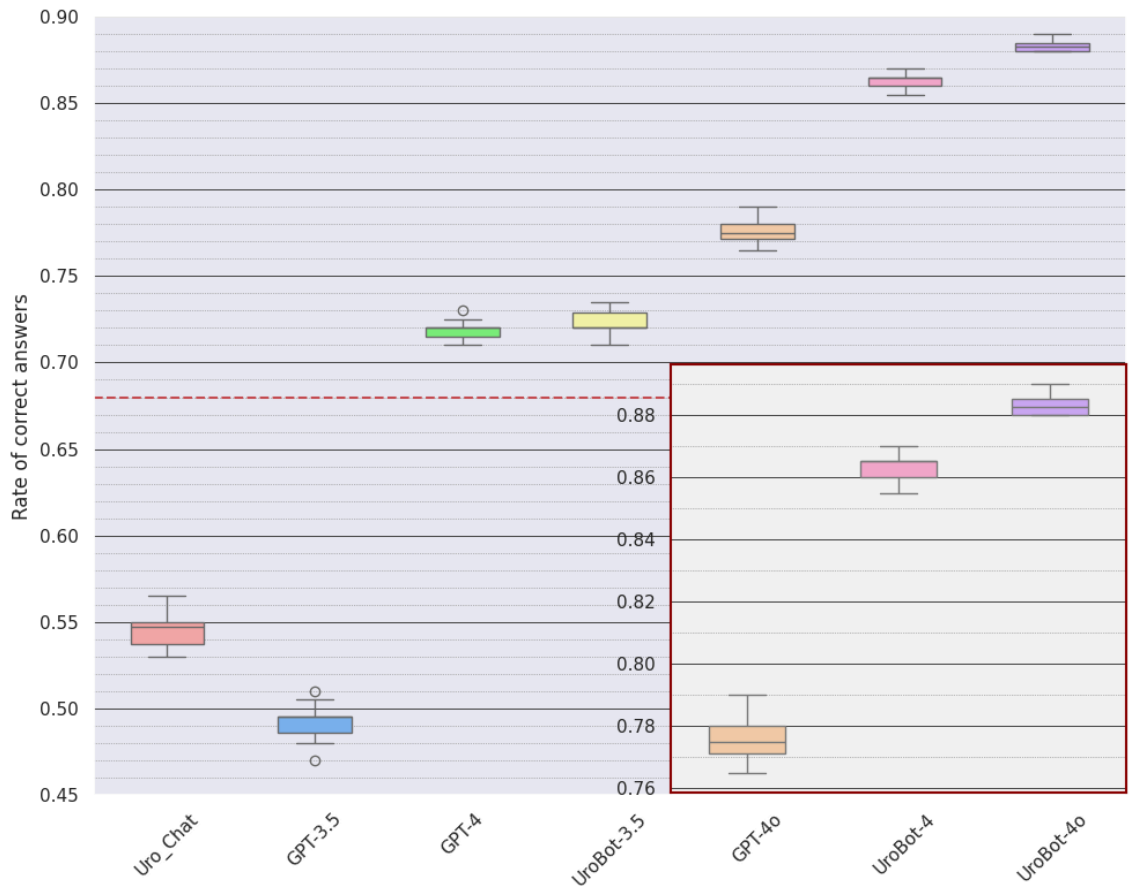

**Figure 3. The figure presents boxplots illustrating the rate of correct answers (RoCA) for all models under investigation.** The bottom right-hand corner of the plot presents a magnified view of the data for the models GPT-4o, UroBot-4 and UroBot-4o. The plot demonstrates that UroBot-4o significantly outperforms all other methods, with a lower variance than the other methods. The dashed line represents the mean performance of urologists.

# 4. Discussion

In this study, UroBot-4o, a LLM retrieving text information from a vector database containing all 20 available EAU guidelines, demonstrated superior performance in the realm of urological board-question answering, achieving a remarkable average RoCA of 88.4%. This performance not only surpassed the GPT-4o model by a significant margin of 10.8 percentage points, but it also greatly exceeded the average performance of urologists on board questions, which is reported at 68.7% in the literature [19]. Furthermore, UroBot-4o's outputs were clinician-verifiable, ensuring that the responses align with medical standards. The model exhibited the highest level of consistency across multiple runs, as evidenced by a Fleiss' Kappa value of 0.979, indicating almost perfect agreement. The retrieval mechanism employed by RAG is crucial in providing contextually appropriate information, which the LLM then effectively utilises to produce accurate answers. Results of the benchmarking show that UroBot significantly outperforms the best available models, surpasses previously reported performance levels in the literature and the average ISA attendee's performance [6,8,17,19]. The lowest results for correctness and consistency were observed in Uro_Chat and GPT-3.5 turbo.

While off-the-shelf LLMs demonstrate impressive capabilities in medical question answering (medQA), a significant limitation is the insufficient performance [6,10]. This constraint can be mitigated through in-context learning, for example via prompt engineering. However, this method is hindered by the fixed length of the input string, which can accommodate only a limited number of pages of text data, rendering this approach impractical [29]. RAG is an advanced form of in-context learning that can utilise extensive knowledge bases. Unlike traditional in-context learning, RAG incorporates an external knowledge retrieval system. Recent models from other AI research groups have also demonstrated impressive improvements using RAG, consistent with the success observed in our study. *Ferber et al.*[12] achieved 84% correct statements in a subset of medical oncology questions with a RAG-enhanced model compared to 57% with the standard model [12].

It is of critical importance to embed medical knowledge into a model, particularly in rapidly evolving fields such as urology, where guidelines are frequently updated. RAG offers scalability and easy-to-implement updates, which provides a method for maintaining current and evidence-based assistance tools in patient care and could therefore benefit clinicians as an informational or educational tool. Our study demonstrates significant performance improvements using a feasible way to incorporate evidence-base knowledge into LLMs using the RAG method. Importantly, RAG might be the key to paving the way for clinically useful large language models.

## 4.2 Limitations

Although leveraging medical state-of-the-art training material for the EBU examination exclusively, the reliance on 200 multiple-choice questions from the EBU Committee may not be fully representative of the full range of scenarios in clinical practice. Future research could build upon this work by testing UroBot with additional questions and clinical situations, allowing practitioners to interact with UroBot in daily tasks. It is recommended that open-ended questions are included in future assessments to further evaluate the reasoning abilities of UroBot. Furthermore, this study does not investigate the effects of different prompts on performance. This is a topic that should be explored in future research. Also, a urologist in a board exam does not have all 20 EAU guidelines available and can retrieve data from them, as our model did. If time allows and a urologist had an 10 hours per board question to look up the information in the guidelines within thousands of pages, he may even achieve 100% accuracy on board questions. Nonetheless, UroBot gives precise and verifiable answers within less than 5 seconds. In terms of speed *and* accuracy, UroBots' performance is superhuman.

Although RAG is effective in reducing the occurrence of hallucinations, it does not entirely prevent them [30,31]. LLMs may still utilise information outside the provided context to answer the question. For clinical use, a mechanism for detecting hallucinations may be necessary. Furthermore, the retrieved text data may exhibit a high degree of textual similarity to the query, yet may lack the necessary relevance to answer the question. We used a commercial LLM as a backbone for RAG (ChatGPT 4o), nonetheless, open source architectures of comparable performance are also available (e.g. LLAMA-3 by Meta Platforms Inc., USA)

In addition, if LLMs were to be used as an information source or if the decision-making process of clinicians would be influenced, LLMs must be approved as a medical device [32,33]. In our ongoing research, we are exploring open-ended prompts and will develop a user-friendly interface displaying outputs similar to EAU guidelines. UroBot will feature distinct physician and patient user-modes, tailored to specific needs, in order to ensure effectiveness and safety. This approach aims to enhance the consistency and accuracy of LLMs in clinical applications, providing reliable AI-assistance tools incorporating evidence-based knowledge and individual patient data at the same time. These improvements act as the cornerstone for further clinical research.

# 5. Conclusion

This study highlights the potential of enhancing LLMs with evidence-based guidelines to improve their performance in specialised medical fields. UroBot is clinician-verifiable and substantially more accurate as compared to both performance of published models and urologists in answering board-questions, encouraging translation to care and showcasing the benefit of retrieval augmented generation. We provide code and instructions to rebuild UroBot and its user-interface for further development. As we further refine these models and expand their knowledge bases, the integration of LLMs into routine medical practice becomes an increasingly viable and beneficial prospect. Nonetheless, regulatory approval as medical devices is imperative to all LLMs prior to their implementation into care [32]. In addition, current software is still unable to replace doctor-patient relationships and can and should not take responsibility for medical decisions since many patients would be opposed to this, especially in oncologic settings [21]. Further research is needed to evaluate the integration of individual patient information into UroBot to answer individual patients' questions.

# Author Contributions

**MJH** developed the text extraction pipeline, the question-answering pipeline and user-interface, designed the methodology, conducted the investigation, performed the formal analysis, validated the findings, authored the original draft, and created visual representations.

**NC** developed the study concept, designed the methodology, conducted the investigation, performed the formal analysis, validated the findings, authored the original draft, and created visual representations.

**SH** supported the conceptualization, contributed to the validation process, participated in the review and editing of the manuscript, and supervised the project team.

**CW** contributed to the statistical analysis and participated in the review and editing of the manuscript.

**MSM** provided resources and participated in the review and editing of the manuscript

**FW** contributed to the conceptualization, reviewed and edited the manuscript and supervised the project.

**TJB** led the conceptualization, contributed to the validation process, provided resources, reviewed and edited the manuscript, supervised the project team, administered the project, and acquired funding.

# Conflict of Interest Statement

The authors declare the following financial interests/personal relationships which may be considered as potential competing interests: Dr. Brinker would like to disclose that he is the owner of Smart Health Heidelberg GmbH (Handschuhsheimer Landstr. 9/1, 69120 Heidelberg, Germany; https://smarthealth.de), outside the submitted work. Dr. Wessels would like to disclose that he advises for AstraZeneca, Janssen and Adon Health outside of the submitted work. The other authors have no conflicts of interest to declare.

# Declaration of Generative AI and AI-assisted Technologies in the Writing Process

During the preparation of this work the author(s) used GPT4 and GPT4o in order to improve readability. After using this tool, the author(s) reviewed and edited the content as needed and take full responsibility for the content of the publication.

## Supplementary Material for the Manuscript:

**Title:** Superhuman Performance in Urology Board Questions by an Explainable Large Language Model Enabled for Context Integration of the European Association of Urology Guidelines: The UroBot Study

**Authors:** Martin J. Hetz[a, †], Nicolas Carl[a,b, †], Sarah Haggenmüller[a], Christoph Wies[a,c], Maurice Stephan Michel[b], Frederik Wessels[b*], Titus J. Brinker[a#*]

[†] These authors contributed equally. [*] These authors jointly supervised this work.

The full code is available via Github under: https://github.com/marjohe/UroBot or using the QR-Code below:

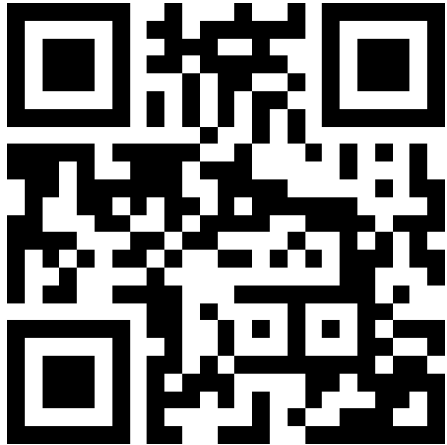

## Supplementary Results:

**Table 1:** Summarised results of the different models over ten runs.

| model | mean_roca | majority_vote_roca | 1 | 2 | 3 | 4 | 5 | 6 | 7 | 8 | 9 | 10 |
|---|---|---|---|---|---|---|---|---|---|---|---|---|
| Uro_Chat | 0.547 | 0.57 | 0.55 | 0.565 | 0.53 | 0.55 | 0.545 | 0.53 | 0.545 | 0.535 | 0.565 | 0.55 |
| GPT_3.5 | 0.492 | 0.5 | 0.495 | 0.48 | 0.495 | 0.495 | 0.49 | 0.505 | 0.495 | 0.51 | 0.485 | 0.47 |
| GPT_4 | 0.719 | 0.72 | 0.72 | 0.72 | 0.72 | 0.73 | 0.725 | 0.715 | 0.715 | 0.72 | 0.715 | 0.71 |
| UroGPT_3.5 | 0.722 | 0.715 | 0.73 | 0.735 | 0.72 | 0.725 | 0.72 | 0.72 | 0.715 | 0.73 | 0.72 | 0.71 |
| UroGPT_4 | 0.863 | 0.865 | 0.865 | 0.87 | 0.865 | 0.865 | 0.865 | 0.87 | 0.86 | 0.86 | 0.855 | 0.86 |
| GPT_4o | 0.776 | 0.78 | 0.78 | 0.77 | 0.775 | 0.79 | 0.775 | 0.77 | 0.775 | 0.78 | 0.765 | 0.78 |
| **UroGPT_4o** | **0.883** | **0.885** | **0.89** | **0.885** | **0.885** | **0.88** | **0.88** | **0.89** | **0.885** | **0.88** | **0.88** | **0.88** |

**Table 2:** Each Spreadsheet represents one LLM with its respective test runs. Runs 1 through 10 are displayed in the columns. The rows represent the Question ID, displayed in the same sequence as in the MCQ ISA booklet 2021-2022. The observations represent correct (1) or incorrect (0) answers. For further reading please address the MCQ ISA booklet. The questions are under copyright and not publicly available. They can be purchased on www.ebu.com.

| ID | Uro_Chat_1 | Uro_Chat_2 | Uro_Chat_3 | Uro_Chat_4 | Uro_Chat_5 | Uro_Chat_6 | Uro_Chat_7 | Uro_Chat_8 | Uro_Chat_9 | Uro_Chat_10 |
|---|---|---|---|---|---|---|---|---|---|---|
| 1 | 1 | 0 | 0 | 0 | 0 | 0 | 0 | 0 | 0 | 0 |
| 2 | 0 | 0 | 0 | 1 | 1 | 0 | 0 | 0 | 0 | 0 |
| 3 | 0 | 0 | 0 | 1 | 1 | 1 | 1 | 1 | 1 | 1 |

| ID | Uro_Chat_1 | Uro_Chat_2 | Uro_Chat_3 | Uro_Chat_4 | Uro_Chat_5 | Uro_Chat_6 | Uro_Chat_7 | Uro_Chat_8 | Uro_Chat_9 | Uro_Chat_10 |
|---|---|---|---|---|---|---|---|---|---|---|
| 4 | 0 | 0 | 0 | 0 | 1 | 0 | 0 | 0 | 0 | 1 |
| 5 | 1 | 1 | 1 | 1 | 1 | 1 | 1 | 1 | 1 | 1 |
| 6 | 1 | 1 | 1 | 1 | 1 | 1 | 1 | 1 | 1 | 1 |
| 7 | 1 | 1 | 1 | 1 | 1 | 1 | 1 | 1 | 1 | 1 |
| 8 | 1 | 0 | 1 | 1 | 1 | 1 | 1 | 1 | 1 | 1 |
| 9 | 1 | 1 | 1 | 1 | 1 | 1 | 1 | 1 | 1 | 1 |
| 10 | 0 | 0 | 0 | 1 | 1 | 1 | 1 | 1 | 1 | 1 |
| 11 | 0 | 0 | 0 | 0 | 0 | 0 | 0 | 0 | 0 | 0 |
| 12 | 1 | 1 | 1 | 1 | 1 | 0 | 0 | 0 | 0 | 0 |
| 13 | 0 | 0 | 0 | 1 | 1 | 1 | 1 | 1 | 1 | 1 |
| 14 | 1 | 1 | 1 | 1 | 1 | 1 | 1 | 1 | 1 | 1 |
| 15 | 1 | 1 | 1 | 1 | 1 | 1 | 1 | 1 | 1 | 1 |
| 16 | 0 | 0 | 0 | 0 | 0 | 0 | 0 | 0 | 0 | 0 |
| 17 | 1 | 1 | 1 | 1 | 1 | 1 | 1 | 1 | 1 | 1 |
| 18 | 0 | 0 | 0 | 0 | 0 | 0 | 0 | 0 | 0 | 0 |
| 19 | 0 | 0 | 0 | 0 | 0 | 0 | 0 | 0 | 0 | 0 |
| 20 | 0 | 0 | 0 | 0 | 0 | 0 | 0 | 0 | 0 | 0 |
| 21 | 0 | 0 | 0 | 1 | 1 | 1 | 1 | 1 | 1 | 1 |
| 22 | 0 | 1 | 0 | 0 | 0 | 0 | 0 | 0 | 0 | 0 |
| 23 | 1 | 0 | 0 | 1 | 1 | 1 | 1 | 1 | 1 | 1 |
| 24 | 0 | 0 | 0 | 0 | 0 | 0 | 0 | 0 | 0 | 0 |
| 25 | 1 | 0 | 0 | 1 | 1 | 1 | 1 | 1 | 1 | 1 |
| 26 | 0 | 0 | 0 | 0 | 1 | 0 | 0 | 0 | 0 | 1 |
| 27 | 0 | 0 | 0 | 0 | 0 | 0 | 0 | 0 | 0 | 0 |
| 28 | 1 | 1 | 1 | 1 | 1 | 1 | 1 | 1 | 1 | 1 |
| 29 | 1 | 1 | 1 | 1 | 1 | 1 | 1 | 1 | 1 | 1 |

| ID | Uro_Chat_1 | Uro_Chat_2 | Uro_Chat_3 | Uro_Chat_4 | Uro_Chat_5 | Uro_Chat_6 | Uro_Chat_7 | Uro_Chat_8 | Uro_Chat_9 | Uro_Chat_10 |
|----|-----------|-----------|-----------|-----------|-----------|-----------|-----------|-----------|-----------|------------|
| 30 | 0 | 0 | 0 | 1 | 1 | 1 | 1 | 1 | 1 | 1 |
| 31 | 1 | 0 | 0 | 0 | 0 | 0 | 0 | 0 | 0 | 0 |
| 32 | 0 | 0 | 0 | 1 | 0 | 0 | 0 | 1 | 1 | 0 |
| 33 | 1 | 1 | 1 | 1 | 1 | 1 | 1 | 1 | 1 | 1 |
| 34 | 1 | 1 | 1 | 1 | 1 | 1 | 1 | 1 | 1 | 1 |
| 35 | 0 | 1 | 1 | 0 | 0 | 0 | 0 | 0 | 0 | 0 |
| 36 | 1 | 1 | 1 | 0 | 0 | 0 | 1 | 0 | 0 | 0 |
| 37 | 0 | 0 | 0 | 0 | 0 | 0 | 0 | 0 | 0 | 0 |
| 38 | 0 | 0 | 0 | 0 | 0 | 0 | 0 | 0 | 0 | 0 |
| 39 | 0 | 1 | 1 | 0 | 1 | 0 | 0 | 0 | 0 | 0 |
| 40 | 0 | 0 | 0 | 1 | 0 | 0 | 0 | 0 | 0 | 0 |
| 41 | 0 | 0 | 0 | 0 | 0 | 0 | 0 | 0 | 0 | 0 |
| 42 | 0 | 0 | 0 | 1 | 0 | 0 | 0 | 0 | 0 | 0 |
| 43 | 0 | 0 | 0 | 0 | 0 | 0 | 0 | 0 | 0 | 0 |
| 44 | 1 | 1 | 1 | 1 | 1 | 1 | 1 | 1 | 1 | 1 |
| 45 | 1 | 1 | 1 | 1 | 1 | 1 | 1 | 1 | 1 | 1 |
| 46 | 1 | 1 | 1 | 1 | 1 | 1 | 1 | 1 | 1 | 1 |
| 47 | 1 | 1 | 1 | 1 | 1 | 1 | 1 | 1 | 1 | 1 |
| 48 | 1 | 1 | 1 | 1 | 0 | 0 | 0 | 0 | 0 | 0 |
| 49 | 1 | 1 | 1 | 0 | 1 | 0 | 1 | 0 | 1 | 0 |
| 50 | 1 | 1 | 0 | 0 | 0 | 1 | 0 | 0 | 0 | 0 |
| 51 | 0 | 0 | 0 | 0 | 0 | 0 | 0 | 0 | 0 | 0 |
| 52 | 0 | 0 | 0 | 0 | 0 | 0 | 0 | 0 | 0 | 0 |
| 53 | 0 | 0 | 0 | 0 | 0 | 0 | 0 | 0 | 0 | 0 |
| 54 | 1 | 0 | 1 | 0 | 0 | 0 | 0 | 0 | 0 | 0 |
| 55 | 1 | 1 | 1 | 1 | 1 | 1 | 1 | 1 | 1 | 1 |

| ID | Uro_Chat_1 | Uro_Chat_2 | Uro_Chat_3 | Uro_Chat_4 | Uro_Chat_5 | Uro_Chat_6 | Uro_Chat_7 | Uro_Chat_8 | Uro_Chat_9 | Uro_Chat_10 |
|---|---|---|---|---|---|---|---|---|---|---|
| 56 | 0 | 1 | 1 | 1 | 1 | 1 | 1 | 1 | 1 | 1 |
| 57 | 0 | 0 | 0 | 0 | 0 | 0 | 0 | 0 | 0 | 0 |
| 58 | 1 | 1 | 1 | 1 | 1 | 1 | 1 | 1 | 1 | 1 |
| 59 | 1 | 1 | 1 | 1 | 1 | 1 | 1 | 1 | 1 | 1 |
| 60 | 1 | 1 | 1 | 1 | 0 | 1 | 1 | 1 | 1 | 1 |
| 61 | 0 | 0 | 0 | 1 | 1 | 1 | 1 | 1 | 1 | 1 |
| 62 | 1 | 0 | 1 | 0 | 0 | 0 | 0 | 0 | 0 | 0 |
| 63 | 1 | 1 | 1 | 1 | 1 | 1 | 1 | 1 | 1 | 1 |
| 64 | 0 | 0 | 1 | 1 | 1 | 1 | 1 | 1 | 1 | 1 |
| 65 | 0 | 0 | 0 | 0 | 0 | 0 | 0 | 0 | 0 | 0 |
| 66 | 1 | 1 | 1 | 1 | 1 | 1 | 1 | 1 | 1 | 1 |
| 67 | 1 | 1 | 1 | 1 | 1 | 1 | 1 | 1 | 1 | 1 |
| 68 | 1 | 1 | 1 | 1 | 1 | 1 | 1 | 1 | 1 | 1 |
| 69 | 1 | 1 | 1 | 1 | 0 | 1 | 1 | 1 | 1 | 1 |
| 70 | 1 | 1 | 1 | 0 | 1 | 1 | 1 | 1 | 1 | 1 |
| 71 | 1 | 1 | 1 | 1 | 0 | 0 | 1 | 1 | 1 | 1 |
| 72 | 1 | 1 | 1 | 1 | 1 | 1 | 1 | 1 | 1 | 1 |
| 73 | 1 | 1 | 1 | 1 | 1 | 1 | 1 | 1 | 1 | 1 |
| 74 | 1 | 1 | 1 | 1 | 1 | 1 | 1 | 1 | 1 | 1 |
| 75 | 1 | 1 | 1 | 1 | 1 | 1 | 1 | 1 | 1 | 1 |
| 76 | 1 | 1 | 1 | 1 | 1 | 1 | 1 | 1 | 1 | 1 |
| 77 | 0 | 0 | 0 | 0 | 0 | 0 | 0 | 0 | 0 | 0 |
| 78 | 0 | 0 | 0 | 0 | 0 | 0 | 0 | 0 | 0 | 0 |
| 79 | 1 | 1 | 1 | 1 | 1 | 1 | 1 | 1 | 1 | 1 |
| 80 | 0 | 0 | 0 | 0 | 0 | 0 | 0 | 0 | 0 | 0 |
| 81 | 0 | 0 | 0 | 0 | 0 | 0 | 0 | 0 | 0 | 0 |

| ID | Uro_Chat_1 | Uro_Chat_2 | Uro_Chat_3 | Uro_Chat_4 | Uro_Chat_5 | Uro_Chat_6 | Uro_Chat_7 | Uro_Chat_8 | Uro_Chat_9 | Uro_Chat_10 |
|-----|-----|-----|-----|-----|-----|-----|-----|-----|-----|-----|
| 82 | 1 | 1 | 1 | 0 | 0 | 0 | 0 | 0 | 0 | 0 |
| 83 | 1 | 1 | 1 | 1 | 1 | 1 | 1 | 1 | 1 | 1 |
| 84 | 0 | 0 | 0 | 0 | 0 | 0 | 0 | 0 | 0 | 0 |
| 85 | 0 | 0 | 0 | 0 | 0 | 0 | 0 | 0 | 0 | 0 |
| 86 | 1 | 1 | 1 | 0 | 0 | 0 | 0 | 0 | 0 | 0 |
| 87 | 1 | 1 | 0 | 0 | 0 | 0 | 0 | 0 | 0 | 0 |
| 88 | 1 | 1 | 1 | 1 | 1 | 1 | 1 | 1 | 1 | 1 |
| 89 | 1 | 1 | 1 | 1 | 1 | 1 | 1 | 1 | 1 | 1 |
| 90 | 1 | 1 | 1 | 1 | 1 | 1 | 1 | 1 | 1 | 1 |
| 91 | 1 | 1 | 1 | 1 | 1 | 1 | 1 | 1 | 1 | 1 |
| 92 | 1 | 1 | 1 | 1 | 1 | 1 | 1 | 1 | 1 | 1 |
| 93 | 1 | 1 | 1 | 1 | 1 | 1 | 1 | 1 | 1 | 1 |
| 94 | 1 | 1 | 1 | 1 | 1 | 1 | 1 | 1 | 1 | 1 |
| 95 | 1 | 1 | 1 | 1 | 1 | 1 | 1 | 1 | 1 | 1 |
| 96 | 1 | 1 | 1 | 1 | 1 | 1 | 1 | 1 | 1 | 1 |
| 97 | 1 | 1 | 1 | 1 | 1 | 1 | 1 | 1 | 1 | 1 |
| 98 | 0 | 0 | 0 | 0 | 1 | 0 | 0 | 0 | 0 | 0 |
| 99 | 0 | 1 | 0 | 0 | 1 | 1 | 1 | 1 | 1 | 1 |
| 100 | 1 | 1 | 1 | 1 | 0 | 0 | 1 | 1 | 1 | 0 |
| 101 | 0 | 1 | 0 | 1 | 1 | 1 | 1 | 1 | 1 | 1 |
| 102 | 1 | 1 | 1 | 1 | 1 | 1 | 1 | 1 | 1 | 1 |
| 103 | 0 | 0 | 0 | 0 | 0 | 0 | 0 | 0 | 0 | 0 |
| 104 | 0 | 0 | 0 | 1 | 1 | 1 | 1 | 1 | 1 | 1 |
| 105 | 1 | 1 | 1 | 1 | 1 | 1 | 1 | 1 | 1 | 1 |
| 106 | 0 | 0 | 0 | 0 | 0 | 0 | 0 | 0 | 0 | 0 |
| 107 | 0 | 1 | 1 | 1 | 1 | 1 | 1 | 1 | 1 | 1 |

| ID | Uro_Chat_1 | Uro_Chat_2 | Uro_Chat_3 | Uro_Chat_4 | Uro_Chat_5 | Uro_Chat_6 | Uro_Chat_7 | Uro_Chat_8 | Uro_Chat_9 | Uro_Chat_10 |
|---|---|---|---|---|---|---|---|---|---|---|
| 108 | 0 | 1 | 1 | 0 | 1 | 1 | 1 | 1 | 1 | 1 |
| 109 | 0 | 0 | 0 | 0 | 0 | 0 | 0 | 0 | 0 | 0 |
| 110 | 1 | 1 | 0 | 0 | 0 | 0 | 0 | 0 | 0 | 0 |
| 111 | 1 | 1 | 1 | 1 | 1 | 1 | 1 | 1 | 1 | 1 |
| 112 | 0 | 0 | 0 | 0 | 0 | 0 | 0 | 0 | 0 | 0 |
| 113 | 1 | 1 | 1 | 1 | 1 | 1 | 1 | 1 | 1 | 1 |
| 114 | 1 | 0 | 1 | 1 | 1 | 1 | 1 | 1 | 1 | 1 |
| 115 | 0 | 0 | 0 | 0 | 0 | 0 | 0 | 0 | 0 | 0 |
| 116 | 1 | 1 | 1 | 0 | 0 | 0 | 0 | 0 | 0 | 0 |
| 117 | 1 | 1 | 1 | 0 | 0 | 0 | 0 | 0 | 0 | 0 |
| 118 | 1 | 1 | 1 | 1 | 1 | 1 | 1 | 1 | 1 | 1 |
| 119 | 1 | 1 | 1 | 1 | 1 | 1 | 1 | 1 | 1 | 1 |
| 120 | 0 | 0 | 0 | 1 | 1 | 1 | 1 | 1 | 1 | 1 |
| 121 | 0 | 1 | 1 | 1 | 0 | 1 | 1 | 1 | 1 | 1 |
| 122 | 0 | 0 | 0 | 1 | 1 | 1 | 1 | 1 | 1 | 1 |
| 123 | 1 | 1 | 1 | 1 | 1 | 1 | 1 | 1 | 1 | 1 |
| 124 | 1 | 1 | 1 | 1 | 1 | 1 | 1 | 1 | 1 | 1 |
| 125 | 1 | 1 | 1 | 1 | 1 | 1 | 1 | 1 | 1 | 1 |
| 126 | 0 | 0 | 0 | 0 | 0 | 0 | 0 | 0 | 0 | 0 |
| 127 | 1 | 1 | 1 | 0 | 0 | 0 | 0 | 0 | 0 | 0 |
| 128 | 1 | 1 | 1 | 1 | 1 | 1 | 1 | 1 | 1 | 1 |
| 129 | 0 | 1 | 0 | 0 | 0 | 0 | 0 | 0 | 0 | 0 |
| 130 | 0 | 0 | 0 | 1 | 1 | 1 | 1 | 1 | 1 | 1 |
| 131 | 0 | 0 | 0 | 0 | 0 | 0 | 0 | 0 | 0 | 0 |
| 132 | 0 | 0 | 0 | 0 | 0 | 0 | 0 | 0 | 0 | 0 |
| 133 | 1 | 1 | 1 | 1 | 1 | 1 | 1 | 1 | 1 | 1 |

| ID | Uro_Chat_1 | Uro_Chat_2 | Uro_Chat_3 | Uro_Chat_4 | Uro_Chat_5 | Uro_Chat_6 | Uro_Chat_7 | Uro_Chat_8 | Uro_Chat_9 | Uro_Chat_10 |
|---|---|---|---|---|---|---|---|---|---|---|
| 134 | 1 | 1 | 1 | 1 | 1 | 1 | 1 | 1 | 1 | 1 |
| 135 | 0 | 0 | 0 | 0 | 0 | 0 | 0 | 0 | 0 | 0 |
| 136 | 1 | 1 | 1 | 1 | 1 | 1 | 1 | 1 | 1 | 1 |
| 137 | 0 | 0 | 0 | 0 | 0 | 0 | 0 | 0 | 0 | 0 |
| 138 | 0 | 1 | 0 | 0 | 0 | 0 | 0 | 0 | 0 | 0 |
| 139 | 0 | 0 | 0 | 0 | 0 | 0 | 0 | 0 | 0 | 0 |
| 140 | 0 | 0 | 0 | 1 | 1 | 1 | 0 | 1 | 1 | 1 |
| 141 | 1 | 1 | 1 | 0 | 0 | 0 | 0 | 0 | 0 | 0 |
| 142 | 1 | 1 | 1 | 1 | 1 | 1 | 1 | 1 | 1 | 1 |
| 143 | 1 | 1 | 1 | 1 | 1 | 1 | 1 | 1 | 1 | 1 |
| 144 | 0 | 0 | 0 | 0 | 0 | 0 | 0 | 0 | 0 | 0 |
| 145 | 1 | 1 | 1 | 1 | 1 | 1 | 1 | 1 | 1 | 1 |
| 146 | 0 | 0 | 0 | 0 | 0 | 0 | 0 | 0 | 0 | 0 |
| 147 | 1 | 1 | 1 | 1 | 1 | 1 | 1 | 1 | 1 | 1 |
| 148 | 0 | 0 | 0 | 0 | 0 | 0 | 0 | 0 | 0 | 0 |
| 149 | 0 | 0 | 0 | 0 | 0 | 0 | 0 | 0 | 0 | 0 |
| 150 | 0 | 0 | 0 | 0 | 0 | 0 | 0 | 0 | 0 | 0 |
| 151 | 1 | 0 | 1 | 1 | 1 | 1 | 1 | 1 | 1 | 1 |
| 152 | 0 | 0 | 0 | 0 | 0 | 0 | 0 | 0 | 0 | 0 |
| 153 | 0 | 0 | 0 | 0 | 0 | 0 | 0 | 0 | 0 | 0 |
| 154 | 0 | 0 | 0 | 0 | 0 | 0 | 0 | 0 | 0 | 0 |
| 155 | 1 | 1 | 1 | 1 | 1 | 1 | 1 | 1 | 1 | 1 |
| 156 | 1 | 1 | 1 | 0 | 0 | 0 | 0 | 0 | 1 | 1 |
| 157 | 0 | 0 | 0 | 0 | 0 | 0 | 0 | 0 | 0 | 0 |
| 158 | 0 | 0 | 1 | 0 | 0 | 0 | 1 | 1 | 1 | 1 |
| 159 | 0 | 0 | 0 | 1 | 1 | 1 | 1 | 1 | 1 | 1 |

| ID | Uro_Chat_1 | Uro_Chat_2 | Uro_Chat_3 | Uro_Chat_4 | Uro_Chat_5 | Uro_Chat_6 | Uro_Chat_7 | Uro_Chat_8 | Uro_Chat_9 | Uro_Chat_10 |
|---|---|---|---|---|---|---|---|---|---|---|
| 160 | 1 | 1 | 1 | 0 | 0 | 0 | 0 | 0 | 0 | 0 |
| 161 | 1 | 1 | 1 | 1 | 1 | 1 | 1 | 1 | 1 | 1 |
| 162 | 1 | 1 | 1 | 1 | 1 | 1 | 1 | 1 | 1 | 1 |
| 163 | 1 | 1 | 1 | 1 | 1 | 1 | 1 | 1 | 1 | 1 |
| 164 | 1 | 1 | 1 | 1 | 1 | 1 | 1 | 1 | 1 | 1 |
| 165 | 1 | 1 | 1 | 1 | 1 | 1 | 1 | 1 | 1 | 1 |
| 166 | 1 | 1 | 1 | 1 | 1 | 1 | 1 | 1 | 1 | 1 |
| 167 | 1 | 1 | 1 | 1 | 1 | 1 | 1 | 1 | 1 | 1 |
| 168 | 1 | 1 | 1 | 0 | 0 | 0 | 0 | 0 | 0 | 0 |
| 169 | 1 | 1 | 0 | 0 | 0 | 0 | 0 | 0 | 0 | 0 |
| 170 | 1 | 0 | 0 | 1 | 1 | 1 | 1 | 1 | 1 | 1 |
| 171 | 0 | 1 | 0 | 0 | 0 | 0 | 0 | 0 | 0 | 0 |
| 172 | 0 | 0 | 0 | 0 | 0 | 0 | 0 | 0 | 0 | 0 |
| 173 | 1 | 1 | 1 | 1 | 1 | 1 | 1 | 1 | 1 | 1 |
| 174 | 0 | 0 | 0 | 0 | 0 | 0 | 0 | 0 | 0 | 0 |
| 175 | 1 | 1 | 1 | 1 | 1 | 1 | 1 | 1 | 1 | 1 |
| 176 | 1 | 0 | 0 | 1 | 0 | 0 | 0 | 0 | 1 | 0 |
| 177 | 0 | 0 | 0 | 0 | 0 | 0 | 0 | 0 | 0 | 0 |
| 178 | 0 | 0 | 0 | 0 | 0 | 0 | 1 | 0 | 0 | 1 |
| 179 | 0 | 0 | 0 | 0 | 0 | 0 | 0 | 0 | 0 | 0 |
| 180 | 0 | 1 | 0 | 0 | 1 | 1 | 0 | 0 | 1 | 1 |
| 181 | 1 | 1 | 1 | 1 | 1 | 1 | 1 | 1 | 1 | 1 |
| 182 | 1 | 1 | 1 | 1 | 1 | 1 | 1 | 1 | 1 | 1 |
| 183 | 1 | 1 | 1 | 1 | 1 | 1 | 1 | 1 | 1 | 1 |
| 184 | 1 | 1 | 1 | 0 | 1 | 0 | 0 | 0 | 1 | 0 |
| 185 | 0 | 1 | 0 | 0 | 0 | 0 | 0 | 0 | 0 | 0 |

| ID | Uro_Chat_1 | Uro_Chat_2 | Uro_Chat_3 | Uro_Chat_4 | Uro_Chat_5 | Uro_Chat_6 | Uro_Chat_7 | Uro_Chat_8 | Uro_Chat_9 | Uro_Chat_10 |
|---|---|---|---|---|---|---|---|---|---|---|
| 186 | 0 | 0 | 0 | 1 | 0 | 0 | 1 | 0 | 0 | 0 |
| 187 | 0 | 0 | 0 | 0 | 0 | 0 | 0 | 0 | 0 | 0 |
| 188 | 1 | 0 | 0 | 0 | 0 | 0 | 0 | 0 | 0 | 0 |
| 189 | 0 | 1 | 0 | 1 | 0 | 1 | 0 | 0 | 1 | 0 |
| 190 | 0 | 0 | 0 | 0 | 0 | 0 | 0 | 0 | 0 | 0 |
| 191 | 1 | 1 | 1 | 1 | 1 | 1 | 1 | 1 | 1 | 1 |
| 192 | 0 | 0 | 0 | 0 | 0 | 0 | 0 | 0 | 0 | 0 |
| 193 | 1 | 1 | 0 | 0 | 1 | 1 | 1 | 1 | 1 | 1 |
| 194 | 1 | 1 | 1 | 1 | 1 | 1 | 1 | 1 | 1 | 1 |
| 195 | 1 | 1 | 1 | 1 | 1 | 1 | 1 | 1 | 1 | 1 |
| 196 | 0 | 0 | 0 | 0 | 0 | 0 | 0 | 0 | 0 | 0 |
| 197 | 1 | 1 | 1 | 1 | 1 | 1 | 1 | 1 | 1 | 1 |
| 198 | 0 | 0 | 0 | 0 | 0 | 0 | 0 | 0 | 0 | 0 |
| 199 | 1 | 1 | 1 | 1 | 1 | 1 | 1 | 1 | 1 | 1 |
| 200 | 0 | 0 | 0 | 0 | 0 | 0 | 0 | 0 | 0 | 0 |

| ID | GPT_3.5_1 | GPT_3.5_2 | GPT_3.5_3 | GPT_3.5_4 | GPT_3.5_5 | GPT_3.5_6 | GPT_3.5_7 | GPT_3.5_8 | GPT_3.5_9 | GPT_3.5_10 |
|---|---|---|---|---|---|---|---|---|---|---|
| 1 | 0 | 0 | 0 | 0 | 0 | 0 | 0 | 0 | 0 | 0 |
| 2 | 0 | 0 | 0 | 0 | 0 | 0 | 0 | 0 | 0 | 0 |
| 3 | 0 | 0 | 0 | 0 | 0 | 0 | 0 | 0 | 0 | 0 |
| 4 | 0 | 0 | 0 | 0 | 0 | 0 | 0 | 0 | 0 | 0 |
| 5 | 1 | 1 | 1 | 1 | 1 | 1 | 1 | 1 | 1 | 1 |
| 6 | 1 | 1 | 1 | 1 | 1 | 1 | 1 | 1 | 1 | 1 |

| ID | GPT_3.5_1 | GPT_3.5_2 | GPT_3.5_3 | GPT_3.5_4 | GPT_3.5_5 | GPT_3.5_6 | GPT_3.5_7 | GPT_3.5_8 | GPT_3.5_9 | GPT_3.5_10 |
|---|---|---|---|---|---|---|---|---|---|---|
| 7 | 0 | 0 | 0 | 0 | 0 | 0 | 0 | 0 | 0 | 0 |
| 8 | 1 | 1 | 1 | 1 | 1 | 1 | 1 | 1 | 1 | 1 |
| 9 | 0 | 1 | 0 | 1 | 1 | 1 | 0 | 1 | 0 | 0 |
| 10 | 1 | 1 | 1 | 1 | 1 | 1 | 1 | 1 | 1 | 1 |
| 11 | 0 | 0 | 0 | 0 | 0 | 0 | 0 | 0 | 0 | 0 |
| 12 | 1 | 1 | 0 | 0 | 0 | 1 | 0 | 0 | 1 | 1 |
| 13 | 0 | 0 | 1 | 0 | 0 | 0 | 0 | 0 | 0 | 0 |
| 14 | 0 | 0 | 0 | 0 | 0 | 0 | 1 | 0 | 0 | 0 |
| 15 | 1 | 1 | 1 | 1 | 1 | 1 | 1 | 1 | 1 | 1 |
| 16 | 1 | 0 | 1 | 1 | 0 | 1 | 1 | 1 | 0 | 0 |
| 17 | 1 | 1 | 1 | 1 | 1 | 1 | 1 | 1 | 1 | 1 |
| 18 | 0 | 0 | 0 | 0 | 0 | 0 | 0 | 0 | 0 | 0 |
| 19 | 0 | 0 | 0 | 0 | 0 | 0 | 0 | 0 | 0 | 0 |
| 20 | 0 | 0 | 0 | 0 | 0 | 0 | 0 | 0 | 0 | 0 |
| 21 | 1 | 1 | 1 | 1 | 1 | 1 | 1 | 1 | 1 | 1 |
| 22 | 0 | 0 | 0 | 0 | 0 | 0 | 0 | 0 | 0 | 0 |
| 23 | 1 | 1 | 1 | 1 | 1 | 1 | 1 | 1 | 1 | 1 |
| 24 | 0 | 0 | 0 | 0 | 0 | 0 | 0 | 0 | 0 | 0 |
| 25 | 0 | 1 | 1 | 1 | 1 | 1 | 1 | 1 | 1 | 1 |
| 26 | 0 | 0 | 0 | 0 | 0 | 0 | 0 | 1 | 0 | 0 |
| 27 | 0 | 0 | 0 | 0 | 0 | 0 | 0 | 0 | 0 | 0 |
| 28 | 1 | 1 | 1 | 1 | 1 | 1 | 1 | 1 | 1 | 1 |
| 29 | 0 | 0 | 0 | 0 | 0 | 0 | 0 | 0 | 0 | 0 |
| 30 | 1 | 1 | 1 | 1 | 1 | 1 | 1 | 1 | 1 | 1 |
| 31 | 1 | 1 | 1 | 1 | 1 | 1 | 1 | 1 | 1 | 1 |
| 32 | 0 | 0 | 0 | 0 | 0 | 0 | 0 | 1 | 0 | 0 |

| ID | GPT_3.5_1 | GPT_3.5_2 | GPT_3.5_3 | GPT_3.5_4 | GPT_3.5_5 | GPT_3.5_6 | GPT_3.5_7 | GPT_3.5_8 | GPT_3.5_9 | GPT_3.5_10 |
|---|---|---|---|---|---|---|---|---|---|---|
| 33 | 1 | 0 | 1 | 1 | 0 | 1 | 0 | 1 | 0 | 0 |
| 34 | 1 | 1 | 1 | 1 | 1 | 1 | 1 | 1 | 1 | 1 |
| 35 | 0 | 0 | 1 | 0 | 0 | 1 | 0 | 1 | 1 | 0 |
| 36 | 1 | 1 | 0 | 1 | 1 | 1 | 0 | 0 | 1 | 1 |
| 37 | 0 | 0 | 0 | 0 | 0 | 0 | 0 | 0 | 0 | 0 |
| 38 | 0 | 0 | 0 | 0 | 0 | 0 | 0 | 0 | 0 | 0 |
| 39 | 0 | 0 | 0 | 0 | 0 | 0 | 0 | 0 | 0 | 0 |
| 40 | 1 | 1 | 1 | 1 | 1 | 1 | 1 | 1 | 1 | 1 |
| 41 | 0 | 0 | 0 | 0 | 0 | 0 | 0 | 0 | 0 | 0 |
| 42 | 0 | 0 | 0 | 0 | 0 | 0 | 0 | 0 | 0 | 0 |
| 43 | 0 | 0 | 0 | 0 | 0 | 0 | 0 | 0 | 0 | 0 |
| 44 | 1 | 1 | 1 | 1 | 1 | 1 | 1 | 1 | 1 | 1 |
| 45 | 1 | 1 | 1 | 1 | 1 | 1 | 1 | 1 | 1 | 1 |
| 46 | 1 | 1 | 1 | 1 | 1 | 1 | 1 | 1 | 1 | 1 |
| 47 | 0 | 0 | 0 | 0 | 0 | 0 | 0 | 0 | 0 | 0 |
| 48 | 0 | 0 | 0 | 0 | 0 | 0 | 0 | 0 | 0 | 0 |
| 49 | 1 | 1 | 1 | 1 | 1 | 1 | 1 | 1 | 1 | 1 |
| 50 | 0 | 0 | 0 | 0 | 0 | 0 | 0 | 0 | 0 | 0 |
| 51 | 0 | 0 | 0 | 0 | 0 | 0 | 0 | 0 | 0 | 0 |
| 52 | 0 | 0 | 0 | 0 | 0 | 0 | 0 | 0 | 0 | 0 |
| 53 | 0 | 0 | 0 | 0 | 0 | 0 | 0 | 0 | 0 | 0 |
| 54 | 0 | 0 | 0 | 0 | 0 | 0 | 0 | 0 | 0 | 0 |
| 55 | 1 | 1 | 1 | 1 | 1 | 1 | 1 | 1 | 1 | 1 |
| 56 | 1 | 1 | 1 | 1 | 1 | 1 | 1 | 1 | 1 | 1 |
| 57 | 0 | 0 | 0 | 0 | 0 | 0 | 0 | 0 | 0 | 0 |
| 58 | 1 | 1 | 1 | 1 | 1 | 1 | 1 | 1 | 1 | 1 |

| ID | GPT_3.5_1 | GPT_3.5_2 | GPT_3.5_3 | GPT_3.5_4 | GPT_3.5_5 | GPT_3.5_6 | GPT_3.5_7 | GPT_3.5_8 | GPT_3.5_9 | GPT_3.5_10 |
|---|---|---|---|---|---|---|---|---|---|---|
| 59 | 1 | 1 | 1 | 1 | 1 | 1 | 1 | 1 | 1 | 1 |
| 60 | 0 | 0 | 0 | 0 | 0 | 0 | 0 | 0 | 0 | 0 |
| 61 | 1 | 1 | 0 | 1 | 0 | 1 | 0 | 1 | 1 | 0 |
| 62 | 1 | 1 | 1 | 1 | 1 | 1 | 1 | 0 | 1 | 1 |
| 63 | 1 | 1 | 1 | 1 | 1 | 1 | 1 | 1 | 1 | 1 |
| 64 | 1 | 1 | 1 | 1 | 1 | 1 | 1 | 1 | 1 | 1 |
| 65 | 0 | 0 | 0 | 0 | 0 | 0 | 0 | 0 | 0 | 0 |
| 66 | 1 | 1 | 1 | 1 | 1 | 1 | 1 | 1 | 1 | 1 |
| 67 | 1 | 1 | 1 | 1 | 1 | 1 | 1 | 1 | 1 | 1 |
| 68 | 1 | 1 | 1 | 1 | 1 | 1 | 1 | 1 | 1 | 1 |
| 69 | 1 | 1 | 1 | 1 | 1 | 1 | 1 | 1 | 1 | 1 |
| 70 | 0 | 0 | 0 | 0 | 0 | 0 | 0 | 0 | 0 | 0 |
| 71 | 0 | 0 | 0 | 0 | 0 | 0 | 0 | 0 | 0 | 0 |
| 72 | 1 | 1 | 1 | 1 | 1 | 1 | 1 | 1 | 1 | 1 |
| 73 | 1 | 1 | 1 | 1 | 1 | 1 | 1 | 1 | 1 | 1 |
| 74 | 1 | 1 | 1 | 1 | 1 | 1 | 1 | 1 | 1 | 1 |
| 75 | 1 | 1 | 1 | 1 | 1 | 1 | 1 | 1 | 1 | 1 |
| 76 | 1 | 1 | 1 | 1 | 1 | 1 | 1 | 1 | 1 | 1 |
| 77 | 0 | 0 | 0 | 0 | 0 | 0 | 0 | 0 | 0 | 0 |
| 78 | 0 | 1 | 0 | 0 | 1 | 1 | 1 | 0 | 0 | 0 |
| 79 | 1 | 1 | 1 | 1 | 1 | 1 | 1 | 1 | 1 | 1 |
| 80 | 0 | 0 | 0 | 0 | 0 | 0 | 0 | 0 | 0 | 0 |
| 81 | 0 | 0 | 0 | 0 | 0 | 0 | 0 | 0 | 0 | 0 |
| 82 | 0 | 0 | 0 | 0 | 0 | 0 | 0 | 0 | 0 | 0 |
| 83 | 1 | 1 | 1 | 1 | 1 | 1 | 1 | 1 | 1 | 1 |
| 84 | 0 | 0 | 0 | 0 | 0 | 0 | 0 | 0 | 0 | 0 |

| ID | GPT_3.5_1 | GPT_3.5_2 | GPT_3.5_3 | GPT_3.5_4 | GPT_3.5_5 | GPT_3.5_6 | GPT_3.5_7 | GPT_3.5_8 | GPT_3.5_9 | GPT_3.5_10 |
|---|---|---|---|---|---|---|---|---|---|---|
| 85 | 0 | 0 | 0 | 0 | 0 | 0 | 0 | 0 | 0 | 0 |
| 86 | 0 | 0 | 0 | 0 | 0 | 0 | 0 | 0 | 0 | 0 |
| 87 | 0 | 0 | 0 | 0 | 0 | 0 | 0 | 0 | 0 | 0 |
| 88 | 1 | 1 | 1 | 1 | 1 | 1 | 1 | 1 | 1 | 1 |
| 89 | 0 | 0 | 0 | 0 | 0 | 0 | 0 | 0 | 0 | 0 |
| 90 | 1 | 1 | 1 | 1 | 1 | 1 | 1 | 1 | 1 | 1 |
| 91 | 1 | 1 | 1 | 1 | 1 | 1 | 1 | 1 | 1 | 1 |
| 92 | 1 | 1 | 1 | 1 | 1 | 1 | 1 | 1 | 1 | 1 |
| 93 | 1 | 1 | 1 | 1 | 1 | 1 | 1 | 1 | 1 | 1 |
| 94 | 0 | 0 | 0 | 0 | 0 | 0 | 0 | 0 | 0 | 0 |
| 95 | 1 | 1 | 0 | 1 | 0 | 1 | 1 | 1 | 1 | 0 |
| 96 | 1 | 1 | 1 | 1 | 1 | 1 | 1 | 1 | 1 | 1 |
| 97 | 1 | 1 | 1 | 1 | 1 | 1 | 1 | 1 | 1 | 1 |
| 98 | 0 | 0 | 0 | 0 | 1 | 1 | 1 | 1 | 1 | 1 |
| 99 | 1 | 0 | 0 | 0 | 0 | 0 | 0 | 0 | 0 | 0 |
| 100 | 1 | 1 | 1 | 1 | 1 | 1 | 1 | 1 | 1 | 1 |
| 101 | 1 | 0 | 1 | 0 | 1 | 1 | 1 | 1 | 1 | 1 |
| 102 | 1 | 1 | 1 | 1 | 1 | 1 | 1 | 1 | 1 | 1 |
| 103 | 0 | 0 | 0 | 0 | 0 | 0 | 0 | 0 | 0 | 0 |
| 104 | 0 | 1 | 1 | 1 | 1 | 1 | 1 | 1 | 1 | 1 |
| 105 | 1 | 1 | 1 | 1 | 1 | 1 | 1 | 1 | 1 | 1 |
| 106 | 0 | 0 | 0 | 0 | 0 | 0 | 0 | 0 | 0 | 0 |
| 107 | 1 | 1 | 1 | 1 | 1 | 1 | 1 | 1 | 1 | 1 |
| 108 | 0 | 0 | 0 | 0 | 0 | 0 | 0 | 0 | 0 | 0 |
| 109 | 1 | 1 | 1 | 1 | 1 | 1 | 1 | 1 | 1 | 1 |
| 110 | 0 | 0 | 0 | 0 | 0 | 0 | 0 | 0 | 0 | 0 |

| ID | GPT_3.5_1 | GPT_3.5_2 | GPT_3.5_3 | GPT_3.5_4 | GPT_3.5_5 | GPT_3.5_6 | GPT_3.5_7 | GPT_3.5_8 | GPT_3.5_9 | GPT_3.5_10 |
|---|---|---|---|---|---|---|---|---|---|---|
| 111 | 1 | 1 | 1 | 1 | 1 | 1 | 1 | 1 | 1 | 1 |
| 112 | 0 | 0 | 0 | 0 | 0 | 0 | 0 | 0 | 0 | 0 |
| 113 | 1 | 1 | 1 | 1 | 1 | 1 | 1 | 1 | 1 | 1 |
| 114 | 1 | 1 | 1 | 1 | 1 | 1 | 1 | 1 | 1 | 1 |
| 115 | 0 | 0 | 0 | 0 | 0 | 0 | 0 | 0 | 0 | 0 |
| 116 | 1 | 0 | 0 | 1 | 1 | 0 | 1 | 1 | 1 | 1 |
| 117 | 0 | 0 | 0 | 0 | 0 | 0 | 0 | 0 | 0 | 0 |
| 118 | 1 | 1 | 1 | 1 | 1 | 1 | 1 | 1 | 1 | 1 |
| 119 | 1 | 1 | 1 | 1 | 1 | 1 | 1 | 1 | 1 | 1 |
| 120 | 1 | 1 | 1 | 1 | 1 | 1 | 1 | 1 | 1 | 1 |
| 121 | 1 | 0 | 1 | 0 | 0 | 0 | 0 | 1 | 0 | 0 |
| 122 | 0 | 0 | 0 | 0 | 0 | 0 | 0 | 0 | 0 | 0 |
| 123 | 1 | 1 | 1 | 1 | 1 | 1 | 1 | 1 | 1 | 1 |
| 124 | 1 | 1 | 1 | 1 | 1 | 1 | 1 | 1 | 1 | 1 |
| 125 | 1 | 1 | 1 | 1 | 1 | 1 | 1 | 1 | 1 | 1 |
| 126 | 0 | 0 | 0 | 0 | 0 | 0 | 0 | 0 | 0 | 0 |
| 127 | 0 | 0 | 0 | 0 | 0 | 0 | 0 | 0 | 0 | 0 |
| 128 | 1 | 1 | 1 | 1 | 1 | 1 | 1 | 1 | 1 | 1 |
| 129 | 0 | 0 | 0 | 0 | 0 | 0 | 0 | 0 | 0 | 0 |
| 130 | 1 | 1 | 1 | 1 | 1 | 1 | 1 | 1 | 1 | 1 |
| 131 | 0 | 0 | 0 | 0 | 0 | 0 | 0 | 0 | 0 | 0 |
| 132 | 0 | 0 | 0 | 0 | 0 | 0 | 0 | 0 | 0 | 0 |
| 133 | 1 | 1 | 1 | 1 | 1 | 1 | 1 | 1 | 1 | 1 |
| 134 | 1 | 1 | 1 | 1 | 1 | 1 | 1 | 1 | 1 | 1 |
| 135 | 0 | 0 | 0 | 0 | 0 | 0 | 0 | 0 | 0 | 0 |
| 136 | 1 | 1 | 1 | 1 | 1 | 1 | 1 | 1 | 1 | 1 |

| ID | GPT_3.5_1 | GPT_3.5_2 | GPT_3.5_3 | GPT_3.5_4 | GPT_3.5_5 | GPT_3.5_6 | GPT_3.5_7 | GPT_3.5_8 | GPT_3.5_9 | GPT_3.5_10 |
|---|---|---|---|---|---|---|---|---|---|---|
| 137 | 0 | 0 | 0 | 0 | 0 | 0 | 0 | 0 | 0 | 0 |
| 138 | 0 | 0 | 0 | 0 | 0 | 0 | 0 | 0 | 0 | 0 |
| 139 | 0 | 0 | 0 | 0 | 0 | 0 | 0 | 0 | 0 | 0 |
| 140 | 1 | 1 | 1 | 1 | 1 | 1 | 1 | 1 | 1 | 1 |
| 141 | 0 | 0 | 0 | 1 | 1 | 0 | 0 | 0 | 0 | 1 |
| 142 | 1 | 1 | 1 | 1 | 1 | 1 | 1 | 1 | 1 | 1 |
| 143 | 0 | 0 | 1 | 1 | 0 | 1 | 1 | 0 | 0 | 0 |
| 144 | 1 | 1 | 1 | 0 | 1 | 0 | 0 | 0 | 1 | 0 |
| 145 | 0 | 0 | 0 | 0 | 0 | 0 | 0 | 0 | 0 | 0 |
| 146 | 0 | 0 | 0 | 0 | 0 | 0 | 0 | 0 | 0 | 0 |
| 147 | 1 | 1 | 1 | 1 | 1 | 1 | 1 | 1 | 1 | 1 |
| 148 | 0 | 0 | 0 | 0 | 0 | 0 | 0 | 0 | 0 | 0 |
| 149 | 0 | 0 | 0 | 0 | 0 | 0 | 0 | 0 | 0 | 0 |
| 150 | 0 | 0 | 0 | 0 | 0 | 0 | 0 | 0 | 0 | 0 |
| 151 | 1 | 1 | 1 | 1 | 1 | 1 | 1 | 1 | 1 | 1 |
| 152 | 0 | 0 | 1 | 0 | 0 | 0 | 0 | 0 | 0 | 0 |
| 153 | 0 | 0 | 0 | 0 | 0 | 0 | 0 | 0 | 0 | 0 |
| 154 | 0 | 0 | 0 | 0 | 0 | 0 | 0 | 0 | 0 | 0 |
| 155 | 1 | 1 | 1 | 1 | 1 | 1 | 1 | 1 | 1 | 1 |
| 156 | 1 | 0 | 0 | 0 | 0 | 0 | 1 | 0 | 0 | 0 |
| 157 | 0 | 0 | 0 | 0 | 0 | 0 | 0 | 0 | 0 | 0 |
| 158 | 0 | 0 | 0 | 1 | 1 | 0 | 1 | 0 | 0 | 0 |
| 159 | 1 | 1 | 1 | 1 | 1 | 1 | 1 | 1 | 1 | 1 |
| 160 | 1 | 1 | 1 | 0 | 0 | 0 | 0 | 1 | 0 | 0 |
| 161 | 1 | 1 | 1 | 1 | 1 | 1 | 1 | 1 | 1 | 1 |
| 162 | 1 | 1 | 1 | 1 | 1 | 1 | 1 | 1 | 1 | 1 |

| ID | GPT_3.5_1 | GPT_3.5_2 | GPT_3.5_3 | GPT_3.5_4 | GPT_3.5_5 | GPT_3.5_6 | GPT_3.5_7 | GPT_3.5_8 | GPT_3.5_9 | GPT_3.5_10 |
|---|---|---|---|---|---|---|---|---|---|---|
| 163 | 1 | 1 | 1 | 1 | 1 | 1 | 1 | 1 | 1 | 1 |
| 164 | 1 | 1 | 1 | 1 | 1 | 1 | 1 | 1 | 1 | 1 |
| 165 | 0 | 1 | 0 | 1 | 1 | 0 | 1 | 1 | 1 | 0 |
| 166 | 1 | 1 | 1 | 1 | 1 | 1 | 1 | 1 | 1 | 1 |
| 167 | 1 | 1 | 1 | 1 | 1 | 1 | 1 | 1 | 1 | 1 |
| 168 | 0 | 0 | 0 | 0 | 0 | 0 | 0 | 0 | 0 | 0 |
| 169 | 0 | 0 | 0 | 0 | 0 | 0 | 0 | 0 | 0 | 0 |
| 170 | 1 | 1 | 1 | 1 | 1 | 1 | 1 | 1 | 1 | 1 |
| 171 | 0 | 0 | 0 | 0 | 0 | 0 | 0 | 0 | 0 | 0 |
| 172 | 0 | 0 | 0 | 0 | 0 | 0 | 0 | 0 | 0 | 0 |
| 173 | 1 | 1 | 1 | 1 | 1 | 1 | 1 | 1 | 1 | 1 |
| 174 | 0 | 0 | 0 | 0 | 0 | 0 | 0 | 0 | 0 | 0 |
| 175 | 1 | 1 | 1 | 1 | 1 | 1 | 1 | 1 | 1 | 1 |
| 176 | 1 | 1 | 1 | 1 | 1 | 1 | 1 | 1 | 0 | 1 |
| 177 | 0 | 0 | 0 | 0 | 0 | 0 | 0 | 0 | 0 | 0 |
| 178 | 0 | 0 | 0 | 0 | 0 | 0 | 0 | 0 | 0 | 0 |
| 179 | 0 | 0 | 0 | 0 | 0 | 0 | 0 | 0 | 0 | 0 |
| 180 | 1 | 1 | 1 | 1 | 1 | 1 | 1 | 1 | 1 | 1 |
| 181 | 1 | 0 | 1 | 1 | 1 | 1 | 1 | 1 | 1 | 1 |
| 182 | 1 | 1 | 1 | 1 | 1 | 1 | 1 | 1 | 1 | 1 |
| 183 | 1 | 1 | 1 | 1 | 1 | 1 | 1 | 1 | 1 | 1 |
| 184 | 0 | 0 | 0 | 0 | 0 | 0 | 0 | 0 | 0 | 0 |
| 185 | 0 | 0 | 0 | 0 | 0 | 0 | 0 | 0 | 0 | 0 |
| 186 | 0 | 0 | 0 | 0 | 0 | 0 | 0 | 0 | 0 | 0 |
| 187 | 0 | 0 | 0 | 0 | 0 | 0 | 0 | 0 | 0 | 0 |
| 188 | 0 | 0 | 0 | 0 | 0 | 0 | 0 | 0 | 0 | 0 |

| ID | GPT_3.5_1 | GPT_3.5_2 | GPT_3.5_3 | GPT_3.5_4 | GPT_3.5_5 | GPT_3.5_6 | GPT_3.5_7 | GPT_3.5_8 | GPT_3.5_9 | GPT_3.5_10 |
|---|---|---|---|---|---|---|---|---|---|---|
| 189 | 0 | 0 | 0 | 0 | 0 | 0 | 0 | 0 | 0 | 0 |
| 190 | 0 | 0 | 0 | 0 | 0 | 0 | 0 | 0 | 0 | 0 |
| 191 | 1 | 1 | 1 | 1 | 1 | 1 | 1 | 1 | 1 | 1 |
| 192 | 0 | 0 | 0 | 0 | 0 | 0 | 0 | 0 | 0 | 0 |
| 193 | 0 | 0 | 0 | 0 | 0 | 0 | 0 | 0 | 0 | 0 |
| 194 | 1 | 1 | 1 | 1 | 1 | 1 | 1 | 1 | 1 | 1 |
| 195 | 1 | 1 | 1 | 1 | 1 | 1 | 1 | 1 | 1 | 1 |
| 196 | 0 | 0 | 1 | 0 | 0 | 1 | 0 | 1 | 0 | 0 |
| 197 | 1 | 1 | 1 | 1 | 1 | 1 | 1 | 1 | 1 | 1 |
| 198 | 0 | 0 | 0 | 0 | 0 | 0 | 0 | 0 | 0 | 0 |
| 199 | 1 | 1 | 1 | 1 | 1 | 1 | 1 | 1 | 1 | 1 |
| 200 | 0 | 0 | 0 | 0 | 0 | 0 | 0 | 0 | 0 | 0 |

| ID | GPT_4_1 | GPT_4_2 | GPT_4_3 | GPT_4_4 | GPT_4_5 | GPT_4_6 | GPT_4_7 | GPT_4_8 | GPT_4_9 | GPT_4_10 |
|---|---|---|---|---|---|---|---|---|---|---|
| 1 | 1 | 1 | 1 | 1 | 1 | 1 | 1 | 1 | 1 | 1 |
| 2 | 0 | 1 | 0 | 1 | 0 | 1 | 1 | 0 | 1 | 1 |
| 3 | 1 | 1 | 1 | 1 | 1 | 1 | 1 | 1 | 1 | 1 |
| 4 | 1 | 1 | 1 | 1 | 1 | 1 | 1 | 1 | 1 | 1 |
| 5 | 1 | 1 | 1 | 1 | 1 | 1 | 1 | 1 | 1 | 1 |
| 6 | 1 | 1 | 1 | 1 | 1 | 1 | 1 | 1 | 1 | 1 |
| 7 | 1 | 1 | 1 | 1 | 1 | 1 | 1 | 1 | 1 | 1 |
| 8 | 0 | 1 | 0 | 0 | 0 | 0 | 0 | 1 | 0 | 0 |
| 9 | 1 | 1 | 1 | 1 | 1 | 1 | 1 | 1 | 1 | 1 |
| 10 | 1 | 1 | 1 | 1 | 1 | 1 | 1 | 1 | 1 | 1 |

| ID | GPT_4_1 | GPT_4_2 | GPT_4_3 | GPT_4_4 | GPT_4_5 | GPT_4_6 | GPT_4_7 | GPT_4_8 | GPT_4_9 | GPT_4_10 |
|---|---|---|---|---|---|---|---|---|---|---|
| 11 | 0 | 0 | 0 | 0 | 0 | 0 | 0 | 0 | 0 | 0 |
| 12 | 0 | 0 | 0 | 0 | 0 | 0 | 0 | 0 | 0 | 0 |
| 13 | 0 | 0 | 0 | 0 | 0 | 0 | 0 | 0 | 0 | 0 |
| 14 | 1 | 1 | 1 | 1 | 1 | 1 | 1 | 1 | 1 | 1 |
| 15 | 1 | 1 | 1 | 1 | 1 | 1 | 1 | 1 | 1 | 1 |
| 16 | 1 | 1 | 0 | 1 | 1 | 0 | 0 | 1 | 1 | 0 |
| 17 | 1 | 1 | 1 | 1 | 1 | 1 | 1 | 1 | 1 | 1 |
| 18 | 1 | 1 | 1 | 1 | 1 | 1 | 1 | 1 | 1 | 1 |
| 19 | 0 | 0 | 0 | 0 | 0 | 0 | 0 | 0 | 0 | 0 |
| 20 | 0 | 0 | 0 | 1 | 0 | 0 | 1 | 0 | 0 | 0 |
| 21 | 1 | 1 | 1 | 1 | 1 | 1 | 1 | 1 | 1 | 1 |
| 22 | 0 | 0 | 0 | 0 | 0 | 0 | 0 | 0 | 0 | 0 |
| 23 | 1 | 1 | 1 | 1 | 1 | 1 | 1 | 1 | 1 | 1 |
| 24 | 0 | 0 | 0 | 0 | 0 | 0 | 0 | 0 | 0 | 0 |
| 25 | 1 | 1 | 1 | 1 | 1 | 1 | 1 | 1 | 1 | 1 |
| 26 | 0 | 0 | 0 | 0 | 0 | 0 | 0 | 0 | 0 | 0 |
| 27 | 0 | 0 | 0 | 0 | 0 | 0 | 0 | 0 | 1 | 0 |
| 28 | 1 | 1 | 1 | 1 | 1 | 1 | 1 | 1 | 1 | 1 |
| 29 | 0 | 0 | 0 | 0 | 0 | 0 | 0 | 0 | 0 | 0 |
| 30 | 1 | 1 | 1 | 1 | 1 | 1 | 1 | 1 | 1 | 1 |
| 31 | 1 | 0 | 1 | 1 | 1 | 0 | 1 | 1 | 0 | 0 |
| 32 | 1 | 1 | 1 | 1 | 1 | 1 | 1 | 1 | 1 | 1 |
| 33 | 1 | 1 | 1 | 1 | 1 | 1 | 1 | 1 | 1 | 1 |
| 34 | 1 | 1 | 1 | 1 | 1 | 1 | 1 | 1 | 1 | 1 |
| 35 | 1 | 1 | 1 | 1 | 1 | 1 | 1 | 1 | 1 | 1 |
| 36 | 1 | 1 | 1 | 1 | 1 | 1 | 1 | 1 | 1 | 1 |
| 37 | 0 | 0 | 0 | 0 | 0 | 0 | 0 | 0 | 0 | 0 |

| ID | GPT_4_1 | GPT_4_2 | GPT_4_3 | GPT_4_4 | GPT_4_5 | GPT_4_6 | GPT_4_7 | GPT_4_8 | GPT_4_9 | GPT_4_10 |
|---|---|---|---|---|---|---|---|---|---|---|
| 38 | 0 | 0 | 0 | 0 | 0 | 0 | 0 | 0 | 0 | 0 |
| 39 | 0 | 0 | 0 | 0 | 0 | 0 | 0 | 0 | 0 | 0 |
| 40 | 1 | 1 | 1 | 1 | 1 | 1 | 1 | 1 | 1 | 1 |
| 41 | 1 | 1 | 1 | 1 | 1 | 1 | 1 | 1 | 1 | 1 |
| 42 | 0 | 0 | 0 | 0 | 0 | 0 | 0 | 0 | 0 | 0 |
| 43 | 0 | 0 | 0 | 0 | 0 | 0 | 0 | 0 | 0 | 0 |
| 44 | 1 | 1 | 1 | 1 | 1 | 1 | 1 | 1 | 1 | 1 |
| 45 | 1 | 1 | 1 | 1 | 1 | 1 | 1 | 1 | 1 | 1 |
| 46 | 1 | 1 | 1 | 1 | 1 | 1 | 1 | 1 | 1 | 1 |
| 47 | 1 | 0 | 1 | 0 | 1 | 0 | 0 | 0 | 0 | 1 |
| 48 | 1 | 0 | 1 | 0 | 1 | 0 | 0 | 1 | 0 | 1 |
| 49 | 1 | 1 | 1 | 1 | 1 | 1 | 1 | 1 | 1 | 1 |
| 50 | 0 | 1 | 1 | 1 | 0 | 0 | 1 | 1 | 1 | 0 |
| 51 | 0 | 0 | 0 | 0 | 0 | 0 | 0 | 0 | 0 | 0 |
| 52 | 1 | 1 | 1 | 1 | 1 | 1 | 1 | 1 | 1 | 1 |
| 53 | 1 | 1 | 1 | 1 | 1 | 1 | 1 | 1 | 1 | 1 |
| 54 | 1 | 1 | 1 | 1 | 1 | 1 | 1 | 1 | 1 | 1 |
| 55 | 1 | 1 | 1 | 1 | 1 | 1 | 1 | 1 | 1 | 1 |
| 56 | 1 | 1 | 1 | 1 | 1 | 1 | 1 | 1 | 1 | 1 |
| 57 | 1 | 1 | 1 | 1 | 1 | 1 | 1 | 1 | 1 | 1 |
| 58 | 1 | 1 | 1 | 1 | 1 | 1 | 1 | 1 | 1 | 1 |
| 59 | 1 | 1 | 1 | 1 | 1 | 1 | 1 | 1 | 1 | 1 |
| 60 | 1 | 1 | 1 | 1 | 1 | 1 | 1 | 1 | 1 | 1 |
| 61 | 1 | 1 | 1 | 1 | 1 | 1 | 1 | 1 | 1 | 1 |
| 62 | 0 | 0 | 0 | 0 | 0 | 0 | 0 | 0 | 0 | 0 |
| 63 | 1 | 1 | 1 | 1 | 1 | 1 | 1 | 1 | 1 | 1 |
| 64 | 1 | 1 | 1 | 1 | 1 | 1 | 1 | 1 | 1 | 1 |

| ID | GPT_4_1 | GPT_4_2 | GPT_4_3 | GPT_4_4 | GPT_4_5 | GPT_4_6 | GPT_4_7 | GPT_4_8 | GPT_4_9 | GPT_4_10 |
|---|---|---|---|---|---|---|---|---|---|---|
| 65 | 1 | 1 | 1 | 1 | 1 | 1 | 1 | 1 | 1 | 1 |
| 66 | 1 | 1 | 1 | 1 | 1 | 1 | 1 | 1 | 1 | 1 |
| 67 | 1 | 1 | 1 | 1 | 1 | 1 | 1 | 1 | 1 | 1 |
| 68 | 0 | 0 | 0 | 0 | 0 | 1 | 0 | 0 | 0 | 0 |
| 69 | 1 | 1 | 1 | 1 | 1 | 1 | 1 | 1 | 1 | 1 |
| 70 | 1 | 1 | 1 | 1 | 1 | 1 | 1 | 1 | 1 | 1 |
| 71 | 1 | 1 | 1 | 1 | 1 | 1 | 1 | 1 | 1 | 1 |
| 72 | 1 | 1 | 1 | 1 | 1 | 1 | 1 | 1 | 1 | 1 |
| 73 | 1 | 1 | 1 | 1 | 1 | 1 | 1 | 1 | 1 | 1 |
| 74 | 1 | 1 | 1 | 1 | 1 | 1 | 1 | 1 | 1 | 1 |
| 75 | 1 | 1 | 1 | 1 | 1 | 1 | 1 | 1 | 1 | 1 |
| 76 | 1 | 1 | 1 | 1 | 1 | 1 | 1 | 1 | 1 | 1 |
| 77 | 1 | 1 | 1 | 1 | 1 | 1 | 1 | 1 | 1 | 1 |
| 78 | 1 | 1 | 1 | 1 | 1 | 1 | 1 | 1 | 1 | 1 |
| 79 | 1 | 1 | 1 | 1 | 1 | 1 | 1 | 1 | 1 | 1 |
| 80 | 0 | 0 | 0 | 0 | 0 | 0 | 0 | 0 | 0 | 0 |
| 81 | 0 | 0 | 0 | 0 | 0 | 0 | 0 | 0 | 0 | 0 |
| 82 | 1 | 1 | 1 | 1 | 1 | 1 | 1 | 1 | 1 | 1 |
| 83 | 1 | 1 | 1 | 1 | 1 | 1 | 1 | 1 | 1 | 1 |
| 84 | 1 | 1 | 1 | 1 | 1 | 1 | 1 | 1 | 1 | 1 |
| 85 | 0 | 0 | 0 | 0 | 0 | 0 | 0 | 0 | 0 | 0 |
| 86 | 0 | 0 | 0 | 0 | 0 | 0 | 0 | 0 | 0 | 0 |
| 87 | 1 | 1 | 1 | 1 | 1 | 1 | 1 | 1 | 1 | 1 |
| 88 | 0 | 0 | 0 | 0 | 0 | 0 | 0 | 0 | 0 | 0 |
| 89 | 1 | 1 | 1 | 1 | 1 | 1 | 1 | 1 | 1 | 1 |
| 90 | 1 | 1 | 1 | 1 | 1 | 1 | 1 | 1 | 1 | 1 |
| 91 | 1 | 1 | 1 | 1 | 1 | 1 | 1 | 1 | 1 | 1 |

| ID | GPT_4_1 | GPT_4_2 | GPT_4_3 | GPT_4_4 | GPT_4_5 | GPT_4_6 | GPT_4_7 | GPT_4_8 | GPT_4_9 | GPT_4_10 |
|---|---|---|---|---|---|---|---|---|---|---|
| 92 | 1 | 1 | 1 | 1 | 1 | 1 | 1 | 1 | 1 | 1 |
| 93 | 1 | 1 | 1 | 1 | 1 | 1 | 1 | 1 | 1 | 1 |
| 94 | 1 | 1 | 1 | 1 | 1 | 1 | 1 | 1 | 1 | 1 |
| 95 | 1 | 1 | 1 | 1 | 1 | 1 | 1 | 1 | 1 | 1 |
| 96 | 1 | 1 | 1 | 1 | 1 | 1 | 1 | 1 | 1 | 1 |
| 97 | 1 | 1 | 1 | 1 | 1 | 1 | 1 | 1 | 1 | 1 |
| 98 | 0 | 0 | 0 | 0 | 0 | 0 | 0 | 0 | 0 | 0 |
| 99 | 1 | 1 | 1 | 1 | 1 | 1 | 1 | 1 | 0 | 1 |
| 100 | 1 | 1 | 1 | 1 | 1 | 1 | 1 | 1 | 1 | 1 |
| 101 | 1 | 1 | 1 | 1 | 1 | 1 | 1 | 1 | 1 | 1 |
| 102 | 1 | 1 | 1 | 1 | 1 | 1 | 1 | 1 | 1 | 1 |
| 103 | 1 | 1 | 1 | 1 | 1 | 1 | 1 | 1 | 1 | 1 |
| 104 | 0 | 0 | 1 | 0 | 1 | 0 | 0 | 0 | 0 | 0 |
| 105 | 1 | 1 | 1 | 1 | 1 | 1 | 1 | 1 | 1 | 1 |
| 106 | 0 | 0 | 0 | 0 | 0 | 0 | 0 | 0 | 0 | 0 |
| 107 | 1 | 1 | 1 | 1 | 1 | 1 | 1 | 1 | 1 | 1 |
| 108 | 1 | 1 | 1 | 1 | 1 | 1 | 1 | 1 | 1 | 1 |
| 109 | 0 | 0 | 0 | 0 | 0 | 0 | 0 | 0 | 0 | 0 |
| 110 | 0 | 0 | 0 | 0 | 0 | 0 | 0 | 0 | 0 | 0 |
| 111 | 1 | 1 | 1 | 1 | 1 | 1 | 1 | 1 | 1 | 1 |
| 112 | 1 | 1 | 1 | 1 | 1 | 1 | 1 | 1 | 1 | 1 |
| 113 | 1 | 1 | 1 | 1 | 1 | 1 | 1 | 1 | 1 | 1 |
| 114 | 1 | 1 | 1 | 1 | 1 | 1 | 1 | 1 | 1 | 1 |
| 115 | 0 | 0 | 0 | 0 | 0 | 0 | 0 | 0 | 0 | 0 |
| 116 | 1 | 1 | 1 | 1 | 1 | 1 | 1 | 1 | 1 | 1 |
| 117 | 0 | 0 | 0 | 0 | 0 | 0 | 0 | 0 | 0 | 0 |
| 118 | 1 | 1 | 1 | 1 | 1 | 1 | 1 | 1 | 1 | 1 |

| ID | GPT_4_1 | GPT_4_2 | GPT_4_3 | GPT_4_4 | GPT_4_5 | GPT_4_6 | GPT_4_7 | GPT_4_8 | GPT_4_9 | GPT_4_10 |
|---|---|---|---|---|---|---|---|---|---|---|
| 119 | 1 | 1 | 1 | 1 | 1 | 1 | 1 | 1 | 1 | 1 |
| 120 | 1 | 1 | 1 | 1 | 1 | 1 | 1 | 1 | 1 | 1 |
| 121 | 0 | 0 | 0 | 0 | 0 | 0 | 0 | 0 | 0 | 0 |
| 122 | 1 | 1 | 1 | 1 | 1 | 1 | 1 | 1 | 1 | 1 |
| 123 | 1 | 1 | 1 | 1 | 1 | 1 | 1 | 1 | 1 | 1 |
| 124 | 1 | 1 | 1 | 1 | 1 | 1 | 1 | 1 | 1 | 1 |
| 125 | 1 | 1 | 1 | 1 | 1 | 1 | 1 | 1 | 1 | 1 |
| 126 | 0 | 0 | 0 | 0 | 0 | 0 | 0 | 0 | 0 | 0 |
| 127 | 0 | 0 | 0 | 0 | 0 | 0 | 0 | 0 | 0 | 0 |
| 128 | 1 | 1 | 1 | 1 | 1 | 1 | 1 | 1 | 1 | 1 |
| 129 | 0 | 0 | 0 | 1 | 0 | 1 | 0 | 0 | 0 | 0 |
| 130 | 1 | 1 | 1 | 1 | 1 | 1 | 1 | 1 | 1 | 1 |
| 131 | 1 | 1 | 1 | 1 | 1 | 1 | 1 | 1 | 1 | 1 |
| 132 | 1 | 1 | 1 | 1 | 1 | 1 | 1 | 1 | 1 | 1 |
| 133 | 1 | 1 | 1 | 1 | 1 | 1 | 1 | 1 | 1 | 1 |
| 134 | 1 | 1 | 1 | 1 | 1 | 1 | 1 | 1 | 1 | 1 |
| 135 | 0 | 0 | 0 | 0 | 0 | 0 | 0 | 0 | 0 | 0 |
| 136 | 1 | 1 | 1 | 1 | 1 | 1 | 1 | 1 | 1 | 1 |
| 137 | 1 | 1 | 1 | 1 | 1 | 1 | 1 | 1 | 1 | 1 |
| 138 | 1 | 1 | 1 | 1 | 1 | 1 | 1 | 1 | 1 | 1 |
| 139 | 0 | 0 | 0 | 0 | 0 | 0 | 0 | 0 | 0 | 0 |
| 140 | 1 | 1 | 1 | 1 | 1 | 1 | 1 | 1 | 1 | 1 |
| 141 | 1 | 1 | 1 | 1 | 1 | 1 | 1 | 1 | 1 | 1 |
| 142 | 1 | 1 | 1 | 1 | 1 | 1 | 1 | 1 | 1 | 1 |
| 143 | 1 | 1 | 1 | 1 | 1 | 1 | 1 | 1 | 1 | 1 |
| 144 | 1 | 1 | 1 | 1 | 1 | 1 | 1 | 1 | 1 | 1 |
| 145 | 1 | 1 | 1 | 1 | 1 | 1 | 1 | 1 | 1 | 1 |

| ID | GPT_4_1 | GPT_4_2 | GPT_4_3 | GPT_4_4 | GPT_4_5 | GPT_4_6 | GPT_4_7 | GPT_4_8 | GPT_4_9 | GPT_4_10 |
|---|---|---|---|---|---|---|---|---|---|---|
| 146 | 1 | 1 | 1 | 1 | 1 | 1 | 1 | 1 | 1 | 1 |
| 147 | 1 | 1 | 1 | 1 | 1 | 1 | 1 | 1 | 1 | 1 |
| 148 | 1 | 1 | 1 | 1 | 1 | 1 | 1 | 1 | 1 | 1 |
| 149 | 1 | 1 | 1 | 1 | 1 | 1 | 1 | 1 | 1 | 1 |
| 150 | 1 | 1 | 1 | 1 | 1 | 1 | 1 | 1 | 1 | 1 |
| 151 | 0 | 0 | 0 | 0 | 0 | 0 | 0 | 0 | 0 | 0 |
| 152 | 0 | 0 | 0 | 0 | 0 | 0 | 0 | 0 | 0 | 0 |
| 153 | 0 | 0 | 0 | 0 | 0 | 0 | 0 | 0 | 0 | 0 |
| 154 | 1 | 1 | 0 | 0 | 1 | 0 | 0 | 0 | 1 | 0 |
| 155 | 1 | 1 | 1 | 1 | 1 | 1 | 1 | 1 | 1 | 1 |
| 156 | 1 | 1 | 1 | 1 | 1 | 1 | 1 | 1 | 1 | 1 |
| 157 | 1 | 1 | 1 | 1 | 1 | 1 | 1 | 1 | 1 | 1 |
| 158 | 1 | 1 | 1 | 1 | 1 | 1 | 1 | 1 | 1 | 1 |
| 159 | 0 | 0 | 0 | 0 | 0 | 0 | 0 | 0 | 0 | 0 |
| 160 | 1 | 1 | 1 | 1 | 1 | 1 | 1 | 1 | 1 | 1 |
| 161 | 1 | 1 | 1 | 1 | 1 | 1 | 1 | 1 | 1 | 1 |
| 162 | 1 | 1 | 1 | 1 | 1 | 1 | 1 | 1 | 1 | 1 |
| 163 | 1 | 1 | 1 | 1 | 1 | 1 | 1 | 1 | 1 | 1 |
| 164 | 1 | 1 | 1 | 1 | 1 | 1 | 1 | 1 | 1 | 1 |
| 165 | 1 | 1 | 1 | 1 | 1 | 1 | 1 | 1 | 1 | 1 |
| 166 | 1 | 1 | 1 | 1 | 1 | 1 | 1 | 1 | 1 | 1 |
| 167 | 1 | 1 | 1 | 1 | 1 | 1 | 1 | 1 | 1 | 1 |
| 168 | 1 | 1 | 1 | 1 | 1 | 1 | 1 | 1 | 1 | 1 |
| 169 | 1 | 1 | 1 | 1 | 1 | 1 | 1 | 1 | 1 | 1 |
| 170 | 1 | 1 | 1 | 1 | 1 | 1 | 1 | 1 | 1 | 1 |
| 171 | 0 | 0 | 0 | 0 | 0 | 0 | 0 | 0 | 0 | 0 |
| 172 | 0 | 0 | 0 | 0 | 0 | 0 | 0 | 0 | 0 | 0 |

| ID | GPT_4_1 | GPT_4_2 | GPT_4_3 | GPT_4_4 | GPT_4_5 | GPT_4_6 | GPT_4_7 | GPT_4_8 | GPT_4_9 | GPT_4_10 |
|---|---|---|---|---|---|---|---|---|---|---|
| 173 | 0 | 0 | 0 | 0 | 0 | 0 | 0 | 0 | 0 | 0 |
| 174 | 1 | 1 | 1 | 1 | 1 | 1 | 1 | 1 | 1 | 1 |
| 175 | 1 | 1 | 1 | 1 | 1 | 1 | 1 | 1 | 1 | 1 |
| 176 | 1 | 1 | 1 | 1 | 1 | 1 | 1 | 1 | 1 | 1 |
| 177 | 1 | 1 | 1 | 1 | 1 | 1 | 1 | 1 | 1 | 1 |
| 178 | 1 | 1 | 1 | 1 | 1 | 1 | 1 | 1 | 0 | 1 |
| 179 | 0 | 0 | 0 | 0 | 0 | 0 | 0 | 0 | 0 | 0 |
| 180 | 0 | 0 | 0 | 0 | 0 | 0 | 0 | 0 | 0 | 0 |
| 181 | 1 | 1 | 1 | 1 | 1 | 1 | 1 | 1 | 1 | 1 |
| 182 | 1 | 1 | 1 | 1 | 1 | 1 | 1 | 1 | 1 | 1 |
| 183 | 1 | 1 | 1 | 1 | 1 | 1 | 1 | 1 | 1 | 1 |
| 184 | 1 | 1 | 1 | 1 | 1 | 1 | 1 | 1 | 1 | 1 |
| 185 | 0 | 0 | 0 | 0 | 0 | 0 | 0 | 0 | 0 | 0 |
| 186 | 0 | 0 | 0 | 0 | 0 | 0 | 0 | 0 | 0 | 0 |
| 187 | 0 | 0 | 0 | 0 | 0 | 0 | 0 | 0 | 0 | 0 |
| 188 | 1 | 1 | 1 | 1 | 1 | 1 | 1 | 1 | 1 | 1 |
| 189 | 1 | 1 | 1 | 1 | 1 | 1 | 1 | 1 | 1 | 1 |
| 190 | 1 | 1 | 1 | 1 | 1 | 1 | 1 | 1 | 1 | 1 |
| 191 | 1 | 1 | 1 | 1 | 1 | 1 | 1 | 1 | 1 | 1 |
| 192 | 0 | 0 | 0 | 0 | 0 | 0 | 0 | 0 | 0 | 0 |
| 193 | 1 | 1 | 1 | 1 | 1 | 1 | 1 | 1 | 1 | 1 |
| 194 | 0 | 0 | 0 | 0 | 0 | 0 | 0 | 0 | 0 | 0 |
| 195 | 1 | 1 | 1 | 1 | 1 | 1 | 1 | 1 | 1 | 1 |
| 196 | 0 | 0 | 0 | 1 | 0 | 1 | 0 | 0 | 1 | 0 |
| 197 | 1 | 1 | 1 | 1 | 1 | 1 | 1 | 1 | 1 | 1 |
| 198 | 0 | 0 | 0 | 0 | 0 | 0 | 0 | 0 | 0 | 0 |
| 199 | 1 | 1 | 1 | 1 | 1 | 1 | 1 | 1 | 1 | 1 |

| ID | GPT_4_1 | GPT_4_2 | GPT_4_3 | GPT_4_4 | GPT_4_5 | GPT_4_6 | GPT_4_7 | GPT_4_8 | GPT_4_9 | GPT_4_10 |
|---|---|---|---|---|---|---|---|---|---|---|
| 200 | 0 | 0 | 0 | 0 | 0 | 0 | 0 | 0 | 0 | 0 |

| ID | UroBot_3.5_1 | UroBot_3.5_2 | UroBot_3.5_3 | UroBot_3.5_4 | UroBot_3.5_5 | UroBot_3.5_6 | UroBot_3.5_7 | UroBot_3.5_8 | UroBot_3.5_9 | UroBot_3.5_10 |
|---|---|---|---|---|---|---|---|---|---|---|
| 1 | 1 | 1 | 1 | 1 | 1 | 1 | 1 | 1 | 1 | 1 |
| 2 | 1 | 1 | 1 | 1 | 1 | 1 | 1 | 1 | 1 | 1 |
| 3 | 1 | 1 | 1 | 1 | 1 | 1 | 1 | 1 | 1 | 1 |
| 4 | 1 | 1 | 1 | 1 | 1 | 1 | 1 | 1 | 1 | 1 |
| 5 | 1 | 1 | 1 | 1 | 1 | 1 | 1 | 1 | 1 | 1 |
| 6 | 1 | 1 | 1 | 1 | 1 | 1 | 1 | 1 | 1 | 1 |
| 7 | 0 | 0 | 0 | 0 | 0 | 0 | 0 | 0 | 0 | 0 |
| 8 | 1 | 1 | 1 | 1 | 1 | 1 | 1 | 1 | 1 | 1 |
| 9 | 1 | 1 | 1 | 1 | 1 | 1 | 1 | 1 | 1 | 1 |
| 10 | 1 | 1 | 1 | 1 | 1 | 1 | 1 | 1 | 1 | 1 |
| 11 | 0 | 0 | 0 | 0 | 0 | 0 | 0 | 0 | 0 | 0 |
| 12 | 0 | 0 | 0 | 0 | 0 | 0 | 0 | 0 | 0 | 0 |
| 13 | 0 | 0 | 0 | 0 | 0 | 0 | 0 | 0 | 0 | 0 |
| 14 | 1 | 1 | 1 | 1 | 1 | 1 | 1 | 1 | 1 | 1 |
| 15 | 1 | 1 | 1 | 1 | 1 | 1 | 1 | 1 | 1 | 1 |
| 16 | 1 | 1 | 1 | 1 | 1 | 1 | 1 | 1 | 1 | 1 |
| 17 | 1 | 1 | 1 | 1 | 1 | 1 | 1 | 1 | 1 | 1 |
| 18 | 1 | 1 | 1 | 1 | 1 | 1 | 1 | 1 | 1 | 1 |
| 19 | 0 | 0 | 0 | 0 | 0 | 0 | 0 | 0 | 0 | 0 |
| 20 | 1 | 1 | 1 | 1 | 1 | 1 | 1 | 1 | 1 | 1 |
| 21 | 1 | 1 | 1 | 1 | 1 | 1 | 1 | 1 | 1 | 1 |
| 22 | 0 | 0 | 0 | 0 | 0 | 0 | 0 | 0 | 0 | 0 |

| ID | UroBot_3.5_1 | UroBot_3.5_2 | UroBot_3.5_3 | UroBot_3.5_4 | UroBot_3.5_5 | UroBot_3.5_6 | UroBot_3.5_7 | UroBot_3.5_8 | UroBot_3.5_9 | UroBot_3.5_10 |
|---|---|---|---|---|---|---|---|---|---|---|
| 23 | 1 | 1 | 1 | 1 | 1 | 1 | 1 | 1 | 1 | 1 |
| 24 | 0 | 0 | 0 | 0 | 0 | 0 | 0 | 0 | 0 | 0 |
| 25 | 1 | 1 | 1 | 1 | 1 | 1 | 1 | 1 | 1 | 1 |
| 26 | 0 | 0 | 0 | 0 | 0 | 0 | 1 | 0 | 1 | 0 |
| 27 | 0 | 0 | 0 | 0 | 0 | 0 | 0 | 0 | 0 | 0 |
| 28 | 1 | 1 | 1 | 1 | 1 | 1 | 1 | 1 | 1 | 1 |
| 29 | 0 | 0 | 0 | 0 | 0 | 0 | 0 | 0 | 0 | 0 |
| 30 | 1 | 1 | 1 | 1 | 1 | 1 | 1 | 1 | 1 | 1 |
| 31 | 1 | 1 | 1 | 1 | 1 | 1 | 1 | 1 | 1 | 1 |
| 32 | 1 | 1 | 1 | 1 | 1 | 1 | 1 | 1 | 1 | 1 |
| 33 | 1 | 1 | 1 | 1 | 1 | 1 | 1 | 1 | 1 | 1 |
| 34 | 1 | 1 | 1 | 1 | 1 | 1 | 1 | 1 | 1 | 1 |
| 35 | 1 | 1 | 1 | 1 | 1 | 1 | 1 | 1 | 1 | 1 |
| 36 | 1 | 1 | 1 | 1 | 1 | 1 | 1 | 1 | 1 | 1 |
| 37 | 0 | 0 | 0 | 0 | 0 | 0 | 0 | 0 | 0 | 0 |
| 38 | 0 | 0 | 0 | 0 | 0 | 0 | 0 | 0 | 0 | 0 |
| 39 | 1 | 1 | 1 | 1 | 1 | 1 | 1 | 1 | 1 | 1 |
| 40 | 1 | 1 | 1 | 1 | 1 | 1 | 1 | 1 | 1 | 1 |
| 41 | 0 | 0 | 0 | 0 | 0 | 0 | 0 | 0 | 0 | 0 |
| 42 | 1 | 1 | 1 | 1 | 1 | 1 | 1 | 1 | 1 | 1 |
| 43 | 0 | 0 | 0 | 0 | 0 | 0 | 0 | 0 | 0 | 0 |
| 44 | 1 | 1 | 1 | 1 | 1 | 1 | 1 | 1 | 1 | 1 |
| 45 | 1 | 1 | 1 | 1 | 1 | 1 | 1 | 1 | 1 | 1 |
| 46 | 1 | 1 | 1 | 1 | 1 | 1 | 1 | 1 | 1 | 1 |
| 47 | 0 | 1 | 0 | 1 | 1 | 1 | 1 | 1 | 0 | 0 |
| 48 | 0 | 0 | 0 | 0 | 0 | 0 | 0 | 0 | 0 | 0 |
| 49 | 1 | 1 | 1 | 1 | 1 | 1 | 1 | 1 | 1 | 1 |

| ID | UroBot_3.5_1 | UroBot_3.5_2 | UroBot_3.5_3 | UroBot_3.5_4 | UroBot_3.5_5 | UroBot_3.5_6 | UroBot_3.5_7 | UroBot_3.5_8 | UroBot_3.5_9 | UroBot_3.5_10 |
|---|---|---|---|---|---|---|---|---|---|---|
| 50 | 1 | 1 | 1 | 1 | 1 | 1 | 1 | 1 | 1 | 1 |
| 51 | 0 | 0 | 0 | 0 | 0 | 0 | 0 | 0 | 0 | 0 |
| 52 | 1 | 1 | 1 | 1 | 1 | 1 | 1 | 1 | 1 | 1 |
| 53 | 1 | 1 | 1 | 1 | 1 | 1 | 0 | 1 | 1 | 1 |
| 54 | 0 | 0 | 0 | 0 | 0 | 0 | 0 | 0 | 0 | 0 |
| 55 | 1 | 1 | 1 | 1 | 1 | 1 | 1 | 1 | 1 | 1 |
| 56 | 1 | 1 | 1 | 1 | 1 | 1 | 1 | 1 | 1 | 1 |
| 57 | 0 | 0 | 0 | 0 | 0 | 0 | 0 | 0 | 0 | 0 |
| 58 | 1 | 1 | 1 | 1 | 1 | 1 | 1 | 1 | 1 | 1 |
| 59 | 1 | 1 | 1 | 1 | 1 | 1 | 1 | 1 | 1 | 1 |
| 60 | 1 | 1 | 0 | 1 | 0 | 0 | 0 | 0 | 1 | 0 |
| 61 | 0 | 0 | 0 | 0 | 0 | 0 | 0 | 0 | 0 | 0 |
| 62 | 1 | 1 | 1 | 1 | 1 | 1 | 1 | 1 | 1 | 1 |
| 63 | 1 | 1 | 1 | 1 | 1 | 1 | 1 | 1 | 1 | 1 |
| 64 | 1 | 1 | 1 | 1 | 1 | 1 | 1 | 1 | 1 | 1 |
| 65 | 1 | 1 | 1 | 1 | 1 | 1 | 1 | 1 | 1 | 1 |
| 66 | 0 | 0 | 0 | 0 | 0 | 0 | 0 | 0 | 0 | 0 |
| 67 | 1 | 1 | 1 | 1 | 1 | 1 | 1 | 1 | 1 | 1 |
| 68 | 0 | 0 | 1 | 1 | 1 | 1 | 1 | 1 | 1 | 1 |
| 69 | 1 | 1 | 1 | 1 | 1 | 1 | 1 | 1 | 1 | 1 |
| 70 | 1 | 1 | 1 | 1 | 1 | 1 | 1 | 1 | 1 | 1 |
| 71 | 0 | 0 | 0 | 0 | 0 | 0 | 0 | 0 | 0 | 0 |
| 72 | 0 | 0 | 1 | 0 | 0 | 0 | 0 | 0 | 0 | 0 |
| 73 | 1 | 1 | 1 | 1 | 1 | 1 | 1 | 1 | 1 | 1 |
| 74 | 1 | 1 | 1 | 1 | 1 | 1 | 1 | 1 | 1 | 1 |
| 75 | 1 | 1 | 1 | 1 | 1 | 1 | 1 | 1 | 1 | 1 |
| 76 | 1 | 1 | 1 | 1 | 1 | 1 | 1 | 1 | 1 | 1 |

| ID | UroBot_3.5_1 | UroBot_3.5_2 | UroBot_3.5_3 | UroBot_3.5_4 | UroBot_3.5_5 | UroBot_3.5_6 | UroBot_3.5_7 | UroBot_3.5_8 | UroBot_3.5_9 | UroBot_3.5_10 |
|---|---|---|---|---|---|---|---|---|---|---|
| 77 | 0 | 0 | 1 | 0 | 0 | 0 | 0 | 1 | 0 | 0 |
| 78 | 1 | 1 | 1 | 1 | 1 | 1 | 1 | 1 | 1 | 1 |
| 79 | 1 | 1 | 1 | 1 | 1 | 1 | 1 | 1 | 1 | 1 |
| 80 | 0 | 0 | 0 | 0 | 0 | 0 | 0 | 0 | 0 | 0 |
| 81 | 1 | 1 | 1 | 1 | 0 | 1 | 0 | 1 | 1 | 1 |
| 82 | 0 | 0 | 0 | 0 | 0 | 0 | 0 | 0 | 0 | 0 |
| 83 | 1 | 1 | 1 | 1 | 1 | 1 | 1 | 1 | 1 | 1 |
| 84 | 0 | 0 | 0 | 0 | 0 | 0 | 0 | 0 | 0 | 0 |
| 85 | 0 | 0 | 0 | 0 | 0 | 0 | 0 | 0 | 0 | 0 |
| 86 | 0 | 1 | 1 | 1 | 1 | 1 | 1 | 0 | 1 | 0 |
| 87 | 0 | 0 | 0 | 0 | 0 | 0 | 0 | 0 | 0 | 0 |
| 88 | 1 | 1 | 1 | 1 | 1 | 1 | 1 | 1 | 1 | 1 |
| 89 | 1 | 1 | 1 | 1 | 1 | 1 | 1 | 1 | 1 | 1 |
| 90 | 1 | 1 | 1 | 1 | 1 | 1 | 1 | 1 | 1 | 1 |
| 91 | 1 | 1 | 1 | 1 | 1 | 1 | 1 | 1 | 1 | 1 |
| 92 | 1 | 1 | 1 | 1 | 1 | 1 | 1 | 1 | 1 | 1 |
| 93 | 1 | 1 | 1 | 1 | 1 | 1 | 1 | 1 | 1 | 1 |
| 94 | 1 | 1 | 1 | 1 | 1 | 1 | 1 | 1 | 1 | 1 |
| 95 | 1 | 1 | 1 | 1 | 1 | 1 | 1 | 1 | 1 | 1 |
| 96 | 1 | 1 | 1 | 1 | 1 | 1 | 1 | 1 | 1 | 1 |
| 97 | 1 | 1 | 1 | 1 | 1 | 1 | 1 | 1 | 1 | 1 |
| 98 | 1 | 1 | 1 | 1 | 1 | 1 | 1 | 1 | 1 | 1 |
| 99 | 1 | 1 | 1 | 1 | 1 | 1 | 1 | 1 | 1 | 0 |
| 100 | 1 | 1 | 1 | 1 | 1 | 1 | 1 | 1 | 1 | 1 |
| 101 | 1 | 1 | 1 | 1 | 1 | 1 | 1 | 1 | 1 | 1 |
| 102 | 1 | 1 | 1 | 1 | 1 | 1 | 1 | 1 | 1 | 1 |
| 103 | 0 | 0 | 0 | 0 | 0 | 0 | 0 | 0 | 0 | 0 |

| ID | UroBot_3.5_1 | UroBot_3.5_2 | UroBot_3.5_3 | UroBot_3.5_4 | UroBot_3.5_5 | UroBot_3.5_6 | UroBot_3.5_7 | UroBot_3.5_8 | UroBot_3.5_9 | UroBot_3.5_10 |
|---|---|---|---|---|---|---|---|---|---|---|
| 104 | 1 | 1 | 1 | 1 | 1 | 1 | 1 | 1 | 1 | 1 |
| 105 | 1 | 1 | 1 | 1 | 1 | 1 | 1 | 1 | 1 | 1 |
| 106 | 1 | 1 | 1 | 1 | 1 | 1 | 1 | 1 | 1 | 1 |
| 107 | 1 | 1 | 1 | 1 | 1 | 1 | 1 | 1 | 1 | 1 |
| 108 | 0 | 0 | 0 | 0 | 0 | 0 | 0 | 0 | 0 | 0 |
| 109 | 1 | 1 | 1 | 1 | 1 | 1 | 1 | 1 | 1 | 1 |
| 110 | 0 | 0 | 0 | 0 | 1 | 0 | 0 | 1 | 0 | 0 |
| 111 | 1 | 1 | 1 | 1 | 1 | 1 | 1 | 1 | 1 | 1 |
| 112 | 1 | 1 | 1 | 1 | 1 | 1 | 1 | 1 | 1 | 1 |
| 113 | 1 | 1 | 1 | 1 | 1 | 1 | 1 | 1 | 1 | 1 |
| 114 | 1 | 1 | 1 | 1 | 1 | 1 | 1 | 1 | 1 | 1 |
| 115 | 0 | 0 | 0 | 0 | 0 | 0 | 0 | 0 | 0 | 0 |
| 116 | 1 | 1 | 1 | 1 | 1 | 1 | 1 | 1 | 1 | 1 |
| 117 | 1 | 1 | 1 | 1 | 1 | 1 | 1 | 1 | 1 | 1 |
| 118 | 1 | 1 | 1 | 1 | 1 | 1 | 1 | 1 | 1 | 1 |
| 119 | 1 | 1 | 0 | 1 | 1 | 1 | 1 | 1 | 1 | 1 |
| 120 | 0 | 1 | 0 | 1 | 0 | 0 | 0 | 0 | 0 | 1 |
| 121 | 1 | 1 | 1 | 1 | 1 | 1 | 1 | 1 | 1 | 1 |
| 122 | 0 | 0 | 0 | 0 | 0 | 0 | 0 | 0 | 0 | 0 |
| 123 | 1 | 1 | 1 | 1 | 1 | 1 | 1 | 1 | 1 | 1 |
| 124 | 1 | 1 | 1 | 1 | 1 | 1 | 1 | 1 | 1 | 1 |
| 125 | 1 | 1 | 1 | 1 | 1 | 1 | 1 | 1 | 1 | 1 |
| 126 | 1 | 1 | 1 | 1 | 1 | 1 | 1 | 1 | 1 | 1 |
| 127 | 0 | 0 | 0 | 0 | 0 | 0 | 0 | 0 | 0 | 0 |
| 128 | 1 | 1 | 1 | 1 | 1 | 1 | 1 | 1 | 1 | 1 |
| 129 | 1 | 1 | 1 | 1 | 1 | 1 | 1 | 1 | 1 | 1 |
| 130 | 1 | 1 | 1 | 1 | 1 | 1 | 1 | 1 | 1 | 1 |

| ID | UroBot_3.5_1 | UroBot_3.5_2 | UroBot_3.5_3 | UroBot_3.5_4 | UroBot_3.5_5 | UroBot_3.5_6 | UroBot_3.5_7 | UroBot_3.5_8 | UroBot_3.5_9 | UroBot_3.5_10 |
|---|---|---|---|---|---|---|---|---|---|---|
| 131 | 1 | 1 | 1 | 1 | 1 | 1 | 1 | 1 | 1 | 1 |
| 132 | 0 | 0 | 0 | 0 | 0 | 0 | 0 | 0 | 0 | 0 |
| 133 | 1 | 1 | 1 | 1 | 1 | 1 | 1 | 1 | 1 | 1 |
| 134 | 1 | 1 | 1 | 1 | 1 | 1 | 1 | 1 | 1 | 1 |
| 135 | 1 | 1 | 1 | 1 | 1 | 1 | 1 | 1 | 1 | 1 |
| 136 | 1 | 1 | 1 | 1 | 1 | 1 | 1 | 1 | 1 | 1 |
| 137 | 0 | 0 | 0 | 0 | 0 | 0 | 0 | 0 | 0 | 0 |
| 138 | 1 | 1 | 1 | 1 | 1 | 1 | 1 | 1 | 1 | 1 |
| 139 | 0 | 0 | 0 | 0 | 0 | 0 | 0 | 0 | 0 | 0 |
| 140 | 1 | 1 | 1 | 1 | 1 | 1 | 1 | 1 | 0 | 1 |
| 141 | 1 | 0 | 0 | 0 | 0 | 0 | 0 | 0 | 0 | 0 |
| 142 | 1 | 1 | 1 | 1 | 1 | 1 | 1 | 1 | 1 | 1 |
| 143 | 1 | 1 | 1 | 1 | 1 | 1 | 1 | 1 | 1 | 1 |
| 144 | 1 | 1 | 1 | 1 | 1 | 1 | 1 | 1 | 1 | 1 |
| 145 | 1 | 1 | 1 | 1 | 1 | 1 | 1 | 1 | 1 | 1 |
| 146 | 0 | 0 | 0 | 0 | 0 | 0 | 0 | 0 | 0 | 0 |
| 147 | 1 | 1 | 1 | 1 | 1 | 1 | 1 | 1 | 1 | 1 |
| 148 | 1 | 1 | 1 | 1 | 1 | 1 | 1 | 1 | 1 | 1 |
| 149 | 1 | 1 | 1 | 1 | 1 | 1 | 1 | 1 | 1 | 1 |
| 150 | 0 | 0 | 0 | 0 | 0 | 0 | 0 | 0 | 0 | 0 |
| 151 | 1 | 1 | 1 | 1 | 1 | 1 | 1 | 1 | 1 | 1 |
| 152 | 1 | 0 | 0 | 0 | 0 | 0 | 0 | 0 | 0 | 0 |
| 153 | 0 | 0 | 0 | 0 | 0 | 0 | 0 | 0 | 0 | 0 |
| 154 | 0 | 1 | 1 | 0 | 1 | 1 | 0 | 1 | 0 | 0 |
| 155 | 0 | 0 | 1 | 0 | 0 | 0 | 0 | 0 | 0 | 0 |
| 156 | 1 | 1 | 1 | 1 | 1 | 1 | 1 | 1 | 1 | 1 |
| 157 | 1 | 1 | 1 | 1 | 1 | 1 | 1 | 1 | 1 | 1 |

| ID | UroBot_3.5_1 | UroBot_3.5_2 | UroBot_3.5_3 | UroBot_3.5_4 | UroBot_3.5_5 | UroBot_3.5_6 | UroBot_3.5_7 | UroBot_3.5_8 | UroBot_3.5_9 | UroBot_3.5_10 |
|---|---|---|---|---|---|---|---|---|---|---|
| 158 | 1 | 1 | 0 | 1 | 0 | 1 | 1 | 1 | 1 | 1 |
| 159 | 1 | 1 | 1 | 1 | 1 | 1 | 1 | 1 | 1 | 1 |
| 160 | 1 | 1 | 1 | 1 | 1 | 1 | 1 | 1 | 1 | 1 |
| 161 | 1 | 1 | 1 | 1 | 1 | 1 | 1 | 1 | 1 | 1 |
| 162 | 1 | 1 | 1 | 1 | 1 | 1 | 1 | 1 | 1 | 1 |
| 163 | 1 | 1 | 1 | 1 | 1 | 1 | 1 | 1 | 1 | 1 |
| 164 | 1 | 1 | 1 | 1 | 1 | 1 | 1 | 1 | 1 | 1 |
| 165 | 0 | 0 | 0 | 0 | 0 | 0 | 0 | 0 | 0 | 0 |
| 166 | 1 | 1 | 1 | 1 | 1 | 1 | 1 | 1 | 1 | 1 |
| 167 | 1 | 1 | 1 | 1 | 1 | 1 | 1 | 1 | 1 | 1 |
| 168 | 1 | 1 | 1 | 1 | 1 | 1 | 1 | 1 | 1 | 1 |
| 169 | 1 | 1 | 1 | 1 | 1 | 1 | 1 | 1 | 1 | 1 |
| 170 | 1 | 1 | 1 | 1 | 1 | 1 | 1 | 1 | 1 | 1 |
| 171 | 1 | 1 | 1 | 1 | 1 | 1 | 1 | 1 | 1 | 1 |
| 172 | 0 | 0 | 0 | 0 | 0 | 0 | 0 | 0 | 0 | 0 |
| 173 | 1 | 1 | 1 | 1 | 1 | 1 | 1 | 1 | 1 | 1 |
| 174 | 1 | 1 | 0 | 1 | 0 | 0 | 0 | 0 | 0 | 0 |
| 175 | 1 | 1 | 1 | 1 | 1 | 1 | 1 | 1 | 1 | 1 |
| 176 | 1 | 1 | 1 | 1 | 1 | 1 | 1 | 1 | 1 | 1 |
| 177 | 1 | 0 | 0 | 0 | 1 | 0 | 1 | 0 | 1 | 0 |
| 178 | 0 | 0 | 0 | 0 | 0 | 0 | 0 | 0 | 0 | 0 |
| 179 | 1 | 1 | 0 | 0 | 0 | 0 | 0 | 1 | 0 | 1 |
| 180 | 1 | 1 | 1 | 1 | 1 | 1 | 1 | 1 | 1 | 1 |
| 181 | 0 | 0 | 0 | 0 | 0 | 0 | 0 | 0 | 0 | 0 |
| 182 | 1 | 1 | 1 | 1 | 1 | 1 | 1 | 1 | 1 | 1 |
| 183 | 1 | 1 | 1 | 1 | 1 | 1 | 1 | 1 | 1 | 1 |
| 184 | 1 | 1 | 1 | 1 | 1 | 1 | 1 | 1 | 1 | 1 |

| ID | UroBot_3.5_1 | UroBot_3.5_2 | UroBot_3.5_3 | UroBot_3.5_4 | UroBot_3.5_5 | UroBot_3.5_6 | UroBot_3.5_7 | UroBot_3.5_8 | UroBot_3.5_9 | UroBot_3.5_10 |
|---|---|---|---|---|---|---|---|---|---|---|
| 185 | 1 | 1 | 1 | 1 | 1 | 1 | 1 | 1 | 1 | 1 |
| 186 | 1 | 1 | 1 | 1 | 1 | 1 | 1 | 1 | 1 | 1 |
| 187 | 1 | 1 | 1 | 1 | 1 | 1 | 1 | 1 | 1 | 1 |
| 188 | 1 | 1 | 1 | 0 | 1 | 1 | 1 | 1 | 1 | 1 |
| 189 | 0 | 0 | 0 | 0 | 0 | 0 | 0 | 0 | 0 | 0 |
| 190 | 1 | 1 | 1 | 1 | 1 | 1 | 1 | 1 | 1 | 1 |
| 191 | 1 | 1 | 1 | 1 | 1 | 1 | 1 | 1 | 1 | 1 |
| 192 | 0 | 0 | 0 | 0 | 0 | 0 | 0 | 0 | 0 | 0 |
| 193 | 1 | 1 | 1 | 1 | 1 | 1 | 1 | 1 | 1 | 1 |
| 194 | 1 | 1 | 1 | 1 | 1 | 1 | 1 | 1 | 1 | 1 |
| 195 | 1 | 1 | 1 | 1 | 1 | 1 | 1 | 1 | 1 | 1 |
| 196 | 1 | 1 | 1 | 1 | 1 | 1 | 1 | 1 | 1 | 1 |
| 197 | 1 | 1 | 1 | 1 | 1 | 1 | 1 | 1 | 1 | 1 |
| 198 | 0 | 0 | 0 | 0 | 0 | 0 | 0 | 0 | 0 | 0 |
| 199 | 1 | 1 | 1 | 1 | 1 | 1 | 1 | 1 | 1 | 1 |
| 200 | 0 | 0 | 0 | 0 | 0 | 0 | 0 | 0 | 0 | 0 |

| ID | UroBot_4_1 | UroBot_4_2 | UroBot_4_3 | UroBot_4_4 | UroBot_4_5 | UroBot_4_6 | UroBot_4_7 | UroBot_4_8 | UroBot_4_9 | UroBot_4_10 |
|---|---|---|---|---|---|---|---|---|---|---|
| 1 | 1 | 1 | 1 | 1 | 1 | 1 | 1 | 1 | 1 | 1 |
| 2 | 1 | 1 | 1 | 1 | 1 | 1 | 1 | 1 | 1 | 1 |
| 3 | 1 | 1 | 1 | 1 | 1 | 1 | 1 | 1 | 1 | 1 |
| 4 | 1 | 1 | 1 | 1 | 1 | 1 | 1 | 1 | 1 | 1 |
| 5 | 1 | 1 | 1 | 1 | 1 | 1 | 1 | 1 | 1 | 1 |

| ID | UroBot_4_1 | UroBot_4_2 | UroBot_4_3 | UroBot_4_4 | UroBot_4_5 | UroBot_4_6 | UroBot_4_7 | UroBot_4_8 | UroBot_4_9 | UroBot_4_10 |
|---|---|---|---|---|---|---|---|---|---|---|
| 6 | 1 | 1 | 1 | 1 | 1 | 1 | 1 | 1 | 1 | 1 |
| 7 | 1 | 1 | 1 | 1 | 1 | 1 | 1 | 1 | 1 | 1 |
| 8 | 1 | 1 | 1 | 1 | 1 | 1 | 1 | 1 | 1 | 1 |
| 9 | 1 | 1 | 1 | 1 | 1 | 1 | 1 | 1 | 1 | 1 |
| 10 | 1 | 1 | 1 | 1 | 1 | 1 | 1 | 1 | 1 | 1 |
| 11 | 1 | 1 | 1 | 1 | 1 | 1 | 1 | 1 | 1 | 1 |
| 12 | 0 | 0 | 0 | 0 | 0 | 0 | 0 | 0 | 0 | 0 |
| 13 | 0 | 0 | 0 | 0 | 0 | 0 | 0 | 0 | 0 | 0 |
| 14 | 1 | 1 | 1 | 1 | 1 | 1 | 1 | 1 | 1 | 1 |
| 15 | 1 | 1 | 1 | 1 | 1 | 1 | 1 | 1 | 1 | 1 |
| 16 | 1 | 1 | 1 | 1 | 1 | 1 | 1 | 1 | 1 | 1 |
| 17 | 1 | 1 | 1 | 1 | 1 | 1 | 1 | 1 | 1 | 1 |
| 18 | 1 | 1 | 1 | 1 | 1 | 1 | 1 | 1 | 1 | 1 |
| 19 | 1 | 1 | 1 | 1 | 1 | 1 | 1 | 1 | 1 | 1 |
| 20 | 1 | 1 | 1 | 1 | 1 | 1 | 1 | 1 | 1 | 1 |
| 21 | 1 | 1 | 1 | 1 | 1 | 1 | 1 | 1 | 1 | 1 |
| 22 | 0 | 0 | 0 | 0 | 0 | 0 | 0 | 0 | 0 | 0 |
| 23 | 1 | 1 | 1 | 1 | 1 | 1 | 1 | 1 | 1 | 1 |
| 24 | 0 | 0 | 0 | 0 | 0 | 0 | 0 | 0 | 0 | 0 |
| 25 | 1 | 1 | 1 | 1 | 1 | 1 | 1 | 1 | 1 | 1 |
| 26 | 1 | 1 | 1 | 1 | 1 | 1 | 1 | 1 | 1 | 1 |
| 27 | 1 | 1 | 1 | 1 | 1 | 1 | 1 | 1 | 1 | 1 |
| 28 | 1 | 1 | 1 | 1 | 1 | 1 | 1 | 1 | 1 | 1 |
| 29 | 0 | 0 | 0 | 0 | 1 | 0 | 0 | 0 | 1 | 0 |
| 30 | 1 | 1 | 1 | 1 | 1 | 1 | 1 | 1 | 1 | 1 |
| 31 | 1 | 1 | 1 | 1 | 1 | 1 | 1 | 1 | 1 | 1 |

| ID | UroBot_4_1 | UroBot_4_2 | UroBot_4_3 | UroBot_4_4 | UroBot_4_5 | UroBot_4_6 | UroBot_4_7 | UroBot_4_8 | UroBot_4_9 | UroBot_4_10 |
|---|---|---|---|---|---|---|---|---|---|---|
| 32 | 1 | 1 | 1 | 1 | 1 | 1 | 1 | 1 | 1 | 1 |
| 33 | 1 | 1 | 1 | 1 | 1 | 1 | 1 | 1 | 1 | 1 |
| 34 | 1 | 1 | 1 | 1 | 1 | 1 | 1 | 1 | 1 | 1 |
| 35 | 1 | 1 | 1 | 1 | 1 | 1 | 1 | 1 | 1 | 1 |
| 36 | 1 | 1 | 1 | 1 | 1 | 1 | 1 | 1 | 1 | 1 |
| 37 | 0 | 0 | 0 | 0 | 0 | 0 | 0 | 0 | 0 | 0 |
| 38 | 0 | 0 | 0 | 0 | 0 | 0 | 0 | 0 | 0 | 0 |
| 39 | 1 | 1 | 1 | 1 | 1 | 1 | 1 | 1 | 1 | 1 |
| 40 | 1 | 1 | 1 | 1 | 1 | 1 | 1 | 1 | 1 | 1 |
| 41 | 0 | 0 | 0 | 0 | 0 | 0 | 0 | 0 | 0 | 0 |
| 42 | 1 | 1 | 1 | 1 | 1 | 1 | 1 | 1 | 1 | 1 |
| 43 | 1 | 1 | 1 | 1 | 1 | 1 | 1 | 1 | 1 | 1 |
| 44 | 1 | 1 | 1 | 1 | 1 | 1 | 1 | 1 | 1 | 1 |
| 45 | 1 | 1 | 1 | 1 | 1 | 1 | 1 | 1 | 1 | 1 |
| 46 | 1 | 1 | 1 | 1 | 1 | 1 | 1 | 1 | 1 | 1 |
| 47 | 1 | 1 | 1 | 1 | 1 | 1 | 1 | 1 | 1 | 1 |
| 48 | 1 | 1 | 1 | 1 | 1 | 1 | 1 | 1 | 1 | 1 |
| 49 | 1 | 1 | 1 | 1 | 1 | 1 | 1 | 1 | 1 | 1 |
| 50 | 1 | 1 | 1 | 1 | 1 | 1 | 0 | 1 | 0 | 0 |
| 51 | 0 | 0 | 1 | 1 | 1 | 1 | 1 | 1 | 0 | 1 |
| 52 | 1 | 1 | 1 | 1 | 1 | 1 | 1 | 1 | 1 | 1 |
| 53 | 0 | 1 | 0 | 1 | 0 | 1 | 0 | 0 | 0 | 0 |
| 54 | 1 | 1 | 1 | 1 | 1 | 1 | 1 | 1 | 1 | 1 |
| 55 | 1 | 1 | 1 | 1 | 1 | 1 | 1 | 1 | 1 | 1 |
| 56 | 1 | 1 | 1 | 1 | 1 | 1 | 1 | 1 | 1 | 1 |
| 57 | 1 | 1 | 1 | 1 | 1 | 1 | 1 | 1 | 1 | 1 |

| ID | UroBot_4_1 | UroBot_4_2 | UroBot_4_3 | UroBot_4_4 | UroBot_4_5 | UroBot_4_6 | UroBot_4_7 | UroBot_4_8 | UroBot_4_9 | UroBot_4_10 |
|---|---|---|---|---|---|---|---|---|---|---|
| 58 | 1 | 1 | 1 | 1 | 1 | 1 | 1 | 1 | 1 | 1 |
| 59 | 1 | 1 | 1 | 1 | 1 | 1 | 1 | 1 | 1 | 1 |
| 60 | 1 | 1 | 1 | 1 | 1 | 1 | 1 | 1 | 1 | 1 |
| 61 | 0 | 0 | 0 | 0 | 0 | 0 | 0 | 0 | 0 | 0 |
| 62 | 1 | 1 | 1 | 1 | 1 | 1 | 1 | 1 | 1 | 1 |
| 63 | 1 | 1 | 1 | 1 | 1 | 1 | 1 | 1 | 1 | 1 |
| 64 | 1 | 1 | 1 | 1 | 1 | 1 | 1 | 1 | 1 | 1 |
| 65 | 1 | 1 | 1 | 1 | 1 | 1 | 1 | 1 | 1 | 1 |
| 66 | 1 | 1 | 1 | 1 | 1 | 1 | 1 | 1 | 1 | 1 |
| 67 | 1 | 1 | 1 | 1 | 1 | 1 | 1 | 1 | 1 | 1 |
| 68 | 0 | 0 | 0 | 0 | 0 | 0 | 0 | 0 | 0 | 0 |
| 69 | 1 | 1 | 1 | 1 | 1 | 1 | 1 | 1 | 1 | 1 |
| 70 | 1 | 1 | 1 | 1 | 1 | 1 | 1 | 1 | 1 | 1 |
| 71 | 1 | 1 | 1 | 1 | 1 | 1 | 1 | 1 | 1 | 1 |
| 72 | 1 | 1 | 1 | 1 | 1 | 1 | 1 | 1 | 1 | 1 |
| 73 | 1 | 1 | 1 | 1 | 1 | 1 | 1 | 1 | 1 | 1 |
| 74 | 1 | 1 | 1 | 1 | 1 | 1 | 1 | 1 | 1 | 1 |
| 75 | 1 | 1 | 1 | 1 | 1 | 1 | 1 | 1 | 1 | 1 |
| 76 | 1 | 1 | 1 | 1 | 1 | 1 | 1 | 1 | 1 | 1 |
| 77 | 1 | 1 | 1 | 0 | 1 | 1 | 1 | 0 | 1 | 0 |
| 78 | 1 | 1 | 1 | 1 | 1 | 1 | 1 | 1 | 1 | 1 |
| 79 | 1 | 1 | 1 | 1 | 1 | 1 | 1 | 1 | 1 | 1 |
| 80 | 0 | 0 | 0 | 0 | 0 | 0 | 0 | 0 | 0 | 0 |
| 81 | 1 | 1 | 1 | 1 | 1 | 1 | 1 | 1 | 1 | 1 |
| 82 | 0 | 0 | 0 | 0 | 0 | 0 | 0 | 0 | 0 | 0 |
| 83 | 1 | 1 | 1 | 1 | 1 | 1 | 1 | 1 | 1 | 1 |

| ID | UroBot_4_1 | UroBot_4_2 | UroBot_4_3 | UroBot_4_4 | UroBot_4_5 | UroBot_4_6 | UroBot_4_7 | UroBot_4_8 | UroBot_4_9 | UroBot_4_10 |
|----|-----------|-----------|-----------|-----------|-----------|-----------|-----------|-----------|-----------|------------|
| 84 | 0 | 0 | 0 | 0 | 0 | 0 | 0 | 0 | 0 | 0 |
| 85 | 0 | 0 | 0 | 0 | 0 | 0 | 0 | 0 | 0 | 0 |
| 86 | 1 | 1 | 1 | 1 | 1 | 1 | 1 | 1 | 1 | 1 |
| 87 | 1 | 1 | 1 | 1 | 1 | 1 | 1 | 1 | 1 | 1 |
| 88 | 1 | 1 | 1 | 1 | 1 | 1 | 1 | 1 | 1 | 1 |
| 89 | 1 | 1 | 1 | 1 | 1 | 1 | 1 | 1 | 1 | 1 |
| 90 | 1 | 1 | 1 | 1 | 1 | 1 | 1 | 1 | 1 | 1 |
| 91 | 1 | 1 | 1 | 1 | 1 | 1 | 1 | 1 | 1 | 1 |
| 92 | 1 | 1 | 1 | 1 | 1 | 1 | 1 | 1 | 1 | 1 |
| 93 | 1 | 1 | 1 | 1 | 1 | 1 | 1 | 1 | 1 | 1 |
| 94 | 1 | 1 | 1 | 1 | 1 | 1 | 1 | 1 | 1 | 1 |
| 95 | 1 | 1 | 1 | 1 | 1 | 1 | 1 | 1 | 1 | 1 |
| 96 | 1 | 1 | 1 | 1 | 1 | 1 | 1 | 1 | 1 | 1 |
| 97 | 1 | 1 | 1 | 1 | 1 | 1 | 1 | 1 | 1 | 1 |
| 98 | 1 | 1 | 1 | 1 | 1 | 1 | 1 | 1 | 1 | 1 |
| 99 | 1 | 1 | 1 | 1 | 1 | 1 | 1 | 1 | 1 | 1 |
| 100 | 1 | 1 | 1 | 1 | 1 | 1 | 1 | 1 | 1 | 1 |
| 101 | 1 | 1 | 1 | 1 | 1 | 1 | 1 | 1 | 1 | 1 |
| 102 | 1 | 1 | 1 | 1 | 1 | 1 | 1 | 1 | 1 | 1 |
| 103 | 1 | 1 | 1 | 1 | 1 | 1 | 1 | 1 | 1 | 1 |
| 104 | 0 | 0 | 0 | 0 | 0 | 0 | 0 | 0 | 0 | 0 |
| 105 | 1 | 1 | 1 | 1 | 1 | 1 | 1 | 1 | 1 | 1 |
| 106 | 1 | 1 | 1 | 1 | 1 | 1 | 1 | 1 | 1 | 1 |
| 107 | 1 | 1 | 1 | 1 | 1 | 1 | 1 | 1 | 1 | 1 |
| 108 | 1 | 1 | 1 | 1 | 1 | 1 | 1 | 1 | 1 | 1 |
| 109 | 1 | 1 | 1 | 1 | 1 | 1 | 1 | 1 | 1 | 1 |

| ID | UroBot_4_1 | UroBot_4_2 | UroBot_4_3 | UroBot_4_4 | UroBot_4_5 | UroBot_4_6 | UroBot_4_7 | UroBot_4_8 | UroBot_4_9 | UroBot_4_10 |
|---|---|---|---|---|---|---|---|---|---|---|
| 110 | 1 | 1 | 1 | 1 | 1 | 1 | 1 | 1 | 1 | 1 |
| 111 | 1 | 1 | 1 | 1 | 1 | 1 | 1 | 1 | 1 | 1 |
| 112 | 1 | 1 | 1 | 1 | 1 | 1 | 1 | 1 | 1 | 1 |
| 113 | 1 | 1 | 1 | 1 | 1 | 1 | 1 | 1 | 1 | 1 |
| 114 | 1 | 1 | 1 | 1 | 1 | 1 | 1 | 1 | 1 | 1 |
| 115 | 0 | 0 | 0 | 0 | 0 | 0 | 0 | 0 | 0 | 0 |
| 116 | 1 | 1 | 1 | 1 | 1 | 1 | 1 | 1 | 1 | 1 |
| 117 | 1 | 1 | 1 | 1 | 0 | 1 | 1 | 1 | 1 | 1 |
| 118 | 1 | 1 | 1 | 1 | 1 | 1 | 1 | 1 | 1 | 1 |
| 119 | 1 | 1 | 1 | 1 | 1 | 1 | 1 | 1 | 1 | 1 |
| 120 | 1 | 1 | 1 | 1 | 1 | 1 | 1 | 1 | 1 | 1 |
| 121 | 1 | 1 | 1 | 1 | 1 | 1 | 1 | 1 | 1 | 1 |
| 122 | 1 | 1 | 1 | 1 | 1 | 1 | 1 | 1 | 1 | 1 |
| 123 | 1 | 1 | 1 | 1 | 1 | 1 | 1 | 1 | 1 | 1 |
| 124 | 1 | 1 | 1 | 1 | 1 | 1 | 1 | 1 | 1 | 1 |
| 125 | 1 | 1 | 1 | 1 | 1 | 1 | 1 | 1 | 1 | 1 |
| 126 | 1 | 1 | 1 | 1 | 1 | 1 | 1 | 1 | 1 | 1 |
| 127 | 1 | 1 | 1 | 1 | 1 | 1 | 1 | 1 | 1 | 1 |
| 128 | 1 | 1 | 1 | 1 | 1 | 1 | 1 | 1 | 1 | 1 |
| 129 | 1 | 1 | 1 | 1 | 1 | 1 | 1 | 1 | 1 | 1 |
| 130 | 1 | 1 | 1 | 1 | 1 | 1 | 1 | 1 | 1 | 1 |
| 131 | 1 | 1 | 1 | 1 | 1 | 1 | 1 | 1 | 1 | 1 |
| 132 | 1 | 1 | 1 | 1 | 1 | 1 | 1 | 1 | 1 | 1 |
| 133 | 1 | 1 | 1 | 1 | 1 | 1 | 1 | 1 | 1 | 1 |
| 134 | 1 | 1 | 1 | 1 | 1 | 1 | 1 | 1 | 1 | 1 |
| 135 | 1 | 1 | 1 | 1 | 1 | 1 | 1 | 1 | 1 | 1 |

| ID | UroBot_4_1 | UroBot_4_2 | UroBot_4_3 | UroBot_4_4 | UroBot_4_5 | UroBot_4_6 | UroBot_4_7 | UroBot_4_8 | UroBot_4_9 | UroBot_4_10 |
|---|---|---|---|---|---|---|---|---|---|---|
| 136 | 1 | 1 | 1 | 1 | 1 | 1 | 1 | 1 | 1 | 1 |
| 137 | 1 | 1 | 1 | 1 | 1 | 1 | 1 | 1 | 1 | 1 |
| 138 | 1 | 1 | 1 | 1 | 1 | 1 | 1 | 1 | 1 | 1 |
| 139 | 0 | 0 | 0 | 0 | 0 | 0 | 0 | 0 | 0 | 0 |
| 140 | 1 | 1 | 1 | 1 | 1 | 1 | 1 | 1 | 1 | 1 |
| 141 | 1 | 1 | 1 | 1 | 1 | 1 | 1 | 1 | 1 | 1 |
| 142 | 1 | 1 | 0 | 0 | 0 | 0 | 0 | 0 | 0 | 1 |
| 143 | 1 | 1 | 1 | 1 | 1 | 1 | 1 | 1 | 1 | 1 |
| 144 | 1 | 1 | 1 | 1 | 1 | 1 | 1 | 1 | 1 | 1 |
| 145 | 1 | 1 | 1 | 1 | 1 | 1 | 1 | 1 | 1 | 1 |
| 146 | 1 | 1 | 1 | 1 | 1 | 1 | 1 | 1 | 1 | 1 |
| 147 | 1 | 1 | 1 | 1 | 1 | 1 | 1 | 1 | 1 | 1 |
| 148 | 1 | 1 | 1 | 1 | 1 | 1 | 1 | 1 | 1 | 1 |
| 149 | 1 | 1 | 1 | 1 | 1 | 1 | 1 | 1 | 1 | 1 |
| 150 | 1 | 1 | 1 | 1 | 1 | 1 | 1 | 1 | 1 | 1 |
| 151 | 1 | 1 | 1 | 1 | 1 | 1 | 1 | 1 | 1 | 1 |
| 152 | 0 | 0 | 0 | 0 | 0 | 0 | 0 | 0 | 0 | 0 |
| 153 | 0 | 0 | 0 | 0 | 0 | 0 | 0 | 0 | 0 | 0 |
| 154 | 1 | 1 | 1 | 1 | 1 | 1 | 1 | 1 | 1 | 1 |
| 155 | 1 | 1 | 1 | 1 | 1 | 1 | 1 | 1 | 1 | 1 |
| 156 | 1 | 1 | 1 | 1 | 1 | 1 | 1 | 1 | 1 | 1 |
| 157 | 1 | 1 | 1 | 1 | 1 | 1 | 1 | 1 | 1 | 1 |
| 158 | 1 | 1 | 1 | 1 | 1 | 1 | 1 | 1 | 1 | 1 |
| 159 | 1 | 1 | 1 | 1 | 1 | 1 | 1 | 1 | 1 | 1 |
| 160 | 1 | 1 | 1 | 1 | 1 | 1 | 1 | 1 | 1 | 1 |
| 161 | 1 | 1 | 1 | 1 | 1 | 1 | 1 | 1 | 1 | 1 |

| ID | UroBot_4_1 | UroBot_4_2 | UroBot_4_3 | UroBot_4_4 | UroBot_4_5 | UroBot_4_6 | UroBot_4_7 | UroBot_4_8 | UroBot_4_9 | UroBot_4_10 |
|---|---|---|---|---|---|---|---|---|---|---|
| 162 | 1 | 1 | 1 | 1 | 1 | 1 | 1 | 1 | 1 | 1 |
| 163 | 1 | 1 | 1 | 1 | 1 | 1 | 1 | 1 | 1 | 1 |
| 164 | 1 | 1 | 1 | 1 | 1 | 1 | 1 | 1 | 1 | 1 |
| 165 | 1 | 1 | 1 | 1 | 1 | 1 | 1 | 1 | 1 | 1 |
| 166 | 1 | 1 | 1 | 1 | 1 | 1 | 1 | 1 | 1 | 1 |
| 167 | 1 | 1 | 1 | 1 | 1 | 1 | 1 | 1 | 1 | 1 |
| 168 | 1 | 1 | 1 | 1 | 1 | 1 | 1 | 1 | 1 | 1 |
| 169 | 1 | 1 | 1 | 1 | 1 | 1 | 1 | 1 | 1 | 1 |
| 170 | 1 | 1 | 1 | 1 | 1 | 1 | 1 | 1 | 1 | 1 |
| 171 | 1 | 1 | 1 | 1 | 1 | 1 | 1 | 1 | 1 | 1 |
| 172 | 0 | 0 | 0 | 0 | 0 | 0 | 0 | 0 | 0 | 0 |
| 173 | 1 | 1 | 1 | 1 | 1 | 1 | 1 | 1 | 1 | 1 |
| 174 | 1 | 1 | 1 | 1 | 1 | 1 | 1 | 1 | 1 | 1 |
| 175 | 1 | 1 | 1 | 1 | 1 | 1 | 1 | 1 | 1 | 1 |
| 176 | 1 | 1 | 1 | 1 | 1 | 1 | 1 | 1 | 1 | 1 |
| 177 | 1 | 1 | 1 | 1 | 1 | 1 | 1 | 1 | 1 | 1 |
| 178 | 1 | 0 | 1 | 1 | 1 | 1 | 1 | 1 | 0 | 1 |
| 179 | 0 | 1 | 0 | 0 | 0 | 0 | 0 | 0 | 0 | 0 |
| 180 | 1 | 1 | 1 | 1 | 1 | 1 | 1 | 1 | 1 | 1 |
| 181 | 1 | 1 | 1 | 1 | 1 | 1 | 1 | 1 | 1 | 1 |
| 182 | 1 | 1 | 1 | 1 | 1 | 1 | 1 | 1 | 1 | 1 |
| 183 | 1 | 1 | 1 | 1 | 1 | 1 | 1 | 1 | 1 | 1 |
| 184 | 1 | 1 | 1 | 1 | 1 | 1 | 1 | 1 | 1 | 1 |
| 185 | 1 | 1 | 1 | 1 | 1 | 1 | 1 | 1 | 1 | 1 |
| 186 | 1 | 1 | 1 | 1 | 1 | 1 | 1 | 1 | 1 | 1 |
| 187 | 1 | 1 | 1 | 1 | 1 | 1 | 1 | 1 | 1 | 1 |

| ID | UroBot_4_1 | UroBot_4_2 | UroBot_4_3 | UroBot_4_4 | UroBot_4_5 | UroBot_4_6 | UroBot_4_7 | UroBot_4_8 | UroBot_4_9 | UroBot_4_10 |
|---|---|---|---|---|---|---|---|---|---|---|
| 188 | 1 | 1 | 1 | 1 | 1 | 1 | 1 | 1 | 1 | 1 |
| 189 | 0 | 0 | 0 | 0 | 0 | 0 | 0 | 0 | 0 | 0 |
| 190 | 1 | 1 | 1 | 1 | 1 | 1 | 1 | 1 | 1 | 1 |
| 191 | 1 | 1 | 1 | 1 | 1 | 1 | 1 | 1 | 1 | 1 |
| 192 | 0 | 0 | 0 | 0 | 0 | 0 | 0 | 0 | 0 | 0 |
| 193 | 0 | 0 | 0 | 0 | 0 | 0 | 0 | 0 | 0 | 0 |
| 194 | 1 | 1 | 1 | 1 | 1 | 1 | 1 | 1 | 1 | 1 |
| 195 | 1 | 1 | 1 | 1 | 1 | 1 | 1 | 1 | 1 | 1 |
| 196 | 1 | 1 | 1 | 1 | 1 | 1 | 1 | 1 | 1 | 1 |
| 197 | 1 | 1 | 1 | 1 | 1 | 1 | 1 | 1 | 1 | 1 |
| 198 | 1 | 1 | 1 | 1 | 1 | 1 | 1 | 1 | 1 | 1 |
| 199 | 1 | 1 | 1 | 1 | 1 | 1 | 1 | 1 | 1 | 1 |
| 200 | 0 | 0 | 0 | 0 | 0 | 0 | 0 | 0 | 0 | 0 |

| ID | GPT_4o_1 | GPT_4o_2 | GPT_4o_3 | GPT_4o_4 | GPT_4o_5 | GPT_4o_6 | GPT_4o_7 | GPT_4o_8 | GPT_4o_9 | GPT_4o_10 |
|---|---|---|---|---|---|---|---|---|---|---|
| 1 | 1 | 1 | 1 | 1 | 1 | 1 | 1 | 1 | 1 | 1 |
| 2 | 1 | 1 | 1 | 1 | 1 | 1 | 1 | 1 | 1 | 1 |
| 3 | 0 | 0 | 0 | 0 | 0 | 0 | 0 | 0 | 0 | 0 |
| 4 | 1 | 1 | 1 | 1 | 1 | 1 | 1 | 1 | 1 | 1 |
| 5 | 1 | 1 | 1 | 1 | 1 | 1 | 1 | 1 | 1 | 1 |
| 6 | 1 | 1 | 1 | 1 | 1 | 1 | 1 | 1 | 1 | 1 |
| 7 | 1 | 1 | 1 | 1 | 1 | 1 | 1 | 1 | 1 | 1 |
| 8 | 1 | 1 | 1 | 1 | 1 | 1 | 1 | 1 | 1 | 1 |

| ID | GPT_4o_1 | GPT_4o_2 | GPT_4o_3 | GPT_4o_4 | GPT_4o_5 | GPT_4o_6 | GPT_4o_7 | GPT_4o_8 | GPT_4o_9 | GPT_4o_10 |
|---|---|---|---|---|---|---|---|---|---|---|
| 9 | 1 | 1 | 1 | 1 | 1 | 1 | 1 | 1 | 1 | 1 |
| 10 | 1 | 1 | 1 | 1 | 1 | 1 | 1 | 1 | 1 | 1 |
| 11 | 1 | 1 | 1 | 1 | 1 | 1 | 1 | 1 | 1 | 1 |
| 12 | 1 | 1 | 1 | 1 | 1 | 1 | 1 | 1 | 1 | 1 |
| 13 | 1 | 1 | 1 | 1 | 1 | 1 | 1 | 1 | 1 | 1 |
| 14 | 1 | 1 | 1 | 1 | 1 | 1 | 1 | 1 | 1 | 1 |
| 15 | 1 | 1 | 1 | 1 | 1 | 1 | 1 | 1 | 1 | 1 |
| 16 | 0 | 0 | 0 | 0 | 0 | 0 | 0 | 0 | 0 | 0 |
| 17 | 1 | 1 | 1 | 1 | 1 | 1 | 1 | 1 | 1 | 1 |
| 18 | 1 | 1 | 1 | 1 | 1 | 1 | 1 | 1 | 1 | 1 |
| 19 | 0 | 0 | 0 | 0 | 0 | 0 | 0 | 0 | 0 | 0 |
| 20 | 1 | 1 | 1 | 1 | 1 | 1 | 1 | 1 | 1 | 1 |
| 21 | 1 | 1 | 1 | 1 | 1 | 1 | 1 | 1 | 1 | 1 |
| 22 | 0 | 0 | 0 | 0 | 0 | 0 | 0 | 0 | 0 | 0 |
| 23 | 1 | 1 | 1 | 1 | 1 | 1 | 1 | 1 | 1 | 1 |
| 24 | 0 | 0 | 0 | 0 | 0 | 0 | 0 | 0 | 0 | 0 |
| 25 | 1 | 1 | 1 | 1 | 1 | 1 | 1 | 1 | 1 | 1 |
| 26 | 1 | 0 | 1 | 0 | 1 | 1 | 1 | 1 | 1 | 1 |
| 27 | 1 | 1 | 1 | 1 | 1 | 1 | 1 | 1 | 1 | 1 |
| 28 | 1 | 1 | 1 | 1 | 1 | 1 | 1 | 1 | 1 | 1 |
| 29 | 1 | 1 | 1 | 1 | 1 | 1 | 1 | 1 | 1 | 1 |
| 30 | 1 | 1 | 1 | 1 | 1 | 1 | 1 | 1 | 1 | 1 |
| 31 | 1 | 1 | 1 | 1 | 1 | 1 | 1 | 1 | 1 | 1 |
| 32 | 0 | 0 | 0 | 0 | 0 | 0 | 0 | 0 | 0 | 0 |
| 33 | 1 | 1 | 1 | 1 | 1 | 1 | 1 | 1 | 1 | 1 |
| 34 | 1 | 1 | 1 | 1 | 1 | 1 | 1 | 1 | 1 | 1 |
| 35 | 1 | 1 | 1 | 1 | 1 | 1 | 1 | 1 | 1 | 1 |

| ID | GPT_4o_1 | GPT_4o_2 | GPT_4o_3 | GPT_4o_4 | GPT_4o_5 | GPT_4o_6 | GPT_4o_7 | GPT_4o_8 | GPT_4o_9 | GPT_4o_10 |
|---|---|---|---|---|---|---|---|---|---|---|
| 36 | 1 | 1 | 1 | 1 | 1 | 1 | 1 | 1 | 1 | 1 |
| 37 | 0 | 0 | 0 | 0 | 0 | 0 | 0 | 0 | 0 | 0 |
| 38 | 0 | 0 | 0 | 0 | 0 | 0 | 0 | 0 | 0 | 0 |
| 39 | 0 | 0 | 0 | 0 | 0 | 0 | 0 | 0 | 0 | 0 |
| 40 | 1 | 1 | 1 | 1 | 1 | 1 | 1 | 1 | 1 | 1 |
| 41 | 1 | 1 | 1 | 1 | 1 | 1 | 1 | 1 | 1 | 1 |
| 42 | 1 | 1 | 1 | 1 | 1 | 1 | 1 | 1 | 1 | 1 |
| 43 | 0 | 0 | 0 | 0 | 0 | 0 | 0 | 0 | 0 | 0 |
| 44 | 1 | 1 | 1 | 1 | 1 | 1 | 1 | 1 | 1 | 1 |
| 45 | 1 | 1 | 1 | 1 | 1 | 1 | 1 | 1 | 1 | 1 |
| 46 | 1 | 1 | 1 | 1 | 1 | 1 | 1 | 1 | 1 | 1 |
| 47 | 0 | 0 | 0 | 0 | 0 | 0 | 0 | 0 | 0 | 0 |
| 48 | 1 | 1 | 1 | 1 | 1 | 1 | 1 | 1 | 1 | 1 |
| 49 | 1 | 1 | 1 | 1 | 1 | 1 | 1 | 1 | 1 | 1 |
| 50 | 1 | 1 | 1 | 1 | 1 | 1 | 1 | 1 | 1 | 1 |
| 51 | 0 | 0 | 1 | 1 | 1 | 1 | 0 | 0 | 1 | 1 |
| 52 | 1 | 1 | 1 | 1 | 1 | 1 | 1 | 1 | 1 | 1 |
| 53 | 1 | 1 | 1 | 1 | 1 | 1 | 1 | 1 | 1 | 1 |
| 54 | 1 | 1 | 1 | 1 | 1 | 0 | 1 | 1 | 0 | 1 |
| 55 | 1 | 1 | 1 | 1 | 1 | 1 | 1 | 1 | 1 | 1 |
| 56 | 1 | 1 | 1 | 1 | 1 | 1 | 1 | 1 | 1 | 1 |
| 57 | 1 | 1 | 1 | 1 | 1 | 1 | 1 | 1 | 1 | 1 |
| 58 | 1 | 1 | 1 | 1 | 1 | 1 | 1 | 1 | 1 | 1 |
| 59 | 1 | 1 | 1 | 1 | 1 | 1 | 1 | 1 | 1 | 1 |
| 60 | 0 | 0 | 0 | 0 | 0 | 0 | 0 | 0 | 0 | 0 |
| 61 | 1 | 1 | 1 | 1 | 1 | 1 | 1 | 1 | 1 | 1 |
| 62 | 1 | 1 | 1 | 1 | 1 | 1 | 1 | 1 | 1 | 1 |

| ID | GPT_4o_1 | GPT_4o_2 | GPT_4o_3 | GPT_4o_4 | GPT_4o_5 | GPT_4o_6 | GPT_4o_7 | GPT_4o_8 | GPT_4o_9 | GPT_4o_10 |
|---|---|---|---|---|---|---|---|---|---|---|
| 63 | 1 | 1 | 1 | 1 | 1 | 1 | 1 | 1 | 1 | 1 |
| 64 | 1 | 1 | 1 | 1 | 1 | 1 | 1 | 1 | 1 | 1 |
| 65 | 1 | 1 | 1 | 1 | 1 | 1 | 1 | 1 | 1 | 1 |
| 66 | 1 | 1 | 1 | 1 | 1 | 1 | 1 | 1 | 1 | 1 |
| 67 | 1 | 1 | 1 | 1 | 1 | 1 | 1 | 1 | 1 | 1 |
| 68 | 0 | 0 | 0 | 0 | 0 | 0 | 0 | 0 | 0 | 0 |
| 69 | 0 | 0 | 0 | 0 | 0 | 0 | 0 | 0 | 0 | 0 |
| 70 | 1 | 1 | 1 | 1 | 1 | 1 | 1 | 1 | 1 | 1 |
| 71 | 1 | 1 | 1 | 1 | 1 | 1 | 1 | 1 | 1 | 1 |
| 72 | 1 | 1 | 1 | 1 | 1 | 1 | 1 | 1 | 1 | 1 |
| 73 | 1 | 1 | 1 | 1 | 1 | 1 | 1 | 1 | 1 | 1 |
| 74 | 1 | 1 | 1 | 1 | 1 | 1 | 1 | 1 | 1 | 1 |
| 75 | 1 | 1 | 1 | 1 | 1 | 1 | 1 | 1 | 1 | 1 |
| 76 | 1 | 1 | 1 | 1 | 1 | 1 | 1 | 1 | 1 | 1 |
| 77 | 1 | 1 | 1 | 1 | 1 | 1 | 1 | 1 | 1 | 1 |
| 78 | 1 | 1 | 1 | 1 | 1 | 1 | 1 | 1 | 1 | 1 |
| 79 | 1 | 1 | 1 | 1 | 1 | 1 | 1 | 1 | 1 | 1 |
| 80 | 0 | 0 | 0 | 0 | 0 | 0 | 0 | 0 | 0 | 0 |
| 81 | 0 | 0 | 0 | 0 | 0 | 0 | 0 | 0 | 0 | 0 |
| 82 | 0 | 0 | 0 | 0 | 0 | 0 | 0 | 0 | 0 | 0 |
| 83 | 1 | 1 | 1 | 1 | 1 | 1 | 1 | 1 | 1 | 1 |
| 84 | 1 | 1 | 1 | 1 | 1 | 1 | 1 | 1 | 1 | 1 |
| 85 | 0 | 0 | 0 | 0 | 0 | 0 | 0 | 0 | 0 | 0 |
| 86 | 1 | 1 | 1 | 1 | 1 | 1 | 1 | 1 | 1 | 1 |
| 87 | 1 | 1 | 1 | 1 | 1 | 1 | 1 | 1 | 1 | 1 |
| 88 | 1 | 1 | 0 | 1 | 1 | 0 | 0 | 1 | 0 | 0 |
| 89 | 1 | 1 | 1 | 1 | 1 | 1 | 1 | 1 | 1 | 1 |

| ID | GPT_4o_1 | GPT_4o_2 | GPT_4o_3 | GPT_4o_4 | GPT_4o_5 | GPT_4o_6 | GPT_4o_7 | GPT_4o_8 | GPT_4o_9 | GPT_4o_10 |
|-----|----------|----------|----------|----------|----------|----------|----------|----------|----------|-----------|
| 90 | 1 | 1 | 1 | 1 | 1 | 1 | 1 | 1 | 1 | 1 |
| 91 | 1 | 1 | 1 | 1 | 1 | 1 | 1 | 1 | 1 | 1 |
| 92 | 1 | 1 | 1 | 1 | 1 | 1 | 1 | 1 | 1 | 1 |
| 93 | 1 | 1 | 1 | 1 | 1 | 1 | 1 | 1 | 1 | 1 |
| 94 | 1 | 1 | 1 | 1 | 1 | 1 | 1 | 1 | 1 | 1 |
| 95 | 1 | 1 | 1 | 1 | 1 | 1 | 1 | 1 | 1 | 1 |
| 96 | 1 | 1 | 1 | 1 | 1 | 1 | 1 | 1 | 1 | 1 |
| 97 | 1 | 1 | 1 | 1 | 1 | 1 | 1 | 1 | 1 | 1 |
| 98 | 0 | 0 | 0 | 0 | 0 | 0 | 0 | 0 | 0 | 0 |
| 99 | 1 | 1 | 1 | 1 | 1 | 1 | 1 | 1 | 1 | 1 |
| 100 | 1 | 1 | 1 | 1 | 1 | 1 | 1 | 1 | 1 | 1 |
| 101 | 1 | 0 | 0 | 1 | 0 | 0 | 0 | 0 | 0 | 0 |
| 102 | 1 | 1 | 1 | 1 | 1 | 1 | 1 | 1 | 1 | 1 |
| 103 | 1 | 1 | 1 | 1 | 1 | 1 | 1 | 1 | 1 | 1 |
| 104 | 0 | 0 | 0 | 0 | 0 | 0 | 0 | 0 | 0 | 0 |
| 105 | 1 | 1 | 1 | 1 | 1 | 1 | 1 | 1 | 1 | 1 |
| 106 | 0 | 0 | 0 | 0 | 0 | 0 | 0 | 0 | 0 | 0 |
| 107 | 1 | 1 | 1 | 1 | 1 | 1 | 1 | 1 | 1 | 1 |
| 108 | 1 | 1 | 1 | 1 | 1 | 1 | 1 | 1 | 1 | 1 |
| 109 | 1 | 1 | 1 | 1 | 1 | 1 | 1 | 1 | 1 | 1 |
| 110 | 0 | 0 | 0 | 0 | 0 | 0 | 0 | 0 | 0 | 0 |
| 111 | 1 | 1 | 1 | 1 | 1 | 1 | 1 | 1 | 1 | 1 |
| 112 | 0 | 0 | 1 | 0 | 1 | 1 | 1 | 1 | 0 | 1 |
| 113 | 1 | 1 | 1 | 1 | 1 | 1 | 1 | 1 | 1 | 1 |
| 114 | 1 | 1 | 1 | 1 | 1 | 1 | 1 | 1 | 1 | 1 |
| 115 | 0 | 0 | 0 | 0 | 0 | 0 | 0 | 0 | 0 | 0 |
| 116 | 1 | 1 | 1 | 1 | 1 | 1 | 1 | 1 | 1 | 1 |

| ID | GPT_4o_1 | GPT_4o_2 | GPT_4o_3 | GPT_4o_4 | GPT_4o_5 | GPT_4o_6 | GPT_4o_7 | GPT_4o_8 | GPT_4o_9 | GPT_4o_10 |
|---|---|---|---|---|---|---|---|---|---|---|
| 117 | 0 | 0 | 0 | 0 | 0 | 0 | 0 | 0 | 0 | 0 |
| 118 | 1 | 1 | 1 | 1 | 1 | 1 | 1 | 1 | 1 | 1 |
| 119 | 1 | 1 | 1 | 1 | 1 | 1 | 1 | 1 | 1 | 1 |
| 120 | 1 | 1 | 1 | 1 | 1 | 1 | 1 | 1 | 1 | 1 |
| 121 | 1 | 1 | 1 | 1 | 1 | 1 | 1 | 1 | 1 | 1 |
| 122 | 1 | 1 | 1 | 1 | 1 | 1 | 1 | 1 | 1 | 1 |
| 123 | 0 | 0 | 0 | 0 | 0 | 0 | 0 | 0 | 0 | 0 |
| 124 | 1 | 1 | 1 | 1 | 1 | 1 | 1 | 1 | 1 | 1 |
| 125 | 1 | 1 | 1 | 1 | 1 | 1 | 1 | 1 | 1 | 1 |
| 126 | 0 | 0 | 0 | 0 | 0 | 0 | 0 | 0 | 0 | 0 |
| 127 | 1 | 1 | 1 | 1 | 1 | 1 | 1 | 1 | 1 | 1 |
| 128 | 1 | 1 | 1 | 1 | 1 | 1 | 1 | 1 | 1 | 1 |
| 129 | 0 | 0 | 0 | 0 | 0 | 0 | 0 | 0 | 0 | 0 |
| 130 | 1 | 1 | 1 | 1 | 1 | 1 | 1 | 1 | 1 | 1 |
| 131 | 0 | 0 | 0 | 0 | 0 | 0 | 0 | 0 | 0 | 0 |
| 132 | 1 | 1 | 1 | 1 | 1 | 1 | 1 | 1 | 1 | 1 |
| 133 | 1 | 1 | 1 | 1 | 1 | 1 | 1 | 1 | 1 | 1 |
| 134 | 1 | 1 | 1 | 1 | 1 | 1 | 1 | 1 | 1 | 1 |
| 135 | 1 | 1 | 1 | 1 | 1 | 1 | 1 | 1 | 1 | 1 |
| 136 | 1 | 1 | 1 | 1 | 1 | 1 | 1 | 1 | 1 | 1 |
| 137 | 1 | 1 | 1 | 1 | 1 | 1 | 1 | 1 | 1 | 1 |
| 138 | 1 | 1 | 1 | 1 | 1 | 1 | 1 | 1 | 1 | 1 |
| 139 | 0 | 0 | 0 | 0 | 0 | 0 | 0 | 0 | 0 | 0 |
| 140 | 0 | 1 | 1 | 0 | 0 | 0 | 1 | 1 | 0 | 0 |
| 141 | 1 | 1 | 1 | 1 | 1 | 1 | 1 | 1 | 1 | 1 |
| 142 | 1 | 1 | 1 | 1 | 1 | 1 | 1 | 1 | 1 | 1 |
| 143 | 1 | 1 | 1 | 1 | 1 | 1 | 1 | 1 | 1 | 1 |

| ID | GPT_4o_1 | GPT_4o_2 | GPT_4o_3 | GPT_4o_4 | GPT_4o_5 | GPT_4o_6 | GPT_4o_7 | GPT_4o_8 | GPT_4o_9 | GPT_4o_10 |
|---|---|---|---|---|---|---|---|---|---|---|
| 144 | 1 | 1 | 1 | 1 | 1 | 1 | 1 | 1 | 1 | 1 |
| 145 | 1 | 1 | 1 | 1 | 1 | 1 | 1 | 1 | 1 | 1 |
| 146 | 0 | 0 | 0 | 0 | 0 | 0 | 0 | 0 | 0 | 0 |
| 147 | 1 | 1 | 1 | 1 | 1 | 1 | 1 | 1 | 1 | 1 |
| 148 | 1 | 1 | 1 | 1 | 1 | 1 | 1 | 1 | 1 | 1 |
| 149 | 1 | 1 | 1 | 1 | 1 | 1 | 1 | 1 | 1 | 1 |
| 150 | 1 | 1 | 1 | 1 | 1 | 1 | 1 | 1 | 1 | 1 |
| 151 | 1 | 1 | 1 | 1 | 1 | 1 | 1 | 1 | 1 | 1 |
| 152 | 0 | 0 | 0 | 0 | 0 | 0 | 0 | 0 | 0 | 0 |
| 153 | 0 | 1 | 1 | 0 | 0 | 1 | 0 | 0 | 0 | 0 |
| 154 | 1 | 1 | 1 | 1 | 1 | 1 | 1 | 1 | 1 | 1 |
| 155 | 1 | 1 | 1 | 1 | 1 | 1 | 1 | 1 | 1 | 1 |
| 156 | 1 | 1 | 1 | 1 | 1 | 1 | 1 | 1 | 1 | 1 |
| 157 | 1 | 1 | 1 | 1 | 1 | 1 | 1 | 1 | 1 | 1 |
| 158 | 1 | 1 | 1 | 1 | 1 | 1 | 1 | 1 | 1 | 1 |
| 159 | 1 | 1 | 1 | 1 | 1 | 1 | 1 | 1 | 1 | 1 |
| 160 | 1 | 1 | 1 | 1 | 1 | 1 | 1 | 1 | 1 | 1 |
| 161 | 1 | 1 | 1 | 1 | 1 | 1 | 1 | 1 | 1 | 1 |
| 162 | 1 | 1 | 1 | 1 | 1 | 1 | 1 | 1 | 1 | 1 |
| 163 | 1 | 1 | 1 | 1 | 1 | 1 | 1 | 1 | 1 | 1 |
| 164 | 1 | 1 | 1 | 1 | 1 | 1 | 1 | 1 | 1 | 1 |
| 165 | 1 | 1 | 1 | 1 | 1 | 1 | 1 | 1 | 1 | 1 |
| 166 | 1 | 1 | 1 | 1 | 1 | 1 | 1 | 1 | 1 | 1 |
| 167 | 1 | 1 | 1 | 1 | 1 | 1 | 1 | 1 | 1 | 1 |
| 168 | 1 | 1 | 1 | 1 | 1 | 1 | 1 | 1 | 1 | 1 |
| 169 | 1 | 1 | 1 | 1 | 1 | 1 | 1 | 1 | 1 | 1 |
| 170 | 1 | 1 | 1 | 1 | 1 | 1 | 1 | 1 | 1 | 1 |

| ID | GPT_4o_1 | GPT_4o_2 | GPT_4o_3 | GPT_4o_4 | GPT_4o_5 | GPT_4o_6 | GPT_4o_7 | GPT_4o_8 | GPT_4o_9 | GPT_4o_10 |
|---|---|---|---|---|---|---|---|---|---|---|
| 171 | 0 | 0 | 0 | 0 | 0 | 0 | 0 | 0 | 0 | 0 |
| 172 | 0 | 0 | 1 | 1 | 1 | 1 | 1 | 1 | 1 | 1 |
| 173 | 1 | 1 | 1 | 1 | 1 | 1 | 1 | 1 | 1 | 1 |
| 174 | 1 | 1 | 1 | 1 | 1 | 1 | 1 | 1 | 1 | 1 |
| 175 | 1 | 1 | 1 | 1 | 1 | 1 | 1 | 1 | 1 | 1 |
| 176 | 1 | 1 | 1 | 1 | 1 | 1 | 1 | 1 | 1 | 1 |
| 177 | 1 | 1 | 1 | 1 | 1 | 1 | 1 | 1 | 1 | 1 |
| 178 | 1 | 1 | 1 | 1 | 1 | 1 | 1 | 1 | 1 | 1 |
| 179 | 0 | 0 | 0 | 0 | 0 | 0 | 0 | 0 | 0 | 0 |
| 180 | 0 | 0 | 0 | 0 | 0 | 0 | 0 | 0 | 0 | 0 |
| 181 | 1 | 1 | 1 | 1 | 1 | 1 | 1 | 1 | 1 | 1 |
| 182 | 1 | 1 | 1 | 1 | 1 | 1 | 1 | 1 | 1 | 1 |
| 183 | 1 | 1 | 0 | 1 | 0 | 0 | 1 | 1 | 1 | 1 |
| 184 | 1 | 1 | 1 | 1 | 1 | 1 | 1 | 1 | 1 | 1 |
| 185 | 0 | 0 | 0 | 0 | 0 | 0 | 0 | 0 | 0 | 0 |
| 186 | 1 | 1 | 1 | 1 | 1 | 1 | 1 | 1 | 1 | 1 |
| 187 | 1 | 0 | 0 | 1 | 0 | 0 | 0 | 0 | 0 | 1 |
| 188 | 1 | 0 | 0 | 1 | 0 | 0 | 0 | 0 | 0 | 0 |
| 189 | 0 | 1 | 0 | 1 | 1 | 1 | 0 | 1 | 0 | 0 |
| 190 | 1 | 1 | 1 | 1 | 1 | 1 | 1 | 1 | 1 | 1 |
| 191 | 1 | 1 | 1 | 1 | 1 | 1 | 1 | 1 | 1 | 1 |
| 192 | 0 | 0 | 0 | 0 | 0 | 0 | 0 | 0 | 0 | 0 |
| 193 | 1 | 1 | 1 | 1 | 1 | 1 | 1 | 1 | 1 | 1 |
| 194 | 1 | 1 | 0 | 1 | 1 | 1 | 1 | 1 | 1 | 1 |
| 195 | 1 | 1 | 1 | 1 | 1 | 1 | 1 | 1 | 1 | 1 |
| 196 | 1 | 0 | 1 | 1 | 0 | 0 | 1 | 0 | 1 | 1 |
| 197 | 1 | 1 | 1 | 1 | 1 | 1 | 1 | 1 | 1 | 1 |

| ID | GPT_4o_1 | GPT_4o_2 | GPT_4o_3 | GPT_4o_4 | GPT_4o_5 | GPT_4o_6 | GPT_4o_7 | GPT_4o_8 | GPT_4o_9 | GPT_4o_10 |
|---|---|---|---|---|---|---|---|---|---|---|
| 198 | 0 | 0 | 0 | 0 | 0 | 0 | 0 | 0 | 0 | 0 |
| 199 | 1 | 1 | 1 | 1 | 1 | 1 | 1 | 1 | 1 | 1 |
| 200 | 0 | 0 | 0 | 0 | 0 | 0 | 0 | 0 | 0 | 0 |

| ID | UroBot_4o_1 | UroBot_4o_2 | UroBot_4o_3 | UroBot_4o_4 | UroBot_4o_5 | UroBot_4o_6 | UroBot_4o_7 | UroBot_4o_8 | UroBot_4o_9 | UroBot_4o_10 |
|---|---|---|---|---|---|---|---|---|---|---|
| 1 | 1 | 1 | 1 | 1 | 1 | 1 | 1 | 1 | 1 | 1 |
| 2 | 1 | 1 | 1 | 1 | 1 | 1 | 1 | 1 | 1 | 1 |
| 3 | 1 | 1 | 1 | 1 | 1 | 1 | 1 | 1 | 1 | 1 |
| 4 | 1 | 1 | 1 | 1 | 1 | 1 | 1 | 1 | 1 | 1 |
| 5 | 1 | 1 | 1 | 1 | 1 | 1 | 1 | 1 | 1 | 1 |
| 6 | 1 | 1 | 1 | 1 | 1 | 1 | 1 | 1 | 1 | 1 |
| 7 | 1 | 1 | 1 | 1 | 1 | 1 | 1 | 1 | 1 | 1 |
| 8 | 1 | 1 | 1 | 1 | 1 | 1 | 1 | 1 | 1 | 1 |
| 9 | 1 | 1 | 1 | 1 | 1 | 1 | 1 | 1 | 1 | 1 |
| 10 | 1 | 1 | 1 | 1 | 1 | 1 | 1 | 1 | 1 | 1 |
| 11 | 1 | 1 | 1 | 1 | 1 | 1 | 1 | 1 | 1 | 1 |
| 12 | 1 | 1 | 1 | 1 | 1 | 1 | 1 | 1 | 1 | 1 |
| 13 | 1 | 1 | 1 | 1 | 1 | 1 | 1 | 1 | 1 | 1 |
| 14 | 1 | 1 | 1 | 1 | 1 | 1 | 1 | 1 | 1 | 1 |
| 15 | 1 | 1 | 1 | 1 | 1 | 1 | 1 | 1 | 1 | 1 |
| 16 | 1 | 1 | 1 | 1 | 1 | 1 | 1 | 1 | 1 | 1 |
| 17 | 1 | 1 | 1 | 1 | 1 | 1 | 1 | 1 | 1 | 1 |
| 18 | 1 | 1 | 1 | 1 | 1 | 1 | 1 | 1 | 1 | 1 |
| 19 | 1 | 1 | 1 | 1 | 1 | 1 | 1 | 1 | 1 | 1 |

| ID | UroBot_4o_1 | UroBot_4o_2 | UroBot_4o_3 | UroBot_4o_4 | UroBot_4o_5 | UroBot_4o_6 | UroBot_4o_7 | UroBot_4o_8 | UroBot_4o_9 | UroBot_4o_10 |
|----|----|----|----|----|----|----|----|----|----|----|
| 20 | 1 | 1 | 1 | 1 | 1 | 1 | 1 | 1 | 1 | 1 |
| 21 | 1 | 1 | 1 | 1 | 1 | 1 | 1 | 1 | 1 | 1 |
| 22 | 0 | 0 | 0 | 0 | 0 | 0 | 0 | 0 | 0 | 0 |
| 23 | 1 | 1 | 1 | 1 | 1 | 1 | 1 | 1 | 1 | 1 |
| 24 | 1 | 0 | 0 | 0 | 0 | 0 | 0 | 0 | 0 | 0 |
| 25 | 1 | 1 | 1 | 1 | 1 | 1 | 1 | 1 | 1 | 1 |
| 26 | 1 | 1 | 1 | 1 | 1 | 1 | 1 | 1 | 1 | 1 |
| 27 | 1 | 1 | 1 | 1 | 1 | 1 | 1 | 1 | 1 | 1 |
| 28 | 1 | 1 | 1 | 1 | 1 | 1 | 1 | 1 | 1 | 1 |
| 29 | 1 | 1 | 1 | 1 | 1 | 1 | 1 | 1 | 1 | 1 |
| 30 | 1 | 1 | 1 | 1 | 1 | 1 | 1 | 1 | 1 | 1 |
| 31 | 1 | 1 | 1 | 1 | 1 | 1 | 1 | 1 | 1 | 1 |
| 32 | 1 | 1 | 1 | 1 | 1 | 1 | 1 | 1 | 1 | 1 |
| 33 | 1 | 1 | 1 | 1 | 1 | 1 | 1 | 1 | 1 | 1 |
| 34 | 1 | 1 | 1 | 1 | 1 | 1 | 1 | 1 | 1 | 1 |
| 35 | 1 | 1 | 1 | 1 | 1 | 1 | 1 | 1 | 1 | 1 |
| 36 | 1 | 1 | 1 | 1 | 1 | 1 | 1 | 1 | 1 | 1 |
| 37 | 0 | 0 | 0 | 0 | 0 | 0 | 0 | 0 | 0 | 0 |
| 38 | 0 | 0 | 0 | 0 | 0 | 0 | 0 | 0 | 0 | 0 |
| 39 | 1 | 1 | 1 | 1 | 1 | 1 | 1 | 1 | 1 | 1 |
| 40 | 1 | 1 | 1 | 1 | 1 | 1 | 1 | 1 | 1 | 1 |
| 41 | 0 | 0 | 0 | 0 | 0 | 0 | 0 | 0 | 0 | 0 |
| 42 | 1 | 1 | 1 | 1 | 1 | 1 | 1 | 1 | 1 | 1 |
| 43 | 1 | 1 | 1 | 1 | 1 | 1 | 1 | 1 | 1 | 1 |
| 44 | 1 | 1 | 1 | 1 | 1 | 1 | 1 | 1 | 1 | 1 |
| 45 | 1 | 1 | 1 | 1 | 1 | 1 | 1 | 1 | 1 | 1 |

| ID | UroBot_4o_1 | UroBot_4o_2 | UroBot_4o_3 | UroBot_4o_4 | UroBot_4o_5 | UroBot_4o_6 | UroBot_4o_7 | UroBot_4o_8 | UroBot_4o_9 | UroBot_4o_10 |
|----|-------------|-------------|-------------|-------------|-------------|-------------|-------------|-------------|-------------|--------------|
| 46 | 1 | 1 | 1 | 1 | 1 | 1 | 1 | 1 | 1 | 1 |
| 47 | 1 | 1 | 1 | 1 | 1 | 1 | 1 | 1 | 1 | 1 |
| 48 | 1 | 1 | 1 | 1 | 1 | 1 | 1 | 1 | 1 | 1 |
| 49 | 1 | 1 | 1 | 1 | 1 | 1 | 1 | 1 | 1 | 1 |
| 50 | 1 | 1 | 1 | 1 | 1 | 1 | 1 | 1 | 1 | 1 |
| 51 | 1 | 1 | 1 | 1 | 1 | 1 | 1 | 1 | 1 | 1 |
| 52 | 1 | 1 | 1 | 1 | 1 | 1 | 1 | 1 | 1 | 1 |
| 53 | 1 | 1 | 1 | 1 | 1 | 1 | 1 | 1 | 1 | 1 |
| 54 | 1 | 1 | 1 | 1 | 1 | 1 | 1 | 1 | 1 | 1 |
| 55 | 1 | 1 | 1 | 1 | 1 | 1 | 1 | 1 | 1 | 1 |
| 56 | 1 | 1 | 1 | 1 | 1 | 1 | 1 | 1 | 1 | 1 |
| 57 | 1 | 1 | 1 | 1 | 1 | 1 | 1 | 1 | 1 | 1 |
| 58 | 1 | 1 | 1 | 1 | 1 | 1 | 1 | 1 | 1 | 1 |
| 59 | 1 | 1 | 1 | 1 | 1 | 1 | 1 | 1 | 1 | 1 |
| 60 | 1 | 1 | 1 | 1 | 1 | 1 | 1 | 1 | 1 | 1 |
| 61 | 0 | 0 | 0 | 0 | 0 | 0 | 0 | 0 | 0 | 0 |
| 62 | 1 | 1 | 1 | 1 | 0 | 1 | 1 | 1 | 1 | 0 |
| 63 | 1 | 1 | 1 | 1 | 1 | 1 | 1 | 1 | 1 | 1 |
| 64 | 1 | 1 | 1 | 1 | 1 | 1 | 1 | 1 | 1 | 1 |
| 65 | 1 | 1 | 1 | 1 | 1 | 1 | 1 | 1 | 1 | 1 |
| 66 | 1 | 1 | 1 | 1 | 1 | 1 | 1 | 1 | 1 | 1 |
| 67 | 1 | 1 | 1 | 1 | 1 | 1 | 1 | 1 | 1 | 1 |
| 68 | 1 | 1 | 1 | 1 | 1 | 1 | 1 | 1 | 1 | 1 |
| 69 | 1 | 1 | 1 | 1 | 1 | 1 | 1 | 1 | 1 | 1 |
| 70 | 1 | 1 | 1 | 1 | 1 | 1 | 1 | 1 | 1 | 1 |
| 71 | 1 | 1 | 1 | 1 | 1 | 1 | 1 | 1 | 1 | 1 |

| ID | UroBot_4o_1 | UroBot_4o_2 | UroBot_4o_3 | UroBot_4o_4 | UroBot_4o_5 | UroBot_4o_6 | UroBot_4o_7 | UroBot_4o_8 | UroBot_4o_9 | UroBot_4o_10 |
|---|---|---|---|---|---|---|---|---|---|---|
| 72 | 1 | 1 | 1 | 1 | 1 | 1 | 1 | 1 | 1 | 1 |
| 73 | 1 | 1 | 1 | 1 | 1 | 1 | 1 | 1 | 1 | 1 |
| 74 | 1 | 1 | 1 | 1 | 1 | 1 | 1 | 1 | 1 | 1 |
| 75 | 1 | 1 | 1 | 1 | 1 | 1 | 1 | 1 | 1 | 1 |
| 76 | 1 | 1 | 1 | 1 | 1 | 1 | 1 | 1 | 1 | 1 |
| 77 | 1 | 1 | 1 | 1 | 1 | 1 | 1 | 1 | 1 | 1 |
| 78 | 1 | 1 | 1 | 1 | 1 | 1 | 1 | 1 | 1 | 1 |
| 79 | 1 | 1 | 1 | 1 | 1 | 1 | 1 | 1 | 1 | 1 |
| 80 | 0 | 0 | 0 | 0 | 0 | 0 | 0 | 0 | 0 | 0 |
| 81 | 1 | 1 | 1 | 1 | 1 | 1 | 1 | 1 | 1 | 1 |
| 82 | 0 | 0 | 0 | 0 | 0 | 0 | 0 | 0 | 0 | 0 |
| 83 | 1 | 1 | 1 | 1 | 1 | 1 | 1 | 1 | 1 | 1 |
| 84 | 0 | 0 | 0 | 0 | 0 | 0 | 0 | 0 | 0 | 0 |
| 85 | 0 | 0 | 0 | 0 | 0 | 0 | 0 | 0 | 0 | 0 |
| 86 | 1 | 1 | 1 | 1 | 1 | 1 | 1 | 1 | 1 | 1 |
| 87 | 1 | 1 | 1 | 1 | 1 | 1 | 1 | 1 | 1 | 1 |
| 88 | 1 | 1 | 1 | 1 | 1 | 1 | 1 | 1 | 1 | 1 |
| 89 | 1 | 1 | 1 | 1 | 1 | 1 | 1 | 1 | 1 | 1 |
| 90 | 1 | 1 | 1 | 1 | 1 | 1 | 1 | 1 | 1 | 1 |
| 91 | 1 | 1 | 1 | 1 | 1 | 1 | 1 | 1 | 1 | 1 |
| 92 | 1 | 1 | 1 | 1 | 1 | 1 | 1 | 1 | 1 | 1 |
| 93 | 1 | 1 | 1 | 1 | 1 | 1 | 1 | 1 | 1 | 1 |
| 94 | 1 | 1 | 1 | 1 | 1 | 1 | 1 | 1 | 1 | 1 |
| 95 | 1 | 1 | 1 | 1 | 1 | 1 | 1 | 1 | 1 | 1 |
| 96 | 1 | 1 | 1 | 1 | 1 | 1 | 1 | 1 | 1 | 1 |
| 97 | 1 | 1 | 1 | 1 | 1 | 1 | 1 | 1 | 1 | 1 |

| ID | UroBot_4o_1 | UroBot_4o_2 | UroBot_4o_3 | UroBot_4o_4 | UroBot_4o_5 | UroBot_4o_6 | UroBot_4o_7 | UroBot_4o_8 | UroBot_4o_9 | UroBot_4o_10 |
|----|-------------|-------------|-------------|-------------|-------------|-------------|-------------|-------------|-------------|--------------|
| 98 | 1 | 1 | 1 | 1 | 1 | 1 | 1 | 1 | 1 | 1 |
| 99 | 1 | 1 | 1 | 1 | 1 | 1 | 1 | 1 | 1 | 1 |
| 100 | 1 | 1 | 1 | 1 | 1 | 1 | 1 | 1 | 1 | 1 |
| 101 | 1 | 1 | 1 | 1 | 1 | 1 | 1 | 1 | 1 | 1 |
| 102 | 1 | 1 | 1 | 1 | 1 | 1 | 1 | 1 | 1 | 1 |
| 103 | 1 | 1 | 1 | 1 | 1 | 1 | 1 | 1 | 1 | 1 |
| 104 | 0 | 0 | 0 | 0 | 0 | 0 | 0 | 0 | 0 | 0 |
| 105 | 1 | 1 | 1 | 1 | 1 | 1 | 1 | 1 | 1 | 1 |
| 106 | 1 | 1 | 1 | 1 | 1 | 1 | 1 | 1 | 1 | 1 |
| 107 | 1 | 1 | 1 | 1 | 1 | 1 | 1 | 1 | 1 | 1 |
| 108 | 1 | 1 | 1 | 1 | 1 | 1 | 1 | 1 | 1 | 1 |
| 109 | 1 | 1 | 1 | 1 | 1 | 1 | 1 | 1 | 1 | 1 |
| 110 | 1 | 1 | 1 | 1 | 1 | 1 | 1 | 1 | 1 | 1 |
| 111 | 1 | 1 | 1 | 1 | 1 | 1 | 1 | 1 | 1 | 1 |
| 112 | 0 | 0 | 0 | 0 | 0 | 0 | 0 | 0 | 0 | 0 |
| 113 | 1 | 1 | 1 | 1 | 1 | 1 | 1 | 1 | 1 | 1 |
| 114 | 1 | 1 | 1 | 1 | 1 | 1 | 1 | 1 | 1 | 1 |
| 115 | 0 | 0 | 0 | 0 | 0 | 0 | 0 | 0 | 0 | 0 |
| 116 | 1 | 1 | 1 | 1 | 1 | 1 | 1 | 1 | 1 | 1 |
| 117 | 0 | 0 | 0 | 0 | 0 | 0 | 0 | 0 | 0 | 0 |
| 118 | 1 | 1 | 1 | 1 | 1 | 1 | 1 | 1 | 1 | 1 |
| 119 | 1 | 1 | 1 | 1 | 1 | 1 | 1 | 1 | 1 | 1 |
| 120 | 1 | 1 | 1 | 1 | 1 | 1 | 1 | 1 | 1 | 1 |
| 121 | 1 | 1 | 1 | 1 | 1 | 1 | 1 | 1 | 1 | 1 |
| 122 | 1 | 1 | 1 | 1 | 1 | 1 | 1 | 1 | 1 | 1 |
| 123 | 0 | 0 | 0 | 0 | 0 | 0 | 0 | 0 | 0 | 0 |

| ID | UroBot_4o_1 | UroBot_4o_2 | UroBot_4o_3 | UroBot_4o_4 | UroBot_4o_5 | UroBot_4o_6 | UroBot_4o_7 | UroBot_4o_8 | UroBot_4o_9 | UroBot_4o_10 |
|---|---|---|---|---|---|---|---|---|---|---|
| 124 | 1 | 1 | 1 | 1 | 1 | 1 | 1 | 1 | 1 | 1 |
| 125 | 1 | 1 | 1 | 1 | 1 | 1 | 1 | 1 | 1 | 1 |
| 126 | 1 | 1 | 1 | 1 | 1 | 1 | 1 | 1 | 1 | 1 |
| 127 | 1 | 1 | 1 | 1 | 1 | 1 | 1 | 1 | 1 | 1 |
| 128 | 1 | 1 | 1 | 1 | 1 | 1 | 1 | 1 | 1 | 1 |
| 129 | 1 | 1 | 1 | 1 | 1 | 1 | 1 | 1 | 1 | 1 |
| 130 | 1 | 1 | 1 | 1 | 1 | 1 | 1 | 1 | 1 | 1 |
| 131 | 1 | 1 | 1 | 1 | 1 | 1 | 1 | 1 | 1 | 1 |
| 132 | 1 | 1 | 1 | 1 | 1 | 1 | 1 | 1 | 1 | 1 |
| 133 | 1 | 1 | 1 | 1 | 1 | 1 | 1 | 1 | 1 | 1 |
| 134 | 1 | 1 | 1 | 1 | 1 | 1 | 1 | 1 | 1 | 1 |
| 135 | 1 | 1 | 1 | 1 | 1 | 1 | 1 | 1 | 1 | 1 |
| 136 | 1 | 1 | 1 | 1 | 1 | 1 | 1 | 1 | 1 | 1 |
| 137 | 1 | 1 | 1 | 1 | 1 | 1 | 1 | 1 | 1 | 1 |
| 138 | 1 | 1 | 1 | 1 | 1 | 1 | 1 | 1 | 1 | 1 |
| 139 | 1 | 0 | 0 | 0 | 0 | 0 | 0 | 0 | 0 | 1 |
| 140 | 1 | 1 | 1 | 1 | 1 | 1 | 1 | 1 | 1 | 1 |
| 141 | 1 | 1 | 1 | 1 | 1 | 1 | 1 | 1 | 1 | 1 |
| 142 | 1 | 1 | 1 | 1 | 1 | 1 | 1 | 1 | 1 | 1 |
| 143 | 1 | 1 | 1 | 1 | 1 | 1 | 1 | 1 | 1 | 1 |
| 144 | 1 | 1 | 1 | 1 | 1 | 1 | 1 | 1 | 1 | 1 |
| 145 | 1 | 1 | 1 | 1 | 1 | 1 | 1 | 1 | 1 | 1 |
| 146 | 1 | 1 | 1 | 1 | 1 | 1 | 1 | 1 | 1 | 1 |
| 147 | 1 | 1 | 1 | 1 | 1 | 1 | 1 | 1 | 1 | 1 |
| 148 | 1 | 1 | 1 | 1 | 1 | 1 | 1 | 0 | 1 | 0 |
| 149 | 1 | 1 | 1 | 1 | 1 | 1 | 1 | 1 | 1 | 1 |

| ID | UroBot_4o_1 | UroBot_4o_2 | UroBot_4o_3 | UroBot_4o_4 | UroBot_4o_5 | UroBot_4o_6 | UroBot_4o_7 | UroBot_4o_8 | UroBot_4o_9 | UroBot_4o_10 |
|---|---|---|---|---|---|---|---|---|---|---|
| 150 | 1 | 1 | 1 | 1 | 1 | 1 | 1 | 1 | 1 | 1 |
| 151 | 1 | 1 | 1 | 1 | 1 | 1 | 1 | 1 | 1 | 1 |
| 152 | 1 | 1 | 1 | 1 | 1 | 1 | 1 | 1 | 1 | 1 |
| 153 | 0 | 0 | 0 | 0 | 0 | 0 | 0 | 0 | 0 | 0 |
| 154 | 1 | 1 | 1 | 1 | 1 | 1 | 1 | 1 | 1 | 1 |
| 155 | 1 | 1 | 1 | 1 | 1 | 1 | 1 | 1 | 1 | 1 |
| 156 | 1 | 1 | 1 | 1 | 1 | 1 | 1 | 1 | 1 | 1 |
| 157 | 1 | 1 | 1 | 1 | 1 | 1 | 1 | 1 | 1 | 1 |
| 158 | 1 | 1 | 1 | 1 | 1 | 1 | 1 | 1 | 1 | 1 |
| 159 | 1 | 1 | 1 | 1 | 1 | 1 | 1 | 1 | 1 | 1 |
| 160 | 1 | 1 | 1 | 1 | 1 | 1 | 1 | 1 | 1 | 1 |
| 161 | 1 | 1 | 1 | 1 | 1 | 1 | 1 | 1 | 1 | 1 |
| 162 | 1 | 1 | 1 | 1 | 1 | 1 | 1 | 1 | 1 | 1 |
| 163 | 1 | 1 | 1 | 1 | 1 | 1 | 1 | 1 | 1 | 1 |
| 164 | 1 | 1 | 1 | 1 | 1 | 1 | 1 | 1 | 1 | 1 |
| 165 | 1 | 1 | 1 | 1 | 1 | 1 | 1 | 1 | 1 | 1 |
| 166 | 1 | 1 | 1 | 1 | 1 | 1 | 1 | 1 | 1 | 1 |
| 167 | 1 | 1 | 1 | 1 | 1 | 1 | 1 | 1 | 1 | 1 |
| 168 | 1 | 1 | 1 | 1 | 1 | 1 | 1 | 1 | 1 | 1 |
| 169 | 1 | 1 | 1 | 1 | 1 | 1 | 1 | 1 | 1 | 1 |
| 170 | 1 | 1 | 1 | 1 | 1 | 1 | 1 | 1 | 1 | 1 |
| 171 | 0 | 0 | 0 | 0 | 0 | 0 | 0 | 0 | 0 | 0 |
| 172 | 0 | 0 | 0 | 0 | 0 | 0 | 0 | 0 | 0 | 0 |
| 173 | 1 | 1 | 1 | 1 | 1 | 1 | 1 | 1 | 1 | 1 |
| 174 | 1 | 1 | 1 | 1 | 1 | 1 | 1 | 1 | 1 | 1 |
| 175 | 1 | 1 | 1 | 1 | 1 | 1 | 1 | 1 | 1 | 1 |

| ID | UroBot_4o_1 | UroBot_4o_2 | UroBot_4o_3 | UroBot_4o_4 | UroBot_4o_5 | UroBot_4o_6 | UroBot_4o_7 | UroBot_4o_8 | UroBot_4o_9 | UroBot_4o_10 |
|---|---|---|---|---|---|---|---|---|---|---|
| 176 | 1 | 1 | 1 | 1 | 1 | 1 | 1 | 1 | 1 | 1 |
| 177 | 1 | 1 | 1 | 1 | 1 | 1 | 1 | 1 | 1 | 1 |
| 178 | 1 | 1 | 1 | 1 | 1 | 1 | 1 | 1 | 1 | 1 |
| 179 | 0 | 0 | 0 | 0 | 0 | 0 | 0 | 0 | 0 | 0 |
| 180 | 1 | 1 | 1 | 1 | 1 | 1 | 1 | 1 | 1 | 1 |
| 181 | 1 | 1 | 1 | 1 | 1 | 1 | 1 | 1 | 1 | 1 |
| 182 | 1 | 1 | 1 | 1 | 1 | 1 | 1 | 1 | 1 | 1 |
| 183 | 1 | 1 | 1 | 1 | 1 | 1 | 1 | 1 | 1 | 1 |
| 184 | 1 | 1 | 1 | 1 | 1 | 1 | 1 | 1 | 1 | 1 |
| 185 | 1 | 1 | 1 | 1 | 1 | 1 | 1 | 1 | 1 | 1 |
| 186 | 1 | 1 | 1 | 1 | 1 | 1 | 1 | 1 | 1 | 1 |
| 187 | 1 | 1 | 1 | 1 | 1 | 1 | 1 | 1 | 1 | 1 |
| 188 | 1 | 1 | 1 | 1 | 1 | 1 | 1 | 1 | 1 | 1 |
| 189 | 0 | 0 | 0 | 0 | 0 | 1 | 0 | 0 | 0 | 0 |
| 190 | 1 | 1 | 1 | 1 | 1 | 1 | 1 | 1 | 1 | 1 |
| 191 | 1 | 1 | 1 | 1 | 1 | 1 | 1 | 1 | 1 | 1 |
| 192 | 0 | 0 | 0 | 0 | 0 | 0 | 0 | 0 | 0 | 0 |
| 193 | 0 | 0 | 0 | 0 | 0 | 0 | 0 | 0 | 0 | 0 |
| 194 | 1 | 1 | 1 | 1 | 1 | 1 | 1 | 1 | 1 | 1 |
| 195 | 1 | 1 | 1 | 1 | 1 | 1 | 1 | 1 | 1 | 1 |
| 196 | 1 | 1 | 1 | 1 | 1 | 1 | 1 | 1 | 1 | 1 |
| 197 | 1 | 1 | 1 | 1 | 1 | 1 | 1 | 1 | 1 | 1 |
| 198 | 1 | 1 | 1 | 1 | 1 | 1 | 1 | 1 | 1 | 1 |
| 199 | 1 | 1 | 1 | 1 | 1 | 1 | 1 | 1 | 1 | 1 |
| 200 | 0 | 1 | 1 | 0 | 1 | 1 | 1 | 1 | 0 | 1 |

**Prompt used for benchmarking:**

**UroBot prompt:**
```
updated_query = "You are answering questions of an exam, you must reply only with the character of the correct answer. " \
                f"You are giving an answer based on the following context: \n" \
                f"---" \
                f"{context}" \
                f"--- \n"
```

**GPT-based model prompt:**

```
updated_query = "You are answering questions of an exam," \
                " you must reply only with the character of the correct answer, for example 'D', or 'A'."
```